\documentclass[opre,nonblindrev]{informs3}

\OneAndAHalfSpacedXI 

\usepackage{endnotes}
\usepackage{diagbox}
\usepackage{tikz}
\usepackage{amsmath}
\allowdisplaybreaks
\usepackage{dsfont, pifont, mathrsfs}
\usepackage{multirow}
\usepackage[hidelinks, colorlinks=true, linkcolor=blue, citecolor=blue]{hyperref}
\let\footnote=\endnote
\let\enotesize=\normalsize
\def\notesname{Endnotes}%
\def\makeenmark{$^{\theenmark}$}
\def\enoteformat{\rightskip0pt\leftskip0pt\parindent=1.75em
  \leavevmode\llap{\theenmark.\enskip}}

\usepackage{natbib}
 \bibpunct[, ]{(}{)}{,}{a}{}{,}%
 \def\bibfont{\small}%
\TheoremsNumberedThrough     
\ECRepeatTheorems

\EquationsNumberedThrough    

\newcommand{\cA}{\mathcal A}
\newcommand{\cB}{\mathcal B}
\newcommand{\cC}{\mathcal C}
\newcommand{\cD}{\mathcal D}
\newcommand{\cE}{\mathcal E}
\newcommand{\cF}{\mathcal F}

\newcommand{\cN}{\mathcal N}

\newcommand{\SG}{\text{subG}}

\newcommand{\mbP}{\mathbb P}
\newcommand{\mbE}{\mathbb E}

\newcommand{\rad}{\text{\rm rad}}

\newcommand{\SE}{\text{\rm SE}}
\newcommand{\UCB}{\text{\rm UCB}}

\newcommand{\poly}{\mathrm{poly}}
\newcommand{\polylog}{\mathrm{polylog}}

\newcommand{\highlight}[1]{{\color{black}{#1}}}

\newcommand{\brs}[1]{\left({#1}\right)} 
\newcommand{\ex}[2]{\mathbb{E}_{#1}\left[#2\right]}

\newcommand*{\circled}[1]{\lower.7ex\hbox{\tikz\draw (0pt, 0pt)%
    circle (.5em) node {\makebox[1em][c]{\small #1}};}}

\usepackage{algorithm}
\usepackage{algorithmicx}
\usepackage{algpseudocode}
\algnewcommand{\algorithmicand}{\textbf{and }}
\algnewcommand{\algorithmicor}{\textbf{or }}
\algnewcommand{\OR}{\algorithmicor}
\algnewcommand{\AND}{\algorithmicand}
\makeatletter
\newenvironment{breakablealgorithm}
  {
  \begin{center}
     \refstepcounter{algorithm}
     \hrule height.8pt depth0pt \kern2pt
     \renewcommand{\caption}[2][\relax]{
      {\raggedright\textbf{\ALG@name~\thealgorithm} ##2\par}%
      \ifx\relax##1\relax 
         \addcontentsline{loa}{algorithm}{\protect\numberline{\thealgorithm}##2}%
      \else 
         \addcontentsline{loa}{algorithm}{\protect\numberline{\thealgorithm}##1}%
      \fi
      \kern2pt\hrule\kern2pt
     }
  }{
     \kern2pt\hrule\relax
  \end{center}
  }

\begin{document}


\RUNAUTHOR{Simchi-Levi, Zheng and Zhu}

\RUNTITLE{Regret Distribution in Stochastic Bandits}



\TITLE{Regret Distribution in Stochastic Bandits: \texorpdfstring{\\}{} Optimal Trade-off between Expectation and Tail Risk}


\ARTICLEAUTHORS{%
\AUTHOR{David Simchi-Levi}
\AFF{Institute for Data, Systems, and Society, Massachusetts Institute of Technology, MA 02139, \EMAIL{dslevi@mit.edu}} 
\AUTHOR{Zeyu Zheng}
\AFF{Department of Industrial Engineering and Operations Research, University of California, Berkeley, CA 94720, \EMAIL{zyzheng@berkeley.edu}}
\AUTHOR{Feng Zhu}
\AFF{Institute for Data, Systems, and Society, Massachusetts Institute of Technology, MA 02139,  \EMAIL{fengzhu@mit.edu}}
} 

\ABSTRACT{%

We study the optimal trade-off between expectation and tail risk for regret distribution in the stochastic multi-armed bandit model. We fully characterize the interplay among three desired properties for policy design: worst-case optimality, instance-dependent consistency, and light-tailed risk. New policies are proposed to characterize the optimal regret tail probability for any regret threshold. In particular, we discover an intrinsic gap of the optimal tail rate depending on whether the time horizon $T$ is known a priori or not. Interestingly, when it comes to the purely worst-case scenario, this gap disappears. Our results reveal insights on how to design policies that balance between efficiency and safety, and highlight extra insights on policy robustness with regard to policy hyper-parameters and model mis-specification. We also conduct a simulation study to validate our theoretical insights and provide practical amendment to our policies. Finally, we discuss extensions of our results to (i) general sub-exponential environments and (ii) general stochastic linear bandits. Furthermore, we find that a special case of our policy design surprisingly coincides with what was adopted in AlphaGo Monte Carlo Tree Search. Our theory provides high-level insights to why their engineered solution is successful and should be advocated in complex decision-making environments.
}%


\maketitle

%


\section{Introduction}

The stochastic multi-armed bandit (MAB) problem is a well-established area of research in online decision-making under uncertainty, applications of which include online advertising, recommendation systems, digital clinical trials, etc. In the stochastic MAB problem, in each time period, based on the information collected previously, the decision maker selects one of several arms, each of which owns an unknown reward distribution, with the goal of maximizing the expected sum of rewards over all time periods. The MAB problem highlights the exploration-exploitation trade-off, where the decision maker must balance between exploring arms with relatively unknown reward distributions and exploiting arms with relatively known high expected rewards. There is a significant amount of literature on MAB, with a comprehensive review provided in \cite{slivkins2019introduction} and \cite{lattimore2020bandit}. 

In order to evaluate policy performance and guide policy design, a commonly used metric is ``expectation". In the MAB setting, one typically use the metric of maximizing the expected total (or cumulative) reward, or equivalently minimizing the ``expected regret", where \textit{regret} is defined as the difference between the cumulative reward of always pulling the best arm and the cumulative reward of a policy. However, a recent work \citet{fan2021fragility} has shown that optimized policy designs may lead the policy to have heavy-tailed risks of incurring a large regret --- the probability of incurring a linear regret slowly decays at a polynomial rate $\Omega(1/T)$ as $T$ tends to infinity. As shown in a subsequent work \citet{simchi2022simple}, all instance-dependent consistent policies (including many renowned algorithms such as Upper Confidence Bound (UCB) \citep{auer2002finite}, Successive Elimination (SE) \citep{even2006action}, Thompson Sampling (TS) \citep{russo2018tutorial}), despite of enjoying optimality on the order of expected regret, can incur a heavy-tailed risk on regret distribution. Roughly speaking, designing a policy that focuses only on the expected regret could be analogous to designing an investment portfolio that focuses only on the expected return without looking at other risks.  In contrast, a ``light-tailed'' risk in this MAB setting means that ideally, the probability of a policy incurring a linear regret decays at an exponential rate $\exp(-\Omega(T^\gamma))$ for some $\gamma > 0$. 

Two important notions that are popularly used in MAB literature to describe properties of a policy --- worst-case \textit{optimality} and instance-dependent \textit{consistency} --- are both defined in terms of expected regret. \cite{simchi2022simple} showed that consistency (e.g., obtaining $\tilde O(1)$ instance-dependent regret) always causes heavy-tailed risk, while optimality (e.g., obtaining $\tilde O(\sqrt{T})$ worst-case regret) allows light-tailed risk. There lacks an understanding in the literature about how much tail risk will arise by adjusting optimality and/or consistency for policy design. It remains an open question and is what we answer in this work: 
\begin{center}
    \emph{What is the optimal trade-off between regret expectation and regret tail risk?} \\
    \emph{How do different levels of optimality and consistency jointly affect the tail risk?}
\end{center}
\highlight{
Along with answering the two questions, we also find that the following message holds ---
\begin{center}
    {\it Controlling regret tail risk leads to extra benefits on improving policy robustness!}
\end{center}}
We summarize our contributions in Section \ref{ssec:contribution}. To facilitate describing the results on regret orders and function orders, we adopt $O(\cdot)$ ($\tilde O(\cdot)$) and $\Omega(\cdot)$ ($\tilde\Omega(\cdot)$) to present upper and lower bounds on the growth rate up to constant (logarithmic) factors, respectively, and $\Theta(\cdot)$ ($\tilde\Theta(\cdot)$) to characterize the rate when the upper and lower bounds match up to constant (logarithmic) factors. We use $o(\cdot)$ and $\omega(\cdot)$ to present strictly dominating upper bounds and strictly dominated lower bounds, respectively.

\subsection{Our Contributions}

\label{ssec:contribution}

\begin{enumerate}
    \item We fully characterize the connections and interplay between the order of expectation and the order of tail risk for regret distribution in stochastic multi-armed bandits. We show how relaxing worst-case expected regret order or instance-dependent expected regret order can help make the regret tail lighter in an information-theoretic way. We characterize that given the family of policies with a worst-case regret of $\tilde O(T^\alpha)$ and an instance-dependent regret of $O(f(T))$ such that $f(T)=\omega(\ln T)$, how fast we can best hope the probability of incurring a regret of $x(T)$ decays with $T$. In particular, we differentiate between the situations where the policy knows the time horizon $T$ in advance or not. We find that in the instance-dependent scenario, knowing $T$ helps make the tail much lighter than the case when $T$ is known a priori; while in a pure worst-case scenario, such gap does not exist.
 
    \item We design simple policies that, for any given $\alpha\in[1/2, 1)$ and $f(\cdot)$ , obtains $ O(K^{1-\alpha}T^\alpha\sqrt{\ln K})$ worst-case regret (\textit{without $\ln T$ factors!}) and $O(f(T))$ instance-dependent regret, whereas obtains the best achievable regret tail probability for both worst-case and instance-dependent scenarios, and for each scenario, whether $T$ is known or not. Table \ref{table:contribution} shows the dependence of our regret tail bounds on the time horizon $T$ and the regret threshold $x$ under $(2\times 2=)4$ different cases. Detailed tail bounds will be provided in the main content. The setting considered in \cite{simchi2022simple} is a special case of ours in the worst-case scenario, and we further improve their result by reducing the $\ln T$ factor into a $\ln K$ one. Our refined analysis might be of independent interest. Our results reveal insights on how to design policies that balance regret expectation and regret tail risk, indicating that (i) less ambitious expectation goals leave space for less tail risk, (ii) a little sacrifice in consistency can greatly reduce tail risk and boost policy robustness under mis-specified volatility parameters, and (iii) knowing the planning horizon in advance can make an intrinsic difference on alleviating tail risk. We also conduct a series of numerical experiments to discuss safety/robustness performance and hyperparameter tuning of our policy designs.

    \begin{table}[!ht]
    \centering
    \renewcommand{\arraystretch}{1.5}
    \begin{tabular}{|c|c|c|}\hline
        & known $T$ & unknown $T$  \\\hline
        \multirow{2}{*}{\shortstack{$\mbP_{\theta, \cD}^\pi(\text{Regret}>x)$ \\ (instance-dependent scenario)}} & \multirow{2}{*}{\shortstack{$\exp(-\Theta(f(T)))$ \\ for $x=\Omega(f(T))$}} & \multirow{2}{*}{\shortstack{$\exp(-\Theta(f(x)))$ \\
        for $x=\Omega(f(T))$}}  \\
         & & \\\hline
        \multirow{2}{*}{\shortstack{$\sup_{\theta, \cD}\mbP_{\theta, \cD}^\pi(\text{Regret}>x)$ \\ (worst-case scenario)}} & \multirow{2}{*}{\shortstack{$\exp\brs{-\Theta((x/T^{1-\alpha})\wedge f(T))}$ \\ for $x=\Omega(T^\alpha)$}}  & \multirow{2}{*}{\shortstack{$\exp\brs{-\Theta((x/T^{1-\alpha})\wedge f(x))}$ \\ for $x=\Omega(T^\alpha)$}} \\
         & & \\\hline
    \end{tabular}
    \caption{\centering Optimal regret tail for the family of policies that obtain both $O(T^\alpha)$ worst-case and $O(f(T))$ instance-dependent expected regret}
    \label{table:contribution}
    \end{table}

    \item We extend the idea and analysis of our policy design to models that allow additional features beyond standard stochastic multi-armed bandits. We consider (i) the stochastic MAB setting with sub-exponential random noises, and (ii) the stochastic linear bandit setting where the decision maker chooses an action in each time period from a potentially time-varying continuous action set. We show that simple modifications to our policy designs allow us to obtain safe and robust performance similar to those for the stochastic MAB model. Moreover, we discuss a surprising relationship between our policy design and the Monte Carlo Tree Search in AlphaGo, revealing theoretical insights on why the engineered solution in AlphaGo should be advocated when facing the exploration-exploitation dilemma in complex decision-making environments.
\end{enumerate}

Our policy designs build upon constructing novel confidence bounds to balance among worst-case optimality, instance-dependent consistency, and light-tailed risk, highlighting a \textit{phase transition} that in order to achieve more light-tailed risk of incurring a large regret, it might be beneficial to have two different phases in the policy design: more exploration at the beginning within the instance-dependent consistency constraint, and more exploitation afterwards within the worst-case optimality condition. As far as we know, we are the very first to optimally and completely characterize the trade-off between expectation and tail risk from different aspects (worst-case \& instance-dependent, known $T$ \& unknown $T$) in the broad online learning literature. Despite of the simplicity of our proposed policy designs, the associated proof techniques are novel and may be useful for broader analysis related with regret tail risk. In particular, for the standard MAB setting, we refine and generalize the \textit{split-and-conquer} technique developed in \cite{simchi2022simple} adaptively  according to different scenarios (worst-case and instance-dependent) to achieve optimal dependence on both $T$ and $K$ under the case when $T$ is known, which is then further improved to handle (i) the any-time case without knowing $T$, (ii) general sub-exponential environments, and (iii) the general linear bandit setting.

\subsection{Related Work}
\label{ssec:literature}

Our work is situated within the stochastic multi-armed bandit (MAB) literature. Relevant reviews can be found in \citet{bubeck2012regret, russo2018tutorial, slivkins2019introduction, lattimore2020bandit}. Below we review the relevant works from several different perspectives based on whether they are concerned with regret tail or not.

{\bf Regret Tail of Bandit Algorithms.} The tail risk of stochastic bandit algorithms remains under-explored compared to their expected performance, and most prior work related with ours studied the concentration properties of regret around the instance-dependent mean. We briefly describe these prior works as follows.

\cite{audibert2009exploration} and \cite{salomon2011deviations} studied the regret concentration properties around the instance-dependent mean of $O(\ln T)$. They found that the regret distribution of standard policies such as $\UCB$ typically only concentrates at a polynomial rate. Specifically, the probability of incurring a regret of $x$ (where $x=\omega(\ln T)$) decays at a polynomial $T$ rate. \cite{salomon2011deviations} also showed that for any-time policies (i.e., policies that do not use the time horizon $T$), it is impossible to achieve even polynomial concentration rate around the instance-dependent $O(\ln T)$ expected regret. These results suggest that standard bandit algorithms may have undesirable concentration properties, and any-time policies may be surprisingly weaker than policies that have access to the time horizon information in advance in terms of high probability bounds. We investigate this phenomenon further in our paper by quantitatively analyzing how relaxing the full consistency constraint (which forces a policy to achieve $O(\ln T)$ instance-dependent regret) can help make the regret tail lighter and influence the gap between knowing the time horizon in advance or not.

In a recent study, \cite{ashutosh2021bandit} demonstrated that an online learning policy aiming to achieve logarithmic expected regret is not robust, in the sense that a mis-specified risk parameter in the policy can cause an instance-dependent expected regret of $\omega(\ln T)$. Such risk parameter may include the parameter for sub-gaussian noises, for example.  To address this issue, they developed robust algorithms. It is worth noting that their primary objective is to handle mis-specification related to risk while still minimizing the order of expected regret.

Our work is built upon the insights from \cite{fan2021fragility} and \cite{simchi2022simple}. \cite{fan2021fragility} analyzed the heavy-tailed risk in bandit and showed that information-theoretically optimized bandit policies suffer from severe heavy-tailed risk: the probability of incurring a linear regret is at least $\Omega(1/T)$. They also showed that UCB algorithms can suffer from the heavy-tailed risk and proposed a modification of UCB algorithms that achieve the desired tail risk polynomially dependent on $T$, improving the robustness of the algorithms to mis-specification. \cite{simchi2022simple} further showed the general incompatibility between instance-dependent consistency and light-tailed risk, illustrating that one can not expect an algorithm to enjoy light-tailed risk if the algorithm achieves instance dependent consistency. They highlighted that a simple policy design maintaining worst-case optimality can achieve optimal light-tailed risk. 

\cite{fan2021fragility} used the metric of expected regret under the instance-dependent scenario, and \cite{simchi2022simple} on the other hand, released requirements on instance-dependent consistency in their policy design to obtain light-tailed risk and worst-case optimality. In addition to these two papers, the optimal trade-off among optimality, consistency, and tail risk remains unclear, which is the question that we hope to address in this work. We fill the gap of the optimal regret tail beyond full consistency (i.e., a policy with $\tilde O(1)$ expected regret) and full optimality (i.e., a policy with $\tilde O(\sqrt{T})$ expected regret), and show the optimal trade-off between  regret expectation and tail risk. Moreover, we differentiate between the cases of knowing $T$ in advance or not, and show a delicate but intrinsic gap between these two cases, which has not been discussed in previous works.

{\bf Limit Behaviour of Bandit Algorithms.} There is also a line of works analyzing the limit behaviour of standard UCB and TS policies by considering the diffusion approximations (see, e.g., \citealt{araman2021diffusion, wager2021diffusion,fan2021diffusion, kalvit2021closer}). While these works typically consider asymptotic limiting regimes that are set such that the gaps between arm means shrink with the total time horizon, we do not consider such limiting regimes but instead consider the original problem setting and study how the tail probability decays with $T$ under original environments without taking the gaps to zero.

{\bf Multi-objective Bandits.} Our work is related to multi-objective bandit problems where the objective is not solely focused on minimizing expected regret. In this context, several works have been proposed, such as \cite{deshmukh2017multi, erraqabi2017trading, yang2017framework, yao2021power, simchi2022multi}. In particular, \cite{simchi2022multi} focused on the trade-off between efficiency (low regret) and statistical power (accurate estimation of arm gaps), and provided an optimal trade-off through an information-theoretic lower bound and a policy-generated upper bound. However, it is important to note that their approach is still centered on the notion of expectation and based on an instance-dependent perspective.

{\bf Risk-averse Bandits.} Another line of related work is risk-averse formulations of the stochastic MAB problem (e.g., \citealt{sani2012risk, galichet2013exploration, maillard2013robust, zimin2014generalized, vakili2016risk, cassel2018general, tamkin2019distributionally, prashanth2020concentration, zhu2020thompson, baudry2021optimal, khajonchotpanya2021revised, chen2022bridging, chang2022unifying}). These formulations consider different notions than expected regret, such as mean-variance or value-at-risk. In contrast, our work focuses on the levels of tail risks and develops policies that maintain low expected regret while achieving light-tailed risk bounds. This leads to different policy design and analysis than the risk-averse formulations.

{\bf Heavy-tailed Bandits.} Many works have contributed to the understanding of heavy-tailed bandit problems and have developed algorithms that can achieve optimal expected regret bounds under heavy-tailed distributions (see, e.g., \citealt{bubeck2013bandits, lattimore2017scale, yu2018pure, lugosi2019mean, lee2020optimal, agrawal2021regret, bhatt2022nearly, tao2022optimal}). In their settings, the rewards generated by the arms have heavy-tailed distributions, an so the challenge lies in efficient estimation of mean rewards from heavy-tailed distributions. As a comparison, we focus on light-tailed reward distributions, and so the mean estimation of arms is not difficult, but the challenge becomes how to achieve a regret distribution as much light-tailed as possible. Nevertheless, we believe our results might be of independent interest to this line of research.

\subsection{Organization and Notation}

The rest of the paper is organized as follows. In Section \ref{sec:setup}, we discuss the basic setup and introduce the concepts related with regret expectation (worst-case optimality, instance-dependent consistency) and regret distribution (tail risk). In Section \ref{sec:lower}, we show the trade-off between regret expectation and light-tailed risk via information-theoretic lower bounds under different scenarios and cases. In Section \ref{sec:upper}, we look into the general stochastic $K$-armed bandit model and design new policies with explicit regret tail upper bounds that match the lower bounds in Section \ref{sec:lower}. In Section \ref{sec:extension}, we show how to extend our policy design into more stochastic bandit settings with structured non-stationarity, and obtain light-tailed regret bounds similar to those in Section \ref{sec:upper}. We also discuss the relationship between our policy designs and AlphaGo. Finally, we conclude in Section \ref{sec:conclusion}. All detailed proofs are left to the supplementary material.

Before proceeding, we introduce some notation.  For any $a, b\in\mathbb R$, $a\wedge b = \min\{a, b\}$ and $a\vee b = \max\{a, b\}$. For any $a\in\mathbb R$, $a_+ = \max\{a, 0\}$. We denote $[N]=\{1, \cdots, N\}$ for any positive integer $N$. Throughout the paper, we use $O(\cdot)$ ($\tilde O(\cdot)$) and $\Omega(\cdot)$ ($\tilde\Omega(\cdot)$) to present upper and lower bounds on the growth rate up to constant (logarithmic) factors, respectively, and $\Theta(\cdot)$ ($\tilde\Theta(\cdot)$) to characterize the rate when the upper and lower bounds match up to constant (logarithmic) factors. We use $o(\cdot)$ and $\omega(\cdot)$ to present strictly dominating upper bounds and strictly dominated lower bounds, respectively.

\section{The Setup} \label{sec:setup}

In this section, we first discuss the model setup. We then formally define the terms that appeared in the introduction and will appear in the rest of this work: $\alpha$-optimality, $\beta$-consistency, and $(\delta, \gamma)$-tail.

Fix a time horizon of $T$ and the number of arms as $K$. Throughout the paper, we assume that $T\geq 3$, $K\geq 2$, and $T\geq K$. In each time $t\in[T]$, based on all the information prior to time $t$, the decision maker (DM) pulls an arm $a_t\in[K]$ and receives a reward $r_{t, a_t}$. More specifically, let $H_t = \{a_1, r_{1, a_1}, \cdots, a_{t-1}, r_{t-1, a_{t-1}}\}$ be the history prior to time $t$. When $t=1$, $H_1=\emptyset$. We differentiate between two cases: knowing $T$ a priori or not. 

\begin{itemize}
    \item ($T$ is known) At time $t$, an admissible \textit{fixed-time} policy $\pi_{t}(T): H_t\cup\{T\}\longmapsto a_t$ maps the history $H_t\cup\{T\}$ to an action $a_t$ that may be realized from a discrete probability distribution on $[K]$. 
    \item ($T$ is unknown) At time $t$, an admissible \textit{any-time} policy $\pi_t: H_t\longmapsto a_t$ maps the history $H_t$ to an action $a_t$ that may be realized from a discrete probability distribution on $[K]$. 
\end{itemize}
In this paper, we will always make clear whether a policy is fixed-time or any-time. When we say $\pi$ is a fixed time policy, we mean that $\pi$ is composed of a series of ``sub-policies":
\begin{align*}
    \pi(1), \cdots, \pi(T), \cdots.
\end{align*}
That is, with the prior knowledge of $T$, $\pi$ executes $\pi(T)$ throughout the whole time horizon. At time $t$, the action taken is determined by $\pi_t(T)$. Different $T$'s may lead to completely different sub-policies. When we say $\pi$ is an any-time policy, we mean that regardless of the value of $T$, the action is always determined by $\pi_t$ at time $t$. One can easily observe that an any-time policy is always a fixed-time policy by taking $\pi(T) = \pi$ for any $T$, but the reverse is not necessarily true.

After an action $a_t$ is taken, the environment independently samples a reward $r_{t, a_t}=\theta_{a_t} + \epsilon_{t, a_t}$ and reveals it to the DM. Here, $\theta_{a_t}$ is the mean reward of arm $a_t$, and $\epsilon_{t, a_t}$ is an independent zero-mean noise term. We assume that $\epsilon_{t, a_t}$ is $\sigma$-sub-gaussian. That is, there exists a $\sigma>0$ such that for any time $t$ and arm $k$, $\max\left\{\mbP\left(\epsilon_{t, k} \geq x\right), \mbP\left(\epsilon_{t, k} \leq -x\right)\right\} \leq \exp(-x^2/(2\sigma^2))$. We will refer to $\sigma$ as the \textit{volatility parameter} or the \textit{volatility profile} as a measure of the intrinsic risk in the environment, borrowing terminologies from finance. We avoid using ``risk parameter'' to separate from the ``risk'' in ``tail risk''. In Section \ref{sec:extension}, we will discuss extensions of environments beyond sub-gaussian noises.

Let $\theta = (\theta_1, \cdots, \theta_K)$ be the mean vector and $\theta_* = \max\{\theta_1, \cdots, \theta_K\}$ be the optimal mean reward among the $K$ arms. Note that DM does not know both information at the beginning, except that $\theta\in[0, 1]^K$. \highlight{The noise distribution is characterized via $\cD = (\cD_1, \cdots, \cD_K)$, where $\cD_k$ is the distribution of the noise term from arm $k$. We assume each $\cD_k$ belongs the class of sub-Gaussian distributions with parameter $\sigma$, denoted as $\SG(\sigma)$ (we will relax this assumption when we discuss extensions in Section \ref{sec:extension}). The empirical regret of the policy $\pi$ (either fixed-time or any-time) under $\cD$ over a time horizon of $T$ is defined as
\begin{align*}
    \hat R_{\theta, \cD}^\pi(T) \triangleq \hat R_{\theta, \cD}^{\pi(T)}(T) = \theta_*\cdot T - \sum_{t=1}^T(\theta_{a_t} + \epsilon_{t, a_t}).
\end{align*}}
Let $\Delta_k = \theta_* - \theta_k$ be the gap between the optimal arm and the $k$th arm. Let $n_{t, k}$ be the number of times arm $k$ has been pulled up to time $t$: $n_{t, k} = \sum_{s=1}^t\mathds 1\{a_s = k\}$. For simplicity, we will also use $n_{k}=n_{T, k}$ to denote the total number of times arm $k$ is pulled throughout the whole time horizon $T$. We define $t_k(n)$ as the time period that arm $k$ is pulled for the $n$th time. Define the pseudo regret and the genuine noise respectively as 
\begin{align*}
    R_{\theta, \cD}^\pi(T) \triangleq R_{\theta, \cD}^{\pi(T)}(T) = \sum_{k=1}^K n_k\Delta_k, \quad N^{\pi}(T) \triangleq N^{\pi(T)}(T) = \sum_{t=1}^T \epsilon_{t, a_t} = \sum_{k=1}^K\sum_{m=1}^{n_k}\epsilon_{t_k(m), k}.
\end{align*}
Then the empirical regret can also be written as $\hat R_{\theta, \cD}^\pi(T) = R_{\theta, \cD}^\pi(T) - N^\pi(T)$. We note that for all the cases considered in this paper, the environment admits $\sigma$-sub-gaussian noises by default, where $\sigma$ is an environment parameter. In our notations, we do not explicitly write $\sigma$ in the definition of regret and noise. The following lemma shows the concentration property of $N^\pi(T)$.
\begin{lemma}\label{lemma:bound-noise}
We have $\mbE[N^\pi(T)] = 0$ and
\[
    \max\left\{\mbP\left(N^\pi(T) \geq x \right), \mbP\left(N^\pi(T) \leq -x\right)\right\} \leq \exp\left(\frac{-x^2}{2\sigma^2T}\right).
\]
\end{lemma}
In the worst-case scenario, the expected regret is at least $\Omega(\sqrt{T})$, and so the tail in Lemma \ref{lemma:bound-noise} is negligible. In the instance-dependent scenario, the expected regret can achieve $o(\sqrt{T})$, making the tail in Lemma \ref{lemma:bound-noise} no longer ignorable. We note that even we always pull the optimal arm, the empirical regret unavoidably incurs a tail in Lemma \ref{lemma:bound-noise} due to the appearance of genuine noise. Therefore, we will focus on pseudo regret $R_{\theta, \cD}^\pi(T)$ in our subsequent discussions (see also, e.g., \citealt{audibert2009exploration, salomon2011deviations}).

\subsection{Regret Expectation and Tail Risk}

Now we describe concepts that are needed to formalize the policy design and analysis.

{\noindent\bf 1. Regret Expectation.} Fix $\alpha\in[1/2, 1)$ and $\beta\in[0, 1)$. We differentiate between two scenarios: worst-case and instance-dependent. 
\begin{itemize}
    \item[(a)] A fixed-time policy $\pi$ is said to be worst-case $\alpha$-optimal or simply, $\alpha$-optimal, if for any $\varepsilon > 0$, we have
    \begin{align*}
        \limsup_{T\to+\infty} \frac{\sup_{\theta, \cD}\mbE\left[R_{\theta, \cD}^{\pi}(T)\right]}{T^{\alpha+\varepsilon}} = 0.
    \end{align*}
    In brief, a fixed-time policy $\pi$ is $\alpha$-optimal if the worst-case expected regret (over all $\theta\in[0, 1]$ and all $\cD\in\SG(\sigma)^K$) can never be growing in $T$ at a polynomial rate faster than $T^\alpha$. Intuitively, the smaller the $\alpha$ is, the better performance a policy has in terms of worst-case expected regret order.

    \item[(b)] A fixed-time policy $\pi$ is said to be instance-dependent $\beta$-consistent or simply, \textbf{$\beta$-consistent}, if for any $\theta$, $\cD$, and any $\varepsilon > 0$, we have
    \begin{align*}
        \limsup_{T\to+\infty} \frac{\mbE\left[R_{\theta, \cD}^{\pi}(T)\right]}{T^{\beta+\varepsilon}} = 0.
    \end{align*}
    In brief, a sequence of policies is $\beta$-consistent if the expected regret can never grow faster than $T^\beta$ for any fixed instance. Intuitively, the smaller the $\beta$ is, the better performance a policy has in terms of instance-dependent expected regret order.
\end{itemize}

We note that the ``worst-case" notion and the ``instance-dependent" notion, in these two items, are most commonly used in the bandits literature, and both notions care about the \textit{expectation} of the regret distribution. The next notion concerns the tail of regret distribution.

{\noindent\bf 2. Regret Tail Risk.} Fix $\delta\in(0, 1]$ and $\gamma\in[0, 1]$. We differentiate between two scenarios: worst-case and instance-dependent.
\begin{itemize}
    \item[(a)] A fixed-time policy is worst-case light-tailed, if there exists some $\gamma > 0$ such that for any constant $c > 0$, there exists a constant $C > 0$ such that
    \begin{align*}
        \limsup_{T\to+\infty} \frac{\ln\left\{\sup_{\theta, \cD}\mbP\left(R_{\theta, \cD}^\pi(T) > cT\right)\right\}}{T^{\gamma}} \leq -C.
    \end{align*}
    More generally, a fixed-time policy $\pi$ is worst-case $(\delta, \gamma)$-tailed, if for any constant $c\in(0, 1/2)$ there exists a constant $C > 0$ such that
    \begin{align*}
        \limsup_{T\to+\infty} \frac{\ln\left\{\sup_{\theta, \cD}\mbP\left(R_{\theta, \cD}^{\pi}(T) > cT^\delta\right)\right\}}{T^{\gamma}} \leq -C.
    \end{align*}
    In brief, a sequence of policies is worst-case $(\delta, \gamma)$-tailed if the worst-case probability of incurring a regret of $T^\delta$ can be bounded by an exponential term of polynomial $T^\gamma$:
    \begin{align*}
        \sup_{\theta, \cD}\mathbb P\left(R_{\theta, \cD}^{\pi}(T)> cT^\delta\right) = \exp(-\Omega(T^{\gamma})).
    \end{align*}
    
    \item[(b)] A fixed-time policy is instance-dependent light-tailed, if there exists some $\gamma > 0$ such that for any underlying true mean vector $\theta$ and any constant $c > 0$, there exists a constant $C > 0$ such that
    \begin{align*}
        \limsup_{T\to+\infty} \frac{\ln\left\{\mbP\left(R_{\theta, \cD}^\pi(T) > cT\right)\right\}}{T^{\gamma}} \leq -C.
    \end{align*}
    More generally, a fixed-time policy $\pi$ is instance-dependent $(\delta, \gamma)$-tailed, if for any underlying true mean vector $\theta$ and any constant $c\in(0, 1/2)$, there exists a constant $C > 0$ such that
    \begin{align*}
        \limsup_{T\to+\infty} \frac{\ln\left\{\mbP\left(R_{\theta, \cD}^{\pi}(T) > cT^\delta\right)\right\}}{T^{\gamma}} \leq -C.
    \end{align*}
    In brief, a sequence of policies is instance-dependent $(\delta, \gamma)$-tailed if the instance-dependent probability of incurring a regret of $T^\delta$ can be bounded by an exponential term of polynomial $T^\gamma$:
    \begin{align*}
        \mathbb P\left(R_{\theta, \cD}^{\pi}(T) > cT^\delta\right) = \exp(-\Omega(T^{\gamma})).
    \end{align*}
\end{itemize}

We would like to give some remarks on the definitions above. 

\begin{enumerate}
    \item For worst-case optimality, here we adopt a relaxed definition, in the sense that we do not clarify how the regret scales with the number of arms $K$ compared to that in literature. The notion of worst-case optimality in this work focuses on the dependence on $T$. For example, a policy with worst-case regret $O(\poly(K)T^\alpha\cdot\poly(\ln T))$ is also $\alpha$-optimal by our definition.

    \item For instance-dependent consistency, here we focus on polynomial growth on expectation and polynomial decay on tail risk to make definitions neat and clear. This is also in accordance with \cite{simchi2022simple} where it is shown that, translated into our language, any $0$-consistent policy is heavy-tailed. Nevertheless, in our main results, we will provide a complete picture for the class of $0$-consistent policies including those that achieve instance-dependent $\polylog(T)$ expected regret.
    
    \item When defining the tail, we impose $c\in(0, 1/2)$ to avoid the corner case when $\delta=1$. In such case, if $c\geq 1$, the tail probability is zero because $\theta\in[0, 1]^K$. We note that when $\delta < 1$, the condition $c\in(0, 1/2)$ is not essential, and here we retain it for simplicity of exposition.
    
    \item An $\alpha$-optimal policy is always $\alpha$-consistent, but the reverse does not hold. Similarly, a worst-case $(\delta, \gamma)$-tailed policy is also instance-dependent $(\delta, \gamma)$-tailed, but the reverse does not hold. We can also claim that if a policy is \textit{not} $\beta$-consistent, then it is also not $\beta$-optimal. Similarly, if a policy is \textit{not} instance-dependent $(\delta, \gamma)$-tailed, then it is also not worst-case $(\delta, \gamma)$-tailed.
    
    \item It is well known that for the stochastic MAB problem, one can design algorithms to achieve both $0$-consistency and $1/2$-optimality using Upper Confidence Bound ($\UCB$, Algorithm \ref{alg:UCB}). The bonus term (or, the confidence radius) $\rad(n)$ is typically set as
    \begin{align} \label{rad:standard}
        \rad(n) = \eta\sqrt{\frac{\ln T}{n}}
    \end{align}
    with $\eta>0$ being some tuning parameter. In both algorithms, $\hat\theta_{t, k}$ is the empirical mean reward of arm $k$ up to time $t$. That said, both the SE policy and UCB policy may not perform well in terms of tail probability of incurring a large regret, as documented in \citet{fan2021fragility} and \citet{simchi2022simple}.
\end{enumerate}

\bigskip
\begin{breakablealgorithm}
\caption{Upper Confidence Bound}
\label{alg:UCB}
\begin{algorithmic}[1]
\State $\cA = [K]$. $t\gets 1$.
\While{$t \leq T$}
    \State Pull the arm with the highest $\UCB$: $\arg\max_k\left\{\hat\theta_{t-1, k} + \rad(n_{t-1, k})\right\}$.
    \State Collect reward $r_{t, a_t}$. $t\gets t+1$.
\EndWhile
\end{algorithmic}
\end{breakablealgorithm}
\bigskip

\section{Tail Lower Bound: The Best to Hope}
\label{sec:lower}

In this section, we show how fast the regret tail can decay as a function of $T$, given that a policy is $\alpha$-optimal or/and $\beta$-consistent. More concretely, if a policy is $\alpha$-optimal or/and $\beta$-consistent, what is the fastest decaying rate we can hope for the probability that the pseudo regret is at least $\Omega(T^\delta)$ (in either the worst-case or instance-dependent scenario). This question is addressed in Theorem \ref{thm:lower-bound-fixed-time}, where we show in an information-theoretic sense, how the regret \textit{tail} can be decaying with $T$ as a function of regret \textit{expectation}. We focus on the simple two-armed bandit setting with Gaussian noises.

\begin{theorem} \label{thm:lower-bound-fixed-time}
Consider the two-armed bandit problem. We have the following arguments.

1. Let $\pi$ be a fixed-time policy such that
\begin{align*}
    \limsup_{T\to+\infty}\frac{\sup_{\theta, \cD}\mbE\left[R_{\theta, \cD}^{\pi}(T)\right]}{T} = 0.
\end{align*}
Let $x(1), \cdots, x(T), \cdots$ be a sequence of numbers such that 
\begin{align*}
    \limsup_{T\to+\infty}\frac{x(T)}{T} < 1/2, \quad \limsup_{T\to+\infty}\frac{\sup_{\theta, \cD}\mbE\left[R_{\theta, \cD}^{\pi}(T)\right]}{x(T)} = 0.
\end{align*}
Then we have
\begin{align*}
    \lim\inf_T\left\{\frac{\ln\left\{\sup_{\theta, \cD}\mbP\left(R_{\theta, \cD}^{\pi}(T) > x(T)\right)\right\}\cdot T}{x(T)\cdot\sup_{\theta, \cD}\mbE\left[R_{\theta, \cD}^{\pi}(T)\right]}\cdot \min\left\{1, \sqrt{\frac{x(T)\cdot\sup_{\theta, \cD}\mbE\left[R_{\theta, \cD}^{\pi}(T)\right]}{T\ln T}}\right\}\right\} \geq -C
\end{align*}
holds for some $C>0$ only dependent on $\sigma$.

2. Let $\pi$ be a fixed-time policy such that
\begin{align*}
    \limsup_{T\to+\infty}\frac{\mbE\left[R_{\theta, \cD}^{\pi}(T)\right]}{T} = 0
\end{align*}
for any $\theta$ and $\cD=(N(0, \sigma^2), N(0, \sigma^2))$. Fix any $\theta$ and $\tilde\theta$ such that $\theta_2>\theta_1 = \tilde\theta_1>\tilde\theta_2$. Let $x(1), \cdots, x(T), \cdots$ be a sequence of numbers such that 
\begin{align*}
    \limsup_{T\to+\infty}\frac{x(T)}{T} < \theta_2 - \theta_1.
\end{align*}
Then we have
\begin{align*}
    \lim\inf_T\frac{\ln\left\{\mbP\left(R_{\theta, \cD}^{\pi}(T) > x(T)\right)\right\}}{\mbE[R_{\tilde\theta, \cD}^{\pi}(T)]} \geq -C
\end{align*}
holds for some $C>0$ only dependent on $\theta, \tilde\theta, \sigma$.
\end{theorem}

The proof of Theorem \ref{thm:lower-bound-fixed-time} builds upon the change of measure argument appeared in \cite{fan2021fragility} and \cite{simchi2022simple}. Our arguments generalize theirs by emphasizing a more precise dependence of the log tail probability on the regret threshold as well as the regret expectation. For the worst-case scenario, we construct a series of instance pairs such that the gap between two arms is $\Theta(x(T)/T)$. For the instance-dependent scenario, we fix the pair of instances and investigates how the tail probability scales in the two environments as $T$ increases. The following lemma is an intermediate step towards completing the proof. It shows that if the policy is ``effective", i.e., achieves sub-linear regret under either case, then the estimation of the \textit{sub-optimal} arm becomes more precise in probability as $T$ increases. In particular, for the worst-case scenario, a more delicate evaluation of the gap between the true mean and the estimated mean is needed compared to that in \cite{simchi2022simple}. Detailed proof is provided in the supplementary material. 

\begin{lemma} \label{lemma:sub-linear}
Consider the two-armed bandit problem. We have the following arguments.

1. Let $\pi$ be a fixed-time policy such that
\begin{align*}
    \limsup_{T\to+\infty}\frac{\sup_{\theta, \cD} \mathbb E\left[R_{\theta, \cD}^{\pi}(T)\right]}{T} = 0.
\end{align*}
Then we have
\begin{align*}
    \limsup_{T\to+\infty}\sup_{\tilde\theta:1/2\geq\tilde\theta_1>\tilde\theta_2}\mbP_{\tilde\theta, \cD}^{\pi(T)}(|\hat\theta_{T, 2} - \tilde\theta_2| > 2\sigma\sqrt{\ln n_{T, 2}}/\sqrt{n_{T, 2}}) = 0.
\end{align*}

2. Let $\pi$ be a fixed-time policy such that for any true mean vector $\theta$ and $\cD=(N(0, \sigma^2), N(0, \sigma^2))$, 
\begin{align*}
    \limsup_{T\to+\infty}\frac{\mathbb E\left[R_{\theta, \cD}^{\pi}(T)\right]}{T} = 0.
\end{align*}
Then for any $\tilde\theta = (\tilde\theta_1, \tilde\theta_2)$ where $\tilde\theta_1 > \tilde\theta_2$, and any $\varepsilon > 0$, we have
\begin{align*}
    \limsup_{T\to+\infty}\mbP_{\tilde\theta, \cD}^{\pi(T)}(|\hat\theta_{T, 2} - \tilde\theta_2| > \varepsilon) = 0.
\end{align*}
\end{lemma}

Theorem \ref{thm:lower-bound-fixed-time} immediately implies Proposition \ref{prop:lower-bound}, which shows the trade-off among $\alpha$-optimality, $\beta$-consistency, and $(\delta, \gamma)$-tail.

\begin{proposition} \label{prop:lower-bound}
We have the following arguments.

1. Fix $\alpha\in[1/2, 1)$. If a policy $\pi$ is $\alpha$-optimal, then for any $\delta>\alpha$ and $\gamma > \delta + \alpha - 1$, $\pi$ is not worst-case $(\delta, \gamma)$-tailed. 

2. Fix $\beta\in[0, 1)$. If a policy $\pi$ is $\beta$-consistent, then
\begin{enumerate}
    \item[(a)] if $T$ is known, then for any $\delta>\beta$ and $\gamma>\beta$, $\pi$ is not instance-dependent $(\delta, \gamma)$-tailed.
    \item[(b)] if $T$ is unknown, then for any $\delta>\beta$ and $\gamma>\delta\beta$, $\pi$ is not instance-dependent $(\delta, \gamma)$-tailed.
\end{enumerate}
\end{proposition}

Now, given $\alpha\in[1/2, 1)$, $\beta\in[0, 1)$ and the family of policies that are \textit{both} $\alpha$-optimal and $\beta$-consistent, the best regret tail we can hope is characterized in Corollary \ref{coro:lower-bound}, which is a direct application of the argument ``if a policy is \textit{not} instance-dependent $(\delta, \gamma)$-tailed, then it is also not worst-case $(\delta, \gamma)$-tailed."

\begin{corollary} \label{coro:lower-bound}
Fix $\alpha\in[1/2, 1)$ and $\beta\in[0, 1)$. If a policy $\pi$ is \textit{both} $\alpha$-optimal and $\beta$-consistent, then we have the following arguments.

1. If $\pi$ has knowledge of $T$, then
\begin{itemize}
    \item for any $\delta>\alpha$ and $\gamma > (\delta+\alpha-1)\wedge\beta$, $\pi$ cannot be worst-case $(\delta, \gamma)$-tailed.
    \item for any $\delta>\beta$ and $\gamma > \beta$, $\pi$ cannot be instance-dependent $(\delta, \gamma)$-tailed.
\end{itemize}

2. If $\pi$ has no knowledge of $T$, then
\begin{itemize}
    \item for any $\delta>\alpha$ and $\gamma > (\delta+\alpha-1)\wedge\delta\beta$, $\pi$ cannot be worst-case $(\delta, \gamma)$-tailed.
    \item for any $\delta>\beta$ and $\gamma > \delta\beta$, $\pi$ cannot be instance-dependent $(\delta, \gamma)$-tailed.
\end{itemize} 
\end{corollary}

\highlight{
We would like to emphasize that another implication from Theorem \ref{thm:lower-bound-fixed-time} is on the general instance-dependent property of any policy. If we take a non-decreasing function $f(T)$ (e.g., $\ln^2 T$) such that $\mbE[R_{\tilde\theta, \cD}^{\pi}(T)]\asymp f(T)$, we have that the tail risk
\begin{align*}
    \mbP(R_{\tilde\theta, \cD}^{\pi}(T)>x) = \exp(-O(f(T))).
\end{align*}
Moreover, if the policy is any-time (without knowing $T$ in advance), then we have
\begin{align*}
    \mbP(R_{\tilde\theta, \cD}^{\pi}(T)>x) \geq \mbP(R_{\tilde\theta, \cD}^{\pi}(\lceil x\rceil)>x) = \exp(-O(f(x))).
\end{align*}
This characterizes the best we can hope for $0$-consistent policies, particularly for those that achieve $\polylog(T)$ expected regret.

To help better understand the above results and discussions in an intuitive way, let a policy be $\alpha$-optimal and achieves a desired instance-dependent expectation growing rate of $f(T)$ ($\beta$-consistency is a special case when $f(T)=T^\beta$). For any regret threshold $x$, the critical values of log tail probability for different scenarios and cases are listed in Table \ref{table:lower-bound}. That said, the best we can hope for the order of the regret tail bounds cannot decay faster than the critical values. 
\begin{table}[!ht]
\centering
\renewcommand{\arraystretch}{1.5}
\begin{tabular}{|c|c|c|}\hline
    & known $T$ & unknown $T$  \\\hline
    \multirow{2}{*}{\shortstack{$\ln\sup_{\theta, \cD}\mbP_{\theta, \cD}^\pi(\text{Regret}>x)$ \\ (worst-case scenario)}} & \multirow{2}{*}{\shortstack{$-(x/T^{1-\alpha})\wedge f(T)$ \\ for large $x$}}  & \multirow{2}{*}{\shortstack{$-(x/T^{1-\alpha})\wedge f(x)$ \\ for large $x$}} \\
     & & \\\hline
    \multirow{2}{*}{\shortstack{$\ln\mbP_{\theta, \cD}^\pi(\text{Regret}>x)$ \\ (instance-dependent scenario)}} & \multirow{2}{*}{\shortstack{$-f(T)$ \\ for large $x$}} & \multirow{2}{*}{\shortstack{$-f(x)$ \\
    for large $x$}}  \\
     & & \\\hline
\end{tabular}
\caption{\centering Critical values of log tail probability for the family of policies that are both $\alpha$-optimal and achieve $f(T)$ instance-dependent expectation growing rate}
\label{table:lower-bound}
\end{table}
}

\section{Tail Upper Bound: The Best to Achieve}
\label{sec:upper}

In this section, we show that ``the best we can hope" is achievable by concrete policies. Without loss of generality, we can assume that $0\leq\beta\leq\alpha\leq 1$. This is because an $\alpha$-suboptimal policy is always $\alpha$-inconsistent. Meanwhile, in Table \ref{table:lower-bound}, we can observe that in the worst-case scenario, if $\beta>\alpha$, then $x^\beta$ (or $T^\beta$) is dominated by $x/T^{1-\alpha}$ (remember that $x=O(T)$). Therefore, we will ignore the case where $\beta>\alpha$. Let $f(T)$ be a non-decreasing function such that $\limsup_{T}f(T)/\ln T = +\infty$. In fact, our results match the lower bounds in Theorem \ref{thm:lower-bound-fixed-time} even if we require the instance-dependent regret expectation grows at $f(T)=o(T^\beta)$ for any $\beta>0$. Our results will achieve the desired optimal tail decaying rate shown in Table \ref{table:lower-bound} by setting $f(T)=\Theta(T^\beta)$.

\subsection{The Fixed-time Design}

\begin{theorem} \label{thm:upper-bound-fixed}
For the $K$-armed bandit problem, $\pi=\UCB$ with
\begin{align} \label{rad:optimal}
    \rad(n) = \eta\frac{(T/K)^\alpha\sqrt{\ln K}}{n} \wedge \sqrt{\frac{f(T)}{n}}
\end{align}
satisfies the following properties: for any $\eta \geq 0$ and any $x>0$, we have

1. (worst-case regret tail) {\small
\begin{align}
    & \quad\; \sup_{\theta, \cD}\mbP(R_{\theta, \cD}^\pi(T)\geq x) \nonumber\\
    & \leq K\exp\left(\frac{\left(x-K-4\eta K^{1-\alpha}T^\alpha\sqrt{\ln K}\right)_+^2}{32\sigma^2KT}\right) + K\exp\brs{-\frac{\eta(x-K)_+\sqrt{\ln K}}{2\sigma^2K^{\alpha} T^{1-\alpha}}} + KT\exp\brs{-\frac{f(T)}{2\sigma^2}}.
    \label{eq:tail-worst-case}
\end{align}}

2. (instance-dependent regret tail) {\small
\begin{align}
    & \quad\; \mbP(R_{\theta, \cD}^\pi(T)\geq x) \nonumber\\
    & \leq K\exp\brs{-\frac{\brs{(x-K)\Delta_0-8f(T)}_+}{16\sigma^2}} + \sum_{k:\Delta_k>0}\exp\brs{-\frac{\eta\Delta_kT^\alpha\sqrt{\ln K}}{\sigma^2K^{\alpha}}} + KT\exp\brs{-\frac{f(T)}{2\sigma^2}}. \label{eq:tail-instance-dependent}
\end{align}}
Here, $\Delta_0$ is such that $1/\Delta_0 = \sum_{k':\Delta_{k'}>0}1/\Delta_{k'}$.
\end{theorem}

The following corollary shows that the upper bounds in Theorem \ref{thm:upper-bound-fixed} match the lower bounds in Corollary \ref{coro:lower-bound}.
\begin{corollary} \label{coro:upper-bound-fixed}
Let $f(T)=T^\beta$. For the $K$-armed bandit problem, $\pi=\UCB$ with \eqref{rad:optimal} enjoys the following tail behavior:
\begin{itemize}
    \item for any $\delta>\alpha$, $\pi$ is $(\delta, (\delta+\alpha-1)\wedge\beta)$-tailed.
    \item for any $\delta>\beta$, $\pi$ is $(\delta, \beta)$-tailed.
\end{itemize}
\end{corollary}

The following proposition shows that the UCB policy with \eqref{rad:optimal-any} obtains $\tilde O(T^\alpha)$ worst-case regret and $O(f(T))$ instance-dependent regret.

\begin{proposition} \label{prop:upper-bound-expected}
Fix any $\alpha\in[1/2, 1)$ and non-decreasing $f(T)=\omega(\ln T)$. For the $K$-armed bandit problem, $\pi=\UCB$ with \eqref{rad:optimal} enjoys the following expected regret bounds (ignoring additive and multiplicative constant terms): for any $\eta>0$, we have
\begin{align*}
    \sup_{\theta, \cD}\mbE\left[R_{\theta, \cD}^\pi(T)\right] = O\left(K^{1-\alpha}T^\alpha\sqrt{\ln K}\right) \quad\text{and}\quad \mbE\left[R_{\theta, \cD}^\pi(T)\right] = O\left(f(T)\sum_{k:\Delta_k>0}\frac{1}{\Delta_k}\right).
\end{align*}
\end{proposition}

{\noindent\bf Remarks.} We would like to give some remarks on Theorem \ref{thm:upper-bound-fixed} and Proposition \ref{prop:upper-bound-expected}.

\begin{enumerate}
\item {\bf Phase transition.} The design of our bonus term is novel and hopefully provides additional insights, as follows. The first component can be interpreted as controlling the worst-case tail risk, while the second one can be regarded as controlling the instance-dependent tail risk. There exhibits a \textit{phase transition} with respect to the size of confidence interval. Take $f(T)=\Theta(T^\beta)$. At the beginning $\tilde\Theta(T^{2\alpha-\beta})$ time periods, the second term dominates, and so the confidence interval shrinks at a rate of $1/\sqrt{n}$, suggesting that we focus more on exploration within the consistency constraint. While in the remaining time periods, the first term dominates, and so the confidence interval shrinks at a rate of $1/n$, suggesting that we focus more on exploitation within the optimality condition. Our policy design suggests that to achieve more light-tailed risk, it might be beneficial to have two different phases in the policy design: more exploration at the beginning, and more exploitation afterwards.

\item {\bf Policy robustness.} Our tail bounds hold for any $\eta\geq 0$. As a result, in Proposition \ref{prop:upper-bound-expected}, we demonstrate that the orders of the regret bounds are independent with the hyper-parameter $\eta$ as well as the specific choice of $f(T)$. It also implies that mis-specifying the variance parameter $\sigma$ does not cost much with regard to the regret tail and expectation, \textit{as long as we are willing to sacrifice a little bit on instance-dependent consistency} --- note that we require $f(T)=\omega(\ln T)$.

\item {\bf Proof techniques.} 
The detailed proof is left to the supplementary material. We would like to emphasize the technical novelty compared to that in \cite{simchi2022simple}. In general, since $\alpha$ and $f(\cdot)$ become flexible constant and function, the proof requires more delicate formulas. When proving the worst-case upper bound, we need a careful manipulation on $\rad(n)$ since we are dealing with the minimum of two different types of bonus terms. When proving the instance-dependent upper bound, we require a careful division of the tail event to make the bound as tight as possible, depending on specific instances ($\theta$). In \cite{simchi2022simple}, the tail bound is only concerned with the worst-case scenario with $\alpha=1/2$, and hence the aforementioned challenges do not exist. Moreover, our results improve over that in \cite{simchi2022simple} --- when $\alpha=1/2$, our results reduce the dependence of the $\sqrt{\ln T}$ to a $\sqrt{\ln K}$ factor, which can be essential in practice since in many experimentation settings $T$ can be large but $K$ remains small.
\end{enumerate}

The remarks above also apply to results in the next section.

\subsection{The Any-time Design}

\begin{theorem} \label{thm:upper-bound-any}
For the $K$-armed bandit problem, $\pi=\UCB$ with
\begin{align} \label{rad:optimal-any}
    \rad_t(n) = \eta\frac{(t/K)^\alpha\sqrt{\ln K}}{n} \wedge \sqrt{\frac{f(t)}{n}}
\end{align}
satisfies the following property: fix any $0\leq\beta\leq \alpha\leq 1$ and $\eta, \eta_2>0$, for any $x > 0$, we have

1. (worst-case regret tail) {\small
\begin{align}
    & \quad\sup_{\theta, \cD}\mbP(R_{\theta, \cD}^\pi(T)\geq x) \nonumber\\
    & \leq K\exp\left(-\frac{\left(c_\alpha(x-K)-4\eta K^{1-\alpha}T^\alpha\sqrt{\ln K}\right)_+^2}{32\sigma^2KT}\right) + K\exp\brs{-c_\alpha\frac{\eta(x-K)_+\sqrt{\ln K}}{2\sigma^2K^{\alpha} T^{1-\alpha}}} + \nonumber\\
    & \quad\quad K\int_{0}^T\exp\brs{-\frac{f(x\vee y)}{2\sigma^2}}dy.
    \label{eq:tail-worst-case-any}
\end{align}}

2. (instance-dependent regret tail) {\small
\begin{align}
    & \quad\; \mbP(R_{\theta, \cD}^\pi(T)\geq x) \nonumber\\
    & \leq 2K\exp\brs{-\frac{\brs{(x-K)\Delta_0-8f(T)}_+}{256\sigma^2}} + \sum_{k:\Delta_k>0}\exp\brs{-\frac{\eta\Delta_kx^\alpha\sqrt{\ln K}}{\sigma^2K^{\alpha}}} + K\int_{0}^T\exp\brs{-\frac{f(x\vee y)}{2\sigma^2}}dy.
    \label{eq:tail-instance-dependent-any}
\end{align}}
Here, $\Delta_0$ is such that $1/\Delta_0 = \sum_{k':\Delta_{k'}>0}1/\Delta_{k'}$.
\end{theorem}

The following corollary shows that the upper bounds in Theorem \ref{thm:upper-bound-any} match the lower bounds in Corollary \ref{coro:lower-bound}.
\begin{corollary} \label{coro:upper-bound-any}
Let $f(t)=t^\beta$. For the $K$-armed bandit problem, $\pi=\UCB$ with \eqref{rad:optimal} enjoys the following tail behavior:
\begin{itemize}
    \item for any $\delta>\alpha$, $\pi$ is $(\delta, (\delta+\alpha-1)\wedge\delta\beta)$-tailed.
    \item for any $\delta>\beta$, $\pi$ is $(\delta, \delta\beta)$-tailed.
\end{itemize}
\end{corollary}

The following proposition shows that the UCB policy with \eqref{rad:optimal-any} obtains $\tilde O(T^\alpha)$ worst-case regret and $O(f(T))$ instance-dependent regret.

\begin{proposition} \label{prop:upper-bound-expected-any}
Fix any $\alpha\in[1/2, 1)$ and non-decreasing $f(t)=\omega(\ln t)$. For the $K$-armed bandit problem, $\pi=\UCB$ with \eqref{rad:optimal-any} enjoys the following expected regret bounds (ignoring additive and multiplicative constant terms): for any $\eta>0$, we have
\begin{align*}
    \sup_{\theta, \cD}\mbE\left[R_{\theta, \cD}^\pi(T)\right] = O\left(K^{1-\alpha}T^\alpha\sqrt{\ln K}\right) \quad\text{and}\quad \mbE\left[R_{\theta, \cD}^\pi(T)\right] = O\left(f(T)\sum_{k:\Delta_k>0}\frac{1}{\Delta_k}\right).
\end{align*}
\end{proposition}

{\noindent\bf Remarks.} We would like to provide some remarks on Theorem \ref{thm:upper-bound-any} and Proposition \ref{prop:upper-bound-expected-any} as follows.
\begin{enumerate}
    \item {\bf Any-time tail behavior.} Comparing Theorem \ref{thm:upper-bound-fixed} and \ref{thm:upper-bound-any} we can find that the main difference of regret tail bounds lies in the last term related to $f$. Let's assume $f(x)\geq 16\sigma^2\ln x$ for sufficiently large $x$. In the fixed-time case, the last term is
    \begin{align*}
        KT\exp\brs{-\frac{f(T)}{2\sigma^2}} = \exp\brs{-\frac{f(T) - O(\ln T)}{2\sigma^2}} = \exp\brs{-\Theta(f(T))}.
    \end{align*}
    In the any-time case, the last term is
    \begin{align*}
        K\int_{0}^T\exp\brs{-\frac{f(x\vee y)}{2\sigma^2}}dy & \leq Kx\exp\brs{-\frac{f(x)}{2\sigma^2}} + K\int_{x}^{+\infty}\exp\brs{-\frac{f(y)-8\sigma^2\ln y}{4\sigma^2} - 2\ln y}dy \\ 
        & = Kx\exp\brs{-\frac{f(x)}{2\sigma^2}} + K\exp\brs{-\frac{f(x)}{8\sigma^2}}\int_{x}^{+\infty}y^{-2}dy \\
        & = \exp\brs{-\Theta(f(x))}.
    \end{align*}
    This resonates with the intrinsic difference between known $T$ and unknown $T$ revealed in Table \ref{table:lower-bound}.

    \item {\bf Assumptions on $f(\cdot)$.} In Theorem \ref{thm:upper-bound-any} we only consider the case when $f(T)=\omega(\ln T)$. The case when $f(T)=\Theta(\ln T)$ is addressed in the MAB literature. Specifically, standard results (see., e.g., \citealt{auer2002finite}) showed that by setting $\eta=+\infty$ and $f(T)=\Theta(\ln T)$ appropriately, the policy achieves $\tilde O(\sqrt T)$ worst-case expected regret and $O(\ln T)$ instance-dependent expected regret. However, the policy becomes heavy-tailed --- the probability of incurring a linear regret must be polynomially dependent on $1/T$ (see also Theorem 1 in \citealt{simchi2022simple}). Therefore, we believe our results are general enough to characterize optimal regret tail bounds for various $\alpha$ and $f(\cdot)$.
\end{enumerate}

\begin{figure}[htbp]
    \centering
    \includegraphics[width=0.95\linewidth]{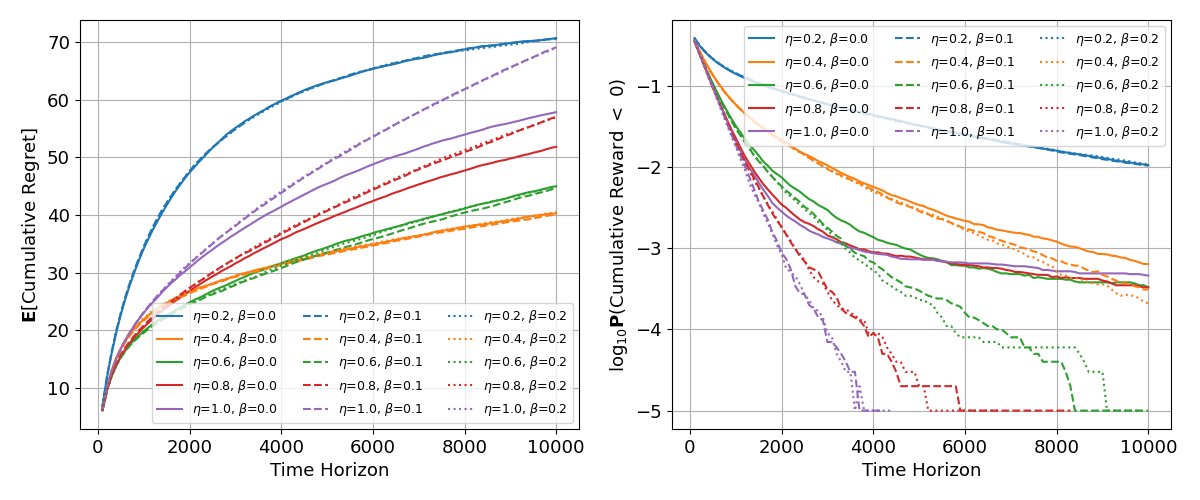}
    \caption{\centering regret expectation vs. tail risk for $(\mathcal N(0.1, 1), \mathcal N(-0.1, 1))$}
    \label{fig:2armed-any}
\end{figure}

We conduct a simulation study to validate the practical performance of our any-time policies (UCB with \eqref{rad:optimal-any}). We consider a $2$-armed bandit environment with $\theta=(0.1, -0.1)$ and $\cD = (\cN(0, 1), \cN(0, 1))$. We choose $\alpha=1/2$, $\eta\in\{0.2, 0.4, 0.6, 0.8, 1.0\}\times\sqrt{\log_2e}$ and $f(T)=T^\beta\log_2 T$ with $\beta\in\{0, 0.1, 0.2\}$ (essentially we are using the base-$2$ logarithm). Each policy is run with the maximum horizon $T=10^4$ and the number of sample paths $N=10^5$. We show the empirical cumulative regret expectation as well as the empirical tail probability of negative cumulative reward (which is a special case of incurring a \textit{linear} regret) in Figure \ref{fig:2armed-any}. One can regard $\eta=1$ and $\beta=0$ as a benchmark case when the UCB bonus term is the standard one \citep{auer2002finite, garivier2011kl, ashutosh2021bandit}.

We would like to point out some observations that are potentially useful in practice as follows:
\begin{itemize}
    \item \textit{A little sacrifice on instance-dependent consistency significantly improves safety.} This phenomenon happens especially when $\eta\in\{0.6, 0.8, 1.0\}$ is not too small --- as we increase $\beta$ from $0$ to $0.1$, we do observe a small increase in regret expectation, but the tail probability decays much faster when $t$ grows. In particular, when $\eta=0.6$, there is no empirical sacrifice in efficiency up to $T=10000$ but a big improvement on safety!
    
    \item \textit{Controlling the worst-case tail using the $1/n$ bonus.} When $\beta$ increases from $0.1$ to $0.2$, we observe little loss in efficiency (and no significant gain in safety). This is because, in \eqref{rad:optimal-any}, we take the minimum of two bonus terms. As $\beta$ increases, the first term --- designed to control the worst-case tail --- begins to dominate. This ensures that efficiency is preserved while tail risk remains under control. Without this term, achieving the same level of safety could result in unnecessary loss of efficiency.
\end{itemize}

\section{Generalization and Extensions} \label{sec:extension}

\subsection{Robustness in Sub-Exponential Environments} \label{ssec:extension:baseline}

In this section, we extend our results in Section \ref{sec:upper} to environments with sub-exponential noise. We will show that through a simple amendment to our bonus design, we are able to achieve the same regret tail decaying rate as in Section \ref{sec:upper}. Along showing the regret tail bounds, we will also highlight the robustness of our design --- our policies achieve desired trade-off between expectation and tail risk even under missing information of the environment profiles such as volatility parameters.

We now briefly restate the setting and highlight those different from the standard stochastic MAB model introduced in Section \ref{sec:setup}. Denote $T$ as the time horizon and $K$ as the number of different arms. Without loss of generality,  we presume that $K\geq 2$, $T\ge 3$, and $T\geq K$. At each time $t$, upon pulling an arm $a_t$, a reward is independently sampled as $r_{t, a_t} = \theta_{a_t} + \epsilon_{t, a_t}$, where $\theta_{a_t}$ is the mean reward of arm $a_t$, and $\epsilon_{t, a_t}$ is a zero-mean random noise, independent across time periods. The random noise $\epsilon_{t, a_t}$ is assumed to be $(\sigma, \nu)$-sub-exponential: for any arm $k$ and time $t$, 
\begin{align*}
    \max\left\{\mbP\left(\epsilon_{t, k} \geq x\right), \mbP\left(\epsilon_{t, k} \leq -x\right)\right\} \leq \exp\brs{-\frac{x^2}{2\sigma^2}\wedge \frac{x}{2\nu}}.
\end{align*}

\begin{theorem} \label{thm:upper-bound-fixed-subexp}
For the $K$-armed bandit problem, $\pi=\UCB$ with
\begin{align} \label{rad:optimal-subexp}
    \rad(n) = \eta\frac{(T/K)^\alpha\sqrt{\ln K}}{n} \wedge \brs{\sqrt{\frac{f(T)}{n}}\vee\kappa\frac{f(T)}{n}}
\end{align}
satisfies the following properties: for any $\eta, \kappa \geq 0$ and any $x>0$, we have

1. (worst-case regret tail) {\small
\begin{align}
    & \quad\; \sup_{\theta, \cD}\mbP(R_{\theta, \cD}^\pi(T)\geq x) \nonumber\\
    & \leq K\exp\left(\frac{\left(x-K-4\eta K^{1-\alpha}T^\alpha\sqrt{\ln K}\right)_+^2}{(32\sigma^2\vee 4\nu)KT}\right) + K\exp\brs{-\frac{\eta x\sqrt{\ln K}}{(2\sigma^2\vee\nu)K^{\alpha} T^{1-\alpha}}} + KT\exp\brs{-\frac{f(T)}{2\sigma^2\vee 2\nu\kappa^{-1}}}.
    \label{eq:tail-worst-case-subexp}
\end{align}}

2. (instance-dependent regret tail) {\small
\begin{align}
    & \quad\; \mbP(R_{\theta, \cD}^\pi(T)\geq x) \nonumber\\
    & \leq K\exp\brs{-\frac{\brs{(x-K)\Delta_0-(8\vee\kappa^2)f(T)}_+}{16\sigma^2\vee 8\nu}} + \sum_{k:\Delta_k>0}\exp\brs{-\frac{\eta\Delta_k(T/K)^\alpha\sqrt{\ln K}}{\sigma^2\vee\nu}} + \nonumber\\
    & \quad\quad KT\exp\brs{-\frac{f(T)}{2\sigma^2\vee 2\nu\kappa^{-1}}}. \label{eq:tail-instance-dependent-subexp}
\end{align}}
Here, $\Delta_0$ is such that $1/\Delta_0 = \sum_{k':\Delta_{k'}>0}1/\Delta_{k'}$.
\end{theorem}

\begin{theorem} \label{thm:upper-bound-any-subexp}
For the $K$-armed bandit problem, $\pi=\UCB$ with
\begin{align} \label{rad:optimal-any-subexp}
    \rad_t(n) = \eta\frac{(t/K)^\alpha\sqrt{\ln K}}{n} \wedge \brs{\sqrt{\frac{f(t)}{n}}\vee\kappa\frac{f(t)}{n}}
\end{align}
satisfies the following property: fix any $\eta, \kappa\geq 0$, and any $x > 0$, we have

1. (worst-case regret tail) {\small
\begin{align}
    & \quad\sup_{\theta, \cD}\mbP(R_{\theta, \cD}^\pi(T)\geq x) \nonumber\\
    & \leq K\exp\left(-\frac{\left(c_{\alpha}(x-K)-4\eta K^{1-\alpha}T^\alpha\sqrt{\ln K}\right)_+^2}{(32\sigma^2\vee 8\nu)KT}\right) + K\exp\brs{-\frac{c_\alpha\eta x\sqrt{\ln K}}{(\sigma^2\vee\nu)K^{\alpha} T^{1-\alpha}}} + \nonumber\\
    & \quad\quad K\int_{0}^T\exp\brs{-\frac{f(x\vee y)}{2\sigma^2\vee 2\nu\kappa^{-1}}}dy.
    \label{eq:tail-worst-case-any-subexp}
\end{align}}

2. (instance-dependent regret tail) {\small
\begin{align}
    & \quad\; \mbP(R_{\theta, \cD}^\pi(T)\geq x) \nonumber\\
    & \leq 2K\exp\brs{-\frac{\brs{(x-K)\Delta_0-(8\vee\kappa^2)f(T)}_+}{256\sigma^2\vee 32\nu}} + \sum_{k:\Delta_k>0}\exp\brs{-\frac{\eta\Delta_k(x/K)^\alpha\sqrt{\ln K}}{\sigma^2\vee\nu}} + \nonumber\\
    & \quad\quad K\int_{0}^T\exp\brs{-\frac{f(x\vee y)}{2\sigma^2\vee 2\nu\kappa^{-1}}}dy.
    \label{eq:tail-instance-dependent-any-subexp}
\end{align}}
Here, $\Delta_0$ is such that $1/\Delta_0 = \sum_{k':\Delta_{k'}>0}1/\Delta_{k'}$.
\end{theorem}

\subsection{Extension to Linear Bandits} \label{sec:extension:linear}
In this section, we further extend our policy design to the setting of linear bandits. We briefly review the setting of linear bandits as follows (see, e.g., \citealt{dani2008stochastic}, \citealt{abbasi2011improved}, for reference of more details). In each time period $t$, the decision maker (DM) is given an action set $\cA_t\subseteq \mathbb R^d$ from which the DM needs to select one action $a_t\in\cA_t$ to take for the time period $t$. Subsequently a reward of $r_t = \theta^{\top}a_t + \epsilon_{t, a_t}$ is collected, where $\theta\in\mathbb R^d$ is an unknown parameter and $\epsilon_{t, a_t}$ is an independent $\sigma$-sub-gaussian mean-zero noise. More specifically, let $H_t = \{a_1, r_{1, a_1}, \cdots, a_{t-1}, r_{t-1, a_{t-1}}\}$ be the history prior to time $t$. When $t=1$, $H_1=\emptyset$. At time $t$, the DM adopts a policy $\pi_t: H_t\longmapsto a_t$ that maps the history $H_t$ to an action $a_t$, where $a_t$ may be realized from some probability distribution on $\cA_t$. Adopting the standard assumptions in the linear bandits literature, we presume that $\|\theta\|_{\infty} \leq 1$ and $\|a\|_2 \leq 1$ for any $a\in\cA_t$ and any $t$. Let $a_t^*=\arg\max_{a\in\cA_t}\theta^\top a$. We assume $\theta^\top (a_t^* - a)\leq 1$ for any $a\in\cA_t$. The empirical regret is defined as
\begin{align*}
    \hat R_{\theta, \cD}^\pi(T) = \sum_{t=1}^T \theta^\top a_t^{*} - \sum_{t=1}^T r_{t, a_t} = \sum_{t=1}^T\theta^\top (a_t^* - a_t) - \sum_{t=1}^T\epsilon_{t, a_t} \triangleq R_{\theta, \cD}^\pi(T) - N^\pi(T).
\end{align*}
In the instance-dependent scenario, we define $\Delta \geq 0$ as the uniform lower bound of the gap between the optimal reward and the second-optimal reward across all time periods. That is, 
\begin{align*}
    \Delta = \inf_{t}\inf_{a\neq a_t^*}\theta^\top a_t^* - \theta^\top a.
\end{align*}
Same as in the MAB setting, $N^\pi(T)$ also enjoys the fast concentration property in Lemma \ref{lemma:bound-noise}.

We provide the Linear UCB policy ($\UCB\text{\rm -L}$, adapted from \citealt{abbasi2011improved} and \citealt{simchi2022simple}) in Algorithm \ref{alg:UCB-lin}. The following theorems show that under carefully specified bonus terms $\rad_t(z)$, we can obtain explicit exponentially decaying regret tail bounds similar to that in the standard stochastic MAB setting, for both the fixed-time case (Theorem \ref{thm:linear}) and the any-time case (Theorem \ref{thm:linear-any}). Note that in standard bonus design, $\rad_t(z)\propto \sqrt{z}$ (see, e.g., the OFUL policy in \citealt{abbasi2011improved}). In our design, $z=a_t^\top V_{t-1}^{-1}a_t$ (denote as $\|a_t\|_t^2$) is regarded as a counterpart of $1/n_i$ in the MAB setting (though they are not equivalent). An additional $\sqrt{dz}$ term is introduced to force exploration at the very beginning and prevent the policy from sticking to a suboptimal action.

\bigskip
\begin{breakablealgorithm}
\caption{UCB-Linear (UCB-L)}
\label{alg:UCB-lin}
\begin{algorithmic}[1]
\State $t\gets 0$, $V_0=I$, $\hat\theta_0=0$.
\While{$t < T$}
    \State $t\gets t+1$. Observe $\cA_t$.
    \State Take the action with the highest $\UCB$:
    \begin{align*}
        a_t=\arg\max_{a\in\cA_t}\left\{\hat\theta_{t-1}^\top a  + \rad_t(a^\top V_{t-1}^{-1}a)\right\}.
    \end{align*}
    \State $V_t=V_{t-1} + a_ta_t^\top$, $\hat\theta_t = V_t^{-1}(\sum_{s\leq t}a_sr_s)$.
\EndWhile
\end{algorithmic}
\end{breakablealgorithm}
\bigskip

\begin{theorem} \label{thm:linear}
Let $T\geq d$. $\pi=\UCB\text{\rm -L}$ with
\begin{align} \label{rad:linear}
    \rad(z) = \eta (T/d)^\alpha \sqrt{d}z\wedge\sqrt{f(T)z}+\sqrt{dz}
\end{align}
satisfies the following property: for any $\alpha\in[1/2, 1)$, $f(T)=\omega(\ln T)$ non-decreasing, $\eta > 0$, $x>0$, we have

1. (worst-case regret tail) {\small
\begin{align*}
    \sup_{\theta, \cD}\mbP(R_{\theta, \cD}^\pi(T)\geq x) & \leq 2d(T/d)^{2d+1}\exp\left(-\frac{\left(x-1-16d\sqrt{T}\ln T - 8\eta d^{\frac{3}{2}-\alpha}T^\alpha\ln T\right)_+^2}{128\sigma^2dT\ln^2T}\right) \\
    & \quad\quad + 2d(T/d)^{2d+1}\exp\left(-\frac{\eta(x-1)_+}{4\sigma^2 d^{\alpha-\frac{1}{2}} T^{1-\alpha}\ln T}\wedge\frac{f(T)}{2\sigma^2}\right).
\end{align*}}

2. (instance-dependent regret tail) {\small
\begin{align*}
    \sup_{\theta, \cD}\mbP(R_{\theta, \cD}^\pi(T)\geq x) & \leq 2d(T/d)^{2d+1}\exp\left(-\frac{\left(\Delta(x-1)/4 - 128d - 32f(T)\right)_+}{32\sigma^2d\ln T}\right) \\
    & \quad\quad + 2d(T/d)^{2d+1}\exp\left(-\frac{\eta\Delta T^\alpha}{2\sigma^2 d^{\alpha-\frac{1}{2}}}\wedge\frac{f(T)}{2\sigma^2}\right).
\end{align*}}
\end{theorem}

\begin{theorem} \label{thm:linear-any}
Let $T\geq d$. $\pi=\UCB\text{\rm -L}$ with
\begin{align} \label{rad:linear-any}
    \rad_t(z) = \eta(t/d)^\alpha \sqrt{d}z\wedge\sqrt{f(t)z}+\sqrt{dz}
\end{align}
satisfies the following property: for any $\alpha\in[1/2, 1)$, $f(T)=\omega(\ln T)$ non-decreasing, $\eta > 0$, $x>0$, we have

1. (worst-case regret tail) {\small
\begin{align*}
    \sup_{\theta, \cD}\mbP(R_{\theta, \cD}^\pi(T)\geq x) & \leq 2d(T/d)^{2d+1}\exp\left(-\frac{\left(x-1-16d\sqrt{T}\ln T - 8\eta d^{\frac{3}{2}-\alpha}T^\alpha\ln T\right)_+^2}{128\sigma^2dT\ln^2T}\right) \\
    & \quad\quad + 2d(T/d)^{2d+1}\exp\left(-\frac{\eta(x-1)_+}{4\sigma^2 d^{\alpha-\frac{1}{2}}T^{1-\alpha}\ln T}\right) + 2(T/d)^{2d}\int_{0}^T\exp\brs{-\frac{f(x\vee y)}{2\sigma^2}}dy.
\end{align*}}

2. (instance-dependent regret tail) {\small
\begin{align*}
    \sup_{\theta, \cD}\mbP(R_{\theta, \cD}^\pi(T)\geq x) & \leq 2d(T/d)^{2d+1}\exp\left(-\frac{\left(\Delta(x-1)/4 - 128d - 32f(T)\right)_+}{32\sigma^2d\ln T}\right) \\
    & \quad\quad + 2d(T/d)^{2d+1}\exp\left(-\frac{\eta\Delta x^\alpha}{2\sigma^2 d^{\alpha-\frac{1}{2}}}\right) + 2(T/d)^{2d}\int_{0}^T\exp\brs{-\frac{f(x\vee y)}{2\sigma^2}}dy.
\end{align*}}
\end{theorem}

Finally, we would like to remark on the the expected regret of the proposed policies, as is shown in Proposition \ref{prop:upper-bound-expected-linear}. In particular, our policies yield optimal worst-case expected regret on \textit{both} $T$ and $d$ up to logarithmic factors, improving over those from \cite{simchi2022simple}. The case when $f(T)=\Theta(\ln T)$ is well addressed in the stochastic bandit literature (see., e.g., \citealt{abbasi2011improved}).

\begin{proposition} \label{prop:upper-bound-expected-linear}
$\pi=\UCB\text{\rm -L}$ with \eqref{rad:linear} or \eqref{rad:linear-any} has the following expected regret bounds (ignoring additive and multiplicative constant terms):
\begin{align*}
    \sup_{\theta, \cD}\mbE[R_{\theta, \cD}^\pi(T)] = O\left(d^{\frac{3}{2}-\alpha}T^{\alpha}(\ln T)^2\right) \quad\text{and}\quad \mbE[R_{\theta, \cD}^\pi(T)] = O\left((f(T)\vee d^2\ln^2 T)\Delta^{-1}\right).
\end{align*}
\end{proposition}

\subsection{Implications on Reinforcement Learning: AlphaGo}
\label{sec:tradeoff:alphago}

Upon concluding this section, we highlight a surprising and interesting coincidence between our any-time design \eqref{rad:optimal-any} and the Monte Carlo Tree Search (MCTS) algorithm employed in AlphaGo --- one of the most successful large-scale reinforcement learning systems, which achieved superhuman performance in the game of Go.

\begin{figure}[htbp]
    \centering
    \includegraphics[width=0.95\linewidth]{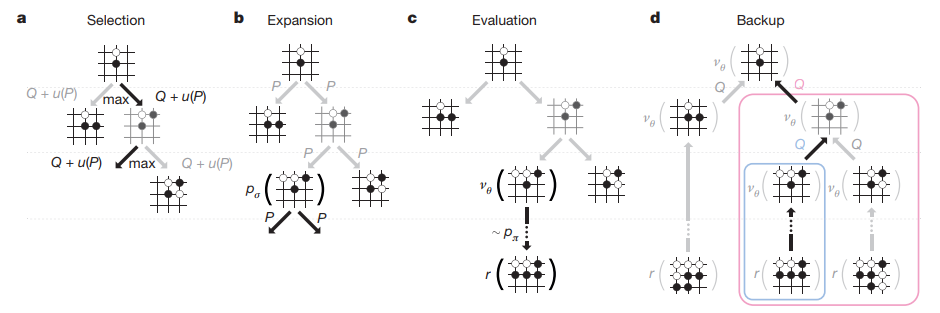}
    \caption{\centering MCTS procedure of one simulation in AlphaGo}
    \label{fig:mcts}
\end{figure}

AlphaGo’s remarkable success is built on two key pillars: MCTS and deep neural networks, which together address challenges from enormous number of possible board states. During training, whenever a state is encountered, and before taking a real action, AlphaGo performs numerous simulations --- executing “virtual” actions --- using a tree search. Each simulation consists of several phases: action selection (within the tree), tree expansion (upon reaching a leaf node), value evaluation (to estimate a reward from the simulation), and value backup (to propagate the reward and update statistics throughout the tree). An illustration of this process is shown in Figure \ref{fig:mcts} (adapted from Page 3 of \citealt{silver2016mastering}).

Admittedly, the concrete procedure can be quite complex. Here, we focus exclusively on the \textit{action selection} phase—that is, assuming the search tree is already available (possibly updated and expanded from previous simulations), we aim to determine the “virtual” action to take during simulations. Although this task is not explicitly formulated as a regret minimization problem, it inherently involves an exploration-exploitation trade-off (see, e.g., \citealt{kocsis2006bandit, browne2012survey, james2017analysis, swiechowski2023monte}):
(i) virtual actions should be sufficiently diverse to improve estimation by exploring a broad range of (state, action) pairs;
(ii) virtual actions should leverage the value estimates and statistics stored in the tree to exploit the information gathered so far.

This trade-off has been extensively studied in the MCTS literature, and it is widely accepted that action selection can be effectively guided by upper confidence bounds \citep{kocsis2006bandit, gelly2006exploration, browne2012survey, swiechowski2023monte} --- a principle that is also adopted in AlphaGo. The critical challenge lies in the design of the bonus term $u(s,a)$. Unlike in classical multi-armed bandit settings, where there is only a single state, reinforcement learning involves a variety of states, and thus the bonus must depend on each specific (state, action) pair.

Prior to AlphaGo, many works follow the standard UCB design, where $\rad_t(n)\asymp 1/\sqrt{n}$. AlphaGo introduced a different strategy by setting the exploration bonus as
\begin{align} \label{alphago}
    u(s, a) = c\cdot P(s, a)\frac{\sqrt{\sum_b n(s, b)}}{1+n(s, a)}.
\end{align}
Here, $c$ is a universal hyperparameter, and $P(s,a)$ is the so-called ``prior probability'', obtained from a policy neural network that predicts the probability of taking action $a$ under state $s$. This prior remains fixed over many simulations and is updated in a batch-wise fashion. The purpose of this design is to modulate the level of exploration for different (state, action) pairs, guided by the policy network, thereby improving exploration efficiency compared to using a uniform exploration coefficient.

In the denominator, $n(s,a)$ denotes the number of times the (state, action) pair $(s,a)$ has been visited. Interestingly, if we remove the dependence on $s$, effectively reducing the problem to a standard multi-armed bandit setting, the bonus term simplifies to approximately $\sqrt{t}/n$. This shares \underline{\textit{exactly the same order of dependence on $t$ and $n$}} as the exploration term $\rad_t(n)$ in \eqref{rad:optimal-any} if we take $\eta_2=+\infty$ (or $\rad_t(n)$ in \eqref{rad:optimal-any} up to logarithmic factors)! Since AlphaGo, this modified exploration design has been adopted in many follow-up works applying MCTS to domains such as game playing \citep{silver2017mastering, schrittwieser2020mastering} and large language model reasoning \citep{xie2024monte, luo2024improve}.

Below, we provide some theoretical insights into this surprising connection and explain why AlphaGo’s engineered solution in \eqref{alphago} proves to be successful.

\begin{enumerate}
    \item \textbf{Small instance-dependent gaps call for worst-case efficiency and safety.} The game of Go represents a highly complex Markov Decision Process with an enormous state space and many available actions. In such settings, the instance-dependent gap between the optimal and sub-optimal actions can be small (as we can observe from the definition of $\Delta$ in Section \ref{sec:extension:linear}) and may vary across different states. This scenario highlights the value of worst-case analysis. Our theoretical results (in particular, Theorem \ref{thm:upper-bound-any}) show that our new design achieves both desired efficiency and optimal safety under the desired efficiency. Specifically, it ensures the correct asymptotic performance in expectation while also providing the highest possible level of reliability, offering a substantial improvement over standard designs.
    
    \item \textbf{Mis-specified priors necessitate a robust design.} In AlphaGo’s design \eqref{alphago}, the adaptive coefficient $P(s, a)$ is set using the output of a policy neural network. These priors are critical for guiding exploration efficiently across different (state, action) pairs. However, neural networks --- though powerful --- can produce inaccurate outputs, especially early in training when they have not yet converged. Our theoretical analysis (see the remark following Theorem \ref{thm:upper-bound-fixed}) indicates that the new design maintains robust performance even under mis-specified volatilities or poorly calibrated priors. This means the algorithm remains reliable and does not suffer from severe performance degradation when a $P(s, a)$ is underestimated.
\end{enumerate}

We hope that these insights not only shed light on the effectiveness of AlphaGo’s design but also offer guidance for designing safe and robust exploration-exploitation strategies in complex decision-making environments.

\section{Conclusion}

\label{sec:conclusion}

In this work, we study the optimal regret tail bound in the stochastic multi-armed bandit problem. We fully characterize the trade-off between regret expectation and tail risk, showing how relaxing the expected regret order can help make the tail probability lighter for incurring large regret. We differentiate between the worst-case scenario and the instance-dependent scenario, and between knowing the whole time horizon in advance or not. Surprisingly, we find that an intrinsic gap of the tail rate appears under the instance-dependent scenario, but disappears under the worst-case one. Our policy design builds upon the confidence bound type policies, while build new bonus terms that reach a delicate balance between worst-case optimality, instance-dependent consistency, and light-tailed risk. We also extend our policies to settings where the standard stochastic MAB problem is combined with structured non-stationarity, including the stochastic MAB problem with non-stationary baseline rewards and the stochastic linear bandit problem.

There are several prospective future directions. Technically, one direction is to improve our policy design for linear bandits on the tail bound. Another direction is to design an any-time policy for the stochastic MAB problem with baseline rewards so that the optimal tail probabilities in the standard stochastic MAB problem are inherited. Empirically, it would be interesting to see how the policy design works in various practical settings. Methodologically, it is tempting to see whether our policy design can be integrated into more complex settings such as reinforcement learning. We hope our results and analysis in this paper may shed new lights on understanding and alleviating the tail risk of learning algorithms under a stochastic environment with different goals (worst-case v.s. instance-dependent) and different prior information (known $T$ v.s. unknown $T$).

\bibliographystyle{informs2014}
\bibliography{main}

\newpage
\setcounter{page}{1}
\begin{APPENDICES} \small
\section{Proofs for Section \ref{sec:lower}}

{\noindent\bf Proof of Lemma \ref{lemma:sub-linear}.} 

1. Define
\begin{align*}
    E_T = \left\{|\hat\theta_{T, 2} - \tilde\theta_2|\leq 2\sigma\sqrt{\ln n_{T, 2}}/\sqrt{n_{T, 2}}\right\}.
\end{align*}
Fix any positive integer $N$, we have
\begin{align*}
& \quad\ \mathbb P_{\tilde\theta}^{\pi(T)}(\bar E_T) \\
& = \mathbb P_{\tilde\theta}^{\pi(T)}(\bar E_T; n_{T, 2} < N) +  \mathbb P_{\tilde\theta}^{\pi(T)}(\bar E_T; n_{T, 2} \geq N) \\
& \leq \mathbb P_{\tilde\theta}^{\pi(T)}(n_{T, 2} < N) + \sum_{n=N}^{+\infty} \mathbb P_{\tilde\theta}^{\pi(T)}(\bar E_T; n_{T, 2} = n) \\
& \leq \mathbb P_{\tilde\theta}^{\pi(T)}(n_{T, 2} < N) + \sum_{n=N}^{+\infty}2\exp(-2\ln n).
\end{align*}
Thus,
\begin{align*}
& \quad \limsup_{T}\sup_{\tilde\theta:1/2\geq\tilde\theta_1>\tilde\theta_2}\mathbb P_{\tilde\theta}^{\pi(T)}(\bar E_T) \\
& \leq \limsup_{T}\sup_{\tilde\theta:1/2\geq\tilde\theta_1>\tilde\theta_2}\mathbb P_{\tilde\theta}^{\pi(T)}(n_{T, 2} < N) + \sum_{n=N}^{+\infty}2n^{-2}
\end{align*}
holds for any $N$. Note that the last term converges to $0$ as $N\to+\infty$. It suffices to show 
\begin{align*}
    \sup_{\tilde\theta:1/2\geq\tilde\theta_1>\tilde\theta_2}\mathbb P_{\tilde\theta}^{\pi(T)}(n_{T, 2} < N)\to 0
\end{align*}
as $T\to+\infty$ for any fixed $N$. Suppose this does not hold, then we can find $p > 0$, a sequence of times $\left\{T(m)\right\}_{m=1}^{+\infty}$ and a sequence of vectors $\{\tilde\theta(m)\}_{m=1}^{+\infty}$ (with $1/2\geq\tilde\theta(m)_1 > \tilde\theta(m)_2$) such that
$$
\mathbb P_{\tilde\theta(m)}^{\pi_{T(m)}}(n_{T(m), 2} < N)>p.
$$
Let $M$ be some large number such that $q\triangleq p - N\exp(-\frac{M^2}{2\sigma^2}) > 0$. For each $m$, consider an alternative environment $\theta(m) = (\theta(m)_1, \theta(m)_2)$ where $\theta(m)_2 > \theta(m)_1 = \tilde\theta(m)_1$. Using the change of measure argument, we have
\begin{align*}
& \quad\ \mathbb P_{\theta(m)}^{\pi_{T(m)}}(n_{T(m), 2} < N) \\
& = \mathbb E_{\theta(m)}^{\pi_{T(m)}}[\mathds 1\{n_{T(m), 2} < N\}] \\
& = \mbE_{\tilde\theta(m)}^{\pi_{T(m)}}\left[\exp\left(\sum_{n=1}^{n_{T(m), 2}}\frac{(r_{t_2(n), 2} - \tilde\theta(m)_2)^2 - (r_{t_2(n), 2} - \theta(m)_2)^2}{2\sigma^2}\right)\mathds 1\{n_{T(m), 2} < N\}\right] \\
& = \mathbb E_{\tilde\theta(m)}^{\pi_{T(m)}}\left[\exp\left(n_{T(m), 2}\left(\frac{\tilde\theta(m)_2^2 - \theta(m)_2^2}{2\sigma^2} + \frac{(\theta(m)_2 - \tilde\theta(m)_2)\hat\theta_{T(m), 2}}{\sigma^2}\right)\right)\mathds 1\{n_{T(m), 2} < N\}\right] \\
& \geq \mathbb E_{\tilde\theta(m)}^{\pi_{T(m)}}\left[\exp\left(n_{T(m), 2}\left(\frac{\tilde\theta(m)_2^2 - \theta(m)_2^2}{2\sigma^2} + \frac{(\theta(m)_2 - \tilde\theta(m)_2)\hat\theta_{T(m), 2}}{\sigma^2}\right)\right) \right.\\
& \quad\quad\quad\quad\quad \left.\mathds 1\{\hat\theta_{T(m), 2} > \tilde\theta(m)_2 - M, n_{T(m), 2} < N\}\right] \\
& \geq \mathbb E_{\tilde\theta(m)}^{\pi_{T(m)}}\left[\exp\left(N\left(-\frac{(\tilde\theta(m)_2 - \theta(m)_2)^2}{2\sigma^2} - \frac{M(\theta(m)_2 - \tilde\theta(m)_2)}{\sigma^2}\right)\right)\right. \\
& \quad\quad\quad\quad\quad \left.\mathds 1\{\hat\theta_{T(m), 2} > \tilde\theta_2 - M, n_{T(m), 2} < N\}\right] \\
& = \exp\left(N\left(-\frac{(\tilde\theta(m)_2 - \theta(m)_2)^2}{2\sigma^2} - \frac{M(\theta(m)_2 - \tilde\theta(m)_2)}{\sigma^2}\right)\right)\mathbb P_{\tilde\theta}^{\pi_{T(m)}}(\hat\theta_{T(m), 2} > \tilde\theta(m)_2 - M, n_{T(m), 2} < N) \\
& \geq \exp\left(N\left(-\frac{1}{2\sigma^2} - \frac{M}{\sigma^2}\right)\right)\mathbb P_{\tilde\theta}^{\pi_{T(m)}}(\hat\theta_{T(m), 2} > \tilde\theta(m)_2 - M, n_{T(m), 2} < N).
\end{align*}
Note that
\begin{align*}
& \quad\ \mathbb P_{\tilde\theta}^{\pi_{T(m)}}(\hat\theta_{T(m), 2} > \tilde\theta(m)_2 - M, n_{T(m), 2} < N) \\
& > p - \sum_{n=1}^{N-1}\mathbb P_{\tilde\theta}^{\pi_{T(m)}}(\hat\theta_{T(m), 2} \leq \tilde\theta(m)_2 - M, n_{T(m), 2} = n) \\
& \geq p - \sum_{n=1}^{N-1}\exp(-\frac{nM^2}{2\sigma^2}) \geq p - N\exp(-\frac{M^2}{2\sigma^2})=q>0.
\end{align*}
Therefore, there exists a constant positive probability such that $\pi_{T(m)}$ pulls arm $2$ no more than $N$ times under $\theta(m)$ for any $m$. As a result, $\{\pi(T)\}$ incurs a worst-case linear expected regret, leading to a contradiction.

2. Define
\begin{align*}
    E_T = \left\{|\hat\theta_{T, 2} - \tilde\theta_2|\leq \varepsilon\right\}.
\end{align*}
Fix any positive integer $N$, we have
\begin{align*}
& \quad\ \mathbb P_{\tilde\theta}^{\pi(T)}(\bar E_T) \\
& = \mathbb P_{\tilde\theta}^{\pi(T)}(\bar E_T; n_{T, 2} < N) +  \mathbb P_{\tilde\theta}^{\pi(T)}(\bar E_T; n_{T, 2} \geq N) \\
& \leq \mathbb P_{\tilde\theta}^{\pi(T)}(n_{T, 2} < N) + \sum_{n=N}^{+\infty} \mathbb P_{\tilde\theta}^{\pi(T)}(\bar E_T; n_{T, 2} = n) \\
& \leq \mathbb P_{\tilde\theta}^{\pi(T)}(n_{T, 2} < N) + \sum_{n=N}^{+\infty}2\exp(-\frac{n\varepsilon^2}{2\sigma^2}).
\end{align*}
Thus,
\begin{align*}
& \quad\ \limsup_{T}\mathbb P_{\tilde\theta}^{\pi(T)}(\bar E_T) \\
& \leq \limsup_{T}\mathbb P_{\tilde\theta}^{\pi(T)}(n_{T, 2} < N) + \sum_{n=N}^{+\infty}2\exp(-\frac{n\varepsilon^2}{2\sigma^2})
\end{align*}
holds for any $N$. Note that the last term converges to $0$ as $N\to+\infty$. It suffices to show $\mathbb P_{\tilde\theta}^{\pi(T)}(n_{T, 2} < N)\to 0$ as $T\to+\infty$ for any fixed $N$. Suppose this does not hold, then we can find $p > 0$ and a sequence $\left\{T(m)\right\}_{m=1}^{+\infty}$ such that
$$
\mathbb P_{\tilde\theta}^{\pi_{T(m)}}(n_{T(m), 2} < N)>p.
$$
Let $M$ be some large number such that $q\triangleq p - N\exp(-\frac{M^2}{2\sigma^2}) > 0$. Consider an alternative environment $\theta = (\theta_1, \theta_2)$ where $\theta_2 > \theta_1 = \tilde\theta_1$. Using the change of measure argument, we have
\begin{align*}
& \quad\ \mathbb P_{\theta, \cD}^{\pi_{T(m)}}(n_{T(m), 2} < N) \\
& = \mathbb E_{\theta, \cD}^{\pi_{T(m)}}[1\{n_{T(m), 2} < N\}] \\
& = \mbE_{\tilde\theta}^{\pi_{T(m)}}\left[\exp\left(\sum_{n=1}^{n_{T(m), 2}}\frac{(r_{t_2(n), 2} - \tilde\theta_2)^2 - (r_{t_2(n), 2} - \theta_2)^2}{2\sigma^2}\right)\mathds 1\{n_{T(m), 2} < N\}\right] \\
& = \mathbb E_{\tilde\theta}^{\pi_{T(m)}}\left[\exp\left(n_{T(m), 2}\left(\frac{\tilde\theta_2^2 - \theta_2^2}{2\sigma^2} + \frac{(\theta_2 - \tilde\theta_2)\hat\theta_{T(m), 2}}{\sigma^2}\right)\right)\mathds 1\{n_{T(m), 2} < N\}\right] \\
& \geq \mathbb E_{\tilde\theta}^{\pi_{T(m)}}\left[\exp\left(n_{T(m), 2}\left(\frac{\tilde\theta_2^2 - \theta_2^2}{2\sigma^2} + \frac{(\theta_2 - \tilde\theta_2)\hat\theta_{T(m), 2}}{\sigma^2}\right)\right)1\{\hat\theta_{T(m), 2} > \tilde\theta_2 - M, n_{T(m), 2} < N\}\right] \\
& \geq \mathbb E_{\tilde\theta}^{\pi_{T(m)}}\left[\exp\left(N\left(-\frac{(\tilde\theta_2 - \theta_2)^2}{2\sigma^2} - \frac{M(\theta_2 - \tilde\theta_2)}{\sigma^2}\right)\right)1\{\hat\theta_{T(m), 2} > \tilde\theta_2 - M, n_{T(m), 2} < N\}\right] \\
& = \exp\left(N\left(-\frac{(\tilde\theta_2 - \theta_2)^2}{2\sigma^2} - \frac{M(\theta_2 - \tilde\theta_2)}{\sigma^2}\right)\right)\mathbb P_{\tilde\theta}^{\pi_{T(m)}}(\hat\theta_{T(m), 2} > \tilde\theta_2 - M, n_{T(m), 2} < N).
\end{align*}
Note that
\begin{align*}
& \quad\ \mathbb P_{\tilde\theta}^{\pi_{T(m)}}(\hat\theta_{T(m), 2} > \tilde\theta_2 - M, n_{T(m), 2} < N) \\
& = p - \sum_{n=1}^{N-1}\mathbb P_{\tilde\theta}^{\pi_{T(m)}}(\hat\theta_{T(m), 2} \leq \tilde\theta_2 - M, n_{T(m), 2} = n) \\
& \geq p - \sum_{n=1}^{N-1}\exp(-\frac{nM^2}{2\sigma^2}) \geq p - N\exp(-\frac{M^2}{2\sigma^2})=q>0.
\end{align*}
Therefore, there exists a constant positive probability such that $\pi$ pulls arm $2$ no more than $N$ times under $\theta$. As a result, $\pi$ incurs a linear expected regret under $\theta$, leading to a contradiction.

$\hfill\Box$

{\noindent\bf Proof of Theorem \ref{thm:lower-bound-fixed-time}.} 

1. We consider the environment where the noise $\epsilon$ is gaussian with standard deviation $\sigma$. Let $\theta_1 = 1/2$. Let $\theta(T) = (\theta_1, \theta_2(T))$ and $\tilde\theta(T) = (\theta_1, \tilde\theta_2(T))$, where $\theta_2(T) = \theta_1 + \frac{x(T)}{cT}$ and $\tilde\theta_2(T) = \theta_1 - \frac{x(T)}{cT}$. Here, $c<1$ is such that
\begin{align*}
    \limsup_{T\to+\infty}\frac{x(T)}{cT} < 1/2.
\end{align*}

For notation simplicity, we will write $\theta$ ($\tilde\theta$) instead of $\theta(T)$ ($\tilde\theta(T)$), but we must keep in mind that $\theta$ ($\tilde\theta$) is dependent on $T$. Also, we write 
$\sup_{\theta, \cD} \mbE\left[R_{\theta, \cD}^{\pi(T)}(T)\right] = R(T)$. Define
\begin{align*}
    E_T = \left\{|\hat\theta_{T, 2} - \tilde\theta_2|\leq 2\sigma\ln n_2/\sqrt{n_{2}}\right\}
\end{align*}
and
\begin{align*}
    F_T = \{n_2\leq 2\frac{T\cdot R(T)}{x(T)}\}.
\end{align*}
Then under the environment $\tilde\theta$, we have
\begin{align*}
    & \quad\ \mbP_{\tilde\theta, \cD}^{\pi(T)}(\bar F_T) = \mbP_{\tilde\theta, \cD}^{\pi(T)}\left(n_2 > 2\frac{T\cdot R(T)}{x(T)}\right) \leq \frac{\mbE_{\tilde\theta}^{\pi(T)}[n_2]}{2\frac{T\cdot R(T)}{x(T)}} \leq \frac{R(T)}{\frac{x(T)}{T}\cdot2\frac{T\cdot R(T)}{x(T)}} = 1/2.
\end{align*}
Combined with Lemma \ref{lemma:sub-linear}, we have
\begin{align*}
    \liminf_T \mbP_{\tilde\theta, \cD}^{\pi(T)}(E_T, F_T) \geq 1/2.
\end{align*}
Now for sufficiently large $T$, we have
\begin{align*}
    & \quad\ \mbP\left(R_{\theta, \cD}^{\pi(T)}(T)\geq x(T)\right) \\
    & \geq \mbP_{\theta, \cD}^{\pi(T)}(n_1\geq cT) \\
    & = \mbP_{\theta, \cD}^{\pi(T)}(n_2\leq (1-c)T) \\
    & \geq \mbP_{\theta, \cD}^{\pi(T)}(n_2\leq 2T\cdot R(T)/x(T)) \\
    & \geq \mbP_{\theta, \cD}^{\pi(T)}(E_T, F_T) \\
    & = \mbE_{\theta, \cD}^{\pi(T)}[\mathds 1\{E_T F_T\}] \\
    & = \mbE_{\tilde\theta}^{\pi(T)}\left[\exp\left(\sum_{n=1}^{n_2}\frac{(r_{t_2(n), 2} - \tilde\theta_2)^2 - (r_{t_2(n), 2} - \theta_2)^2}{2\sigma^2}\right)\mathds 1\{E_T F_T\}\right] \\
    & = \mbE_{\tilde\theta}^{\pi(T)}\left[\exp\left(n_2\left(\frac{\tilde\theta_2^2 - \theta_2^2}{2\sigma^2} + \frac{(\theta_2 - \tilde\theta_2)\hat\theta_{T, 2}}{\sigma^2}\right)\right)\mathds 1\{E_T F_T\}\right] \\
    & \geq \mbE_{\tilde\theta}^{\pi(T)}\left[\exp\left(n_2\left(\frac{\tilde\theta_2^2 - \theta_2^2}{2\sigma^2} + \frac{(\theta_2 - \tilde\theta_2)(\tilde\theta_{2} - 2\sigma\sqrt{\ln n_2}/\sqrt{n_2}}{\sigma^2}\right)\right)\mathds 1\{E_T F_T\}\right] \\
    & = \mbE_{\tilde\theta}^{\pi(T)}\left[\exp\left(-n_2\frac{(\tilde\theta_2 - \theta_2)^2}{2\sigma^2} - \frac{2\sigma\sqrt{\ln T}\sqrt{n_2}(\theta_2 - \tilde\theta_2)}{\sigma^2}\right)\mathds 1\{E_T F_T\}\right] \\
    & \geq \mbE_{\tilde\theta}^{\pi(T)}\left[\exp\left(-2\frac{T\cdot R(T)}{x(T)}\frac{(\tilde\theta_2 - \theta_2)^2}{2\sigma^2} - \frac{2\sigma\sqrt{\ln T} \sqrt{\frac{T\cdot R(T)}{x(T)}}(\theta_2 - \tilde\theta_2)}{\sigma^2}\right)\mathds 1\{E_T F_T\}\right] \\
    & = \exp(-4x(T)\cdot R(T)/Tc^2\sigma^2 - 4\sqrt{x(T)\cdot R(T)\ln T}/\sqrt{T}c\sigma)\mbP_{\tilde\theta, \cD}^{\pi(T)}(E_T, F_T).
\end{align*}
Therefore, together with $c$ being arbitrarily close to $1$, we have
\begin{align*}
    & \quad \liminf_T\frac{\ln\left\{\sup_{\theta, \cD}\mbP\left(R_{\theta, \cD}^{\pi(T)}(T) > x(T)\right)\right\}\cdot T}{x(T)\cdot\sup_{\theta, \cD}\mbE\left[R_{\theta, \cD}^{\pi(T)}(T)\right]}\cdot \min\left\{1, \sqrt{\frac{x(T)\cdot\sup_{\theta, \cD}\mbE\left[R_{\theta, \cD}^{\pi(T)}(T)\right]}{T\ln T}}\right\} \\
    & = \liminf_T\frac{\ln\left\{\sup_{\theta, \cD}\mbP\left(R_{\theta, \cD}^{\pi(T)}(T) > x(T)\right)\right\}}{\max\left\{x(T)R(T)/T, \sqrt{x(T)R(T)\ln T}/\sqrt{T}\right\}}
    \\
    & \geq -\inf_{0 < c < 1}\left\{\frac{4}{c^2\sigma^2} + \frac{4}{c\sigma}\right\} + \liminf_T \frac{\ln 1/2}{ \sqrt{x(T)R(T)\ln T}/\sqrt{T}}\\
    & = -\left(\frac{4}{\sigma^2} + \frac{4}{\sigma}\right) \triangleq -C.
\end{align*}

2. We consider the environment where the noise $\epsilon$ is gaussian with standard deviation $\sigma$. Denote $\Delta=\theta_2 - \theta_1$ and $\tilde\Delta = \tilde\theta_1 - \tilde\theta_2$. Define
\begin{align*}
    E_T = \left\{|\hat\theta_{T, 2} - \tilde\theta_2|\leq \varepsilon\right\}
\end{align*}
where $\varepsilon>0$ is a small number, and
\begin{align*}
    F_T = \{n_2\leq 2\mbE[R_{\tilde\theta, \cD}^{\pi(T)}(T)]/\tilde\Delta\}.
\end{align*}
with $\xi\in(\beta, \gamma)$. Under the environment $\tilde\theta$, we have
\begin{align*}
    & \quad\ \mbP_{\tilde\theta, \cD}^{\pi(T)}(\bar F_T) = \mbP_{\tilde\theta, \cD}^{\pi(T)}\left(n_2 > 2\mbE[R_{\tilde\theta, \cD}^{\pi(T)}(T)]/\tilde\Delta\right) \leq \frac{\mbE_{\tilde\theta}^{\pi(T)}[n_2]}{2\mbE[R_{\tilde\theta, \cD}^{\pi(T)}(T)]/\tilde\Delta} \leq \frac{1}{2}. 
\end{align*}
Combined with Lemma \ref{lemma:sub-linear}, we have
\begin{align*}
    \liminf_T \mbP_{\tilde\theta, \cD}^{\pi(T)}(E_T, F_T) \geq 1/2.
\end{align*}

Let $c\in(0, 1)$ such that
\begin{align*}
    \limsup \frac{x(T)}{cT} < \Delta.
\end{align*}
Take $T$ to be sufficiently large. Now
\begin{align*}
    & \quad\ \mbP\left(R_{\theta, \cD}^{\pi(T)}(T)\geq x(T)\right) \\
    & = \mbP_{\theta, \cD}^{\pi(T)}(n_1\geq x(T)/\Delta) \\
    & \geq \mbP_{\theta, \cD}^{\pi(T)}(n_1\geq cT) \\
    & \geq \mbP_{\theta, \cD}^{\pi(T)}(n_2\leq (1-c)T) \\
    & \geq \mbP_{\theta, \cD}^{\pi(T)}(n_2\leq 2\mbE[R_{\tilde\theta, \cD}^{\pi(T)}(T)]/\tilde\Delta) \\
    & \geq \mbP_{\theta, \cD}^{\pi(T)}(E_T, F_T) \\
    & = \mbE_{\theta, \cD}^{\pi(T)}[\mathds 1\{E_T F_T\}] \\
    & = \mbE_{\tilde\theta}^{\pi(T)}\left[\exp\left(\sum_{n=1}^{n_2}\frac{(r_{t_2(n), 2} - \tilde\theta_2)^2 - (r_{t_2(n), 2} - \theta_2)^2}{2\sigma^2}\right)\mathds 1\{E_T F_T\}\right] \\
    & = \mbE_{\tilde\theta}^{\pi(T)}\left[\exp\left(n_2\left(\frac{\tilde\theta_2^2 - \theta_2^2}{2\sigma^2} + \frac{(\theta_2 - \tilde\theta_2)\hat\theta_{T, 2}}{\sigma^2}\right)\right)\mathds 1\{E_T F_T\}\right] \\
    & \geq \mbE_{\tilde\theta}^{\pi(T)}\left[\exp\left(n_2\left(\frac{\tilde\theta_2^2 - \theta_2^2}{2\sigma^2} + \frac{(\theta_2 - \tilde\theta_2)(\tilde\theta_{2} - \varepsilon)}{\sigma^2}\right)\right)\mathds 1\{E_T F_T\}\right] \\
    & = \mbE_{\tilde\theta}^{\pi(T)}\left[\exp\left(n_2\left(-\frac{(\tilde\theta_2 - \theta_2)^2}{2\sigma^2} - \frac{\varepsilon(\theta_2 - \tilde\theta_2)}{\sigma^2}\right)\right)\mathds 1\{E_T F_T\}\right] \\
    & \geq \mbE_{\tilde\theta}^{\pi(T)}\left[\exp\left(\frac{2\mbE[R_{\tilde\theta, \cD}^{\pi(T)}(T)]}{\Delta}\left(-\frac{(\tilde\theta_2 - \theta_2)^2}{2\sigma^2} - \frac{\varepsilon(\theta_2 - \tilde\theta_2)}{\sigma^2}\right)\right)\mathds 1\{E_T F_T\}\right] \\
    & = \exp\left(-\frac{2\mbE[R_{\tilde\theta, \cD}^{\pi(T)}(T)]}{\tilde\Delta}\left(\frac{(\Delta+\tilde\Delta)^2}{2\sigma^2} + \frac{\varepsilon(\Delta+\tilde\Delta)}{\sigma^2}\right)\right)\mbP_{\tilde\theta, \cD}^{\pi(T)}(E_T, F_T).
\end{align*}
Therefore, together with $\varepsilon$ being arbitrarily small, we have
\begin{align*}
    & \quad \liminf_T\frac{\ln\left\{\mbP\left(R_{\theta, \cD}^{\pi(T)}(T)\geq x(T)\right)\right\}}{\mbE[R_{\tilde\theta, \cD}^{\pi(T)}(T)]} \\
    & \geq -\frac{2}{\tilde\Delta}\inf_{\varepsilon > 0}\left\{\frac{(\Delta+\tilde\Delta)^2}{2\sigma^2} + \frac{\varepsilon(\Delta+\tilde\Delta)}{\sigma^2}\right\} + \liminf_T \frac{\ln 1/2}{\mbE[R_{\tilde\theta, \cD}^{\pi(T)}(T)]}\\
    & = -\frac{(\Delta+\tilde\Delta)^2}{\tilde\Delta\sigma^2} \triangleq -C.
\end{align*}

$\hfill\Box$

{\noindent\bf Proof of Proposition \ref{prop:lower-bound}.}

1. In Theorem \ref{thm:lower-bound-fixed-time}, we take $x(T) = cT^{\delta}$ with $\delta>\alpha\geq 1/2$. Take $\gamma > \alpha+\delta-1$. Then
\begin{align*}
    x(T)\cdot\sup_{\theta, \cD}\mbE\left[R_{\theta, \cD}^{\pi(T)}(T)\right] & = \Omega(T^{\delta+1/2}) = \omega(T\ln T), \\
    x(T)\cdot\sup_{\theta, \cD}\mbE\left[R_{\theta, \cD}^{\pi(T)}(T)\right] & = o(T^{\alpha+\delta}) = o(T^{1+\gamma}).
\end{align*}
We have
\begin{align*}
    & \quad \liminf_T\frac{\ln\left\{\sup_{\theta, \cD}\mbP\left(R_{\theta, \cD}^{\pi(T)}(T) > cT^\delta\right)\right\}}{T^\gamma} \\
    & = \liminf_T\frac{\ln\left\{\sup_{\theta, \cD}\mbP\left(R_{\theta, \cD}^{\pi(T)}(T) > x(T)\right)\right\}\cdot T}{x(T)\cdot\sup_{\theta, \cD}\mbE\left[R_{\theta, \cD}^{\pi(T)}(T)\right]}\cdot \frac{\sup_{\theta, \cD} x(T)\cdot\mbE\left[R_{\theta, \cD}^{\pi(T)}(T)\right]}{T^{1+\gamma}} \\
    & = \liminf_T\frac{\ln\left\{\sup_{\theta, \cD}\mbP\left(R_{\theta, \cD}^{\pi(T)}(T) > x(T)\right)\right\}\cdot T}{x(T)\cdot\sup_{\theta, \cD}\mbE\left[R_{\theta, \cD}^{\pi(T)}(T)\right]}\cdot \min\left\{1, \sqrt{\frac{x(T)\cdot\sup_{\theta, \cD}\mbE\left[R_{\theta, \cD}^{\pi(T)}(T)\right]}{T\ln T}}\right\}\cdot\\
    & \quad\quad\quad \frac{x(T)\cdot\sup_{\theta, \cD}\mbE\left[R_{\theta, \cD}^{\pi(T)}(T)\right]}{T^{1+\gamma}} \\
    & \geq -C\cdot 0 = 0.
\end{align*}

2. In Theorem \ref{thm:lower-bound-fixed-time}, we take $x(T)=cT^\delta$ with $\delta>\beta$. Take $\gamma>\beta$. Then
\begin{align*}
    & \quad \lim\inf_T\frac{\ln\left\{\mbP\left(R_{\theta, \cD}^{\pi(T)}(T) > cT^\delta\right)\right\}}{T^\gamma} \\
    & = \lim\inf_T\frac{\ln\left\{\mbP\left(R_{\theta, \cD}^{\pi(T)}(T) > x(T)\right)\right\}}{\mbE\left[R_{\tilde\theta, \cD}^{\pi(T)}(T)\right]}\cdot\frac{\mbE\left[R_{\tilde\theta, \cD}^{\pi(T)}(T)\right]}{T^\gamma} \\
    & \geq -C\cdot 0 = 0.
\end{align*}

If $\pi$ does not know $T$ a priori, then we take $\pi(1) = \cdots = \pi(T) = \cdots = \pi$. Take $\gamma>\delta\beta$. Let $\Delta=|\theta_1-\theta_2|$ be the gap between the two arms. Define
\begin{align*}
    T_k = \left\lceil\frac{2ck^\delta}{\Delta}\right\rceil
\end{align*}
Then
\begin{align*}
    \mbE\left[R_{\tilde\theta, \cD}^{\pi_{T_k}}(T_k)\right] = o\left(T_k^{\gamma/\delta}\right) = o(k^\gamma).
\end{align*}
The first equality holds because $\pi$ is $\beta$-consistent and $\beta < \gamma/\delta$. We have 
\begin{align*}
    & \quad \lim\inf_T\frac{\ln\left\{\mbP\left(R_{\theta, \cD}^{\pi(T)}(T) > cT^\delta\right)\right\}}{T^\gamma} \\
    & = \lim\inf_k\frac{\ln\left\{\mbP\left(R_{\theta, \cD}^{\pi_k}(k) > ck^\delta\right)\right\}}{k^\gamma} \\
    & \geq \lim\inf_k\frac{\ln\left\{\mbP\left(R_{\theta, \cD}^{\pi_{T_k}}(T_k) > ck^\delta\right)\right\}}{k^\gamma} \\
    & \geq \lim\inf_k\frac{\ln\left\{\mbP\left(R_{\theta, \cD}^{\pi_{T_k}}(T) > \Delta T_k/2\right)\right\}}{\mbE\left[R_{\tilde\theta, \cD}^{\pi_{T_k}}(T_k)\right]}\cdot\frac{\mbE\left[R_{\tilde\theta, \cD}^{\pi_{T_k}}(T_k)\right]}{k^\gamma} \\
    & \geq \lim\inf_{k}\frac{\ln\left\{\mbP\left(R_{\theta, \cD}^{\pi_{T_k}}(T) > \Delta T_k/2\right)\right\}}{\mbE\left[R_{\tilde\theta, \cD}^{\pi_{T_k}}(T_k)\right]}\cdot\frac{\mbE\left[R_{\tilde\theta, \cD}^{\pi_{T_k}}(T_k)\right]}{k^\gamma} \\
    & \geq -C\cdot 0 = 0.
\end{align*}

$\hfill\Box$

\section{Proofs for Section \ref{sec:upper}}

The following simple inequalities would be useful when proving instance-dependent bounds. For any $a, b\geq 0$, 
\begin{align*}
    (\sqrt{a}-\sqrt{b})^2 & \geq (a/2-b)_+, \\
    (a - \sqrt{ab})_+ & \geq (a/2 - b/2)_+.
\end{align*}

{\noindent\bf Proof of Theorem \ref{thm:upper-bound-fixed}.}
Without loss of generality, we assume $\theta_1  = \theta_*$. Let $\pi = \UCB$ and $x\geq K$. We prove the bounds for two scenarios separately.

{\noindent\bf 1. Worst-case scenario.} Define
\begin{align*}
    \cA^* = \left\{k\neq 1: n_k\leq 1 + T/K\right\}.
\end{align*}
We have
\begin{align*}
    & \quad\;\mbP\left(R_\theta^\pi(T)\geq x\right) \\
    & = \mbP\left(\sum_{k\in \cA^*}n_k\Delta_k + \sum_{k\notin \cA^*}n_k\Delta_k\geq x\right) \\
    & \leq \mbP\left(\sum_{k\in \cA^*}(n_k-1)\Delta_k + \sum_{k\notin \cA^*}(n_k-1)\Delta_k\geq x - K\right) \\
    & \leq \mbP\left(\left(\bigcup_{k\in\cA^*}\left\{(n_k-1)\Delta_k\geq\frac{x-K}{2K}\right\}\right)\bigcup\left(\bigcup_{k\notin\cA^*}\left\{\Delta_k\geq \frac{x-K}{2T}\right\}\right)\right) \\
    & \leq \sum_{k\neq 1}\mbP\left(\left\{(n_k-1)\Delta_k\geq \frac{x-K}{2K}, \ k\in\cA^*\right\} \bigcup \left\{\Delta_k\geq \frac{x-K}{2T}, \ k\notin\cA^*\right\}\right) \\
    & \triangleq \sum_{k\neq 1}\mbP\left(\cB_k\cup\cC_k\right).
\end{align*}
To prove the second inequality, we only need to show that the following cannot hold simultaneously:
\begin{align*}
    (n_k-1)\Delta_k < \frac{x-K}{2K}, \quad \forall k\in\cA^*; \quad \quad \Delta_k < \frac{x-K}{2T}, \quad \forall k\notin\cA^*.
\end{align*}
If not, then we have
\begin{align*}
    & \quad\;\sum_{k\neq 1} (n_k-1)\Delta_k \\
    & = \sum_{k\in\cA^*}(n_k-1)\Delta_k + \sum_{k\notin\cA^*}(n_k-1)\Delta_k \\
    & < \frac{(x-K)|\cA^*|}{2K} + \frac{x-K}{2} \\
    & \leq \frac{x-K}{2} + \frac{x-K}{2} \\
    & = x - K.
\end{align*}

Fix $k\neq 1$. We let $m_k = \frac{x-K}{2K\Delta_k}$. 
\begin{itemize}
\item $\cB_k$ happens. Then
\begin{align*}
    \frac{T}{K}\geq n_k - 1\geq m_k \quad \Longrightarrow \quad \frac{T}{K} \wedge (n_k-1) \geq m_k.
\end{align*}
\item $\cC_k$ happens. Then
\begin{align*}
    n_k - 1 \geq \frac{T}{K} \geq m_k \quad \Longrightarrow \quad \frac{T}{K} \wedge (n_k-1) \geq m_k.
\end{align*}
\end{itemize}
Consider the time we pull arm $k$ for the $\lceil m_k\rceil+1$th time (which is $t_k(\lceil m_k\rceil+1)$; for simplicity we write it as $t_k$) We know that the following happens w.p. 1:
\begin{align*}
    \hat\mu_{t_k-1, 1} + \rad(n_{t_k-1, 1})\leq \hat\mu_{t_k-1, k} + \rad(\lceil m_k\rceil).
\end{align*}
We have
\begin{align*}
    & \quad\;\mbP\brs{\cB_k\cup\cC_k} \\
    & \leq \mbP\brs{\hat\mu_{t_k-1, 1} + \rad(n_{t_k-1, 1})\leq \hat\mu_{t_k-1, k} + \rad(\lceil m_k\rceil),\ \frac{T}{K} \wedge (n_k-1) \geq m_k} \\
    & = \mbP\brs{\mu_1 + \frac{\sum_{\ell=1}^{n_{t_k-1, 1}}\epsilon_{t_1(\ell), 1}}{n_{t_k-1, 1}} + \rad(n_{t_k-1, 1})\leq \mu_k + \frac{\sum_{\ell=1}^{\lceil m_k\rceil}\epsilon_{t_k(\ell), k}}{\lceil m_k\rceil} + \rad(\lceil m_k\rceil),\ \frac{T}{K} \wedge (n_k-1) \geq m_k} \\
    & \leq \mbP\brs{\frac{\sum_{\ell=1}^{\lceil m_k\rceil}\epsilon_{t_k(\ell), k}}{\lceil m_k\rceil} \geq \frac{\Delta_k}{2} - \rad(\lceil m_k\rceil), \ \frac{T}{K} \wedge (n_k-1) \geq m_k} + \\ 
    & \quad\quad \mbP\brs{\frac{\sum_{\ell=1}^{n_{t_k-1, 1}}\epsilon_{t_1(\ell), 1}}{n_{t_k-1, 1}} \leq - \frac{\Delta_k}{2} - \rad(n_{t_k-1, 1}), \ \frac{T}{K} \wedge (n_k-1) \geq m_k} \\
    & \leq \mbP\brs{\frac{\sum_{\ell=1}^{\lceil m_k\rceil}\epsilon_{t_k(\ell), k}}{\lceil m_k\rceil} \geq \frac{\Delta_k}{2} - \rad(\lceil m_k\rceil), \ \frac{T}{K} \geq m_k} + \mbP\brs{\exists n: \frac{\sum_{\ell=1}^{n}\epsilon_{t_1(\ell), 1}}{n} \leq - \frac{\Delta_k}{2} - \rad(n), \ \frac{T}{K} \geq m_k} \\
    & \triangleq \mbP\brs{\cE_k} + \mbP\brs{\cF_k}.
\end{align*}

We have the following bounds on the two tail probabilities.
\begin{align*}
    \mbP\brs{\cE_k} 
    & \leq \exp\brs{-\lceil m_k\rceil\frac{\brs{\frac{\Delta_k}{2} - \rad(\lceil m_k\rceil)}^2}{2\sigma^2}} \\
    & \leq \exp\brs{-m_k\frac{\brs{\frac{\Delta_k}{2} - \rad(m_k)}^2}{2\sigma^2}} \\
    & \leq \exp\brs{-\frac{x-K}{2K\Delta_k}\cdot\frac{\brs{\frac{\Delta_k}{2} - \eta\frac{(T/K)^\alpha\sqrt{\ln K}}{x-K}(2K\Delta_k)}^2}{2\sigma^2}} \\
    & = \exp\brs{-\Delta_k\frac{(x-K)}{4K\sigma^2}\brs{\frac{1}{2}-\eta\frac{2K^{1-\alpha}T^{\alpha}\sqrt{\ln K}}{x-K}}^2} \\
    & \leq \exp\brs{-\frac{\brs{x-K-4\eta K^{1-\alpha}T^{\alpha}\sqrt{\ln K}}_+^2}{32\sigma^2KT}}.
\end{align*}
Meanwhile, 
\begin{align*}
    \mbP(\cF_k) & \leq \mbP\brs{\exists n: \frac{\sum_{\ell=1}^{n}\epsilon_{t_1(\ell), 1}}{n} \leq - \frac{\Delta_k}{2} - \eta\frac{(T/K)^\alpha\sqrt{\ln K}}{n}, \ \frac{T}{K} \geq m_k} + \\
    & \quad\quad \mbP\brs{\exists n: \frac{\sum_{\ell=1}^{n}\epsilon_{t_1(\ell), 1}}{n} \leq - \frac{\Delta_k}{2} - \sqrt{\frac{f(T)}{n}}, \ \frac{T}{K} \geq m_k} \\
    & \leq \exp\brs{-2\frac{\eta}{\sigma^2}\frac{x-K}{4T}(T/K)^\alpha\sqrt{\ln K}} + \sum_{n=1}^T\exp\brs{-\frac{f(T)}{2\sigma^2}} \\
    & \leq \exp\brs{-\frac{\eta(x-K)_+\sqrt{\ln K}}{2\sigma^2K^{\alpha} T^{1-\alpha}}} + T\exp\brs{-\frac{f(T)}{2\sigma^2}}.
\end{align*}

Note that the equations above hold for any instance $\theta$. Combining all the equations above yields
\begin{align*}
    & \quad\; \sup_{\theta}\mbP(R_\theta^\pi(T)\geq x) \\
    & \leq K\exp\left(\frac{\left(x-K-4\eta K^{1-\alpha}T^\alpha\sqrt{\ln K}\right)_+^2}{32\sigma^2KT}\right) + K\exp\brs{-\frac{\eta(x-K)_+\sqrt{\ln K}}{2\sigma^2K^{\alpha} T^{1-\alpha}}} + KT\exp\brs{-\frac{f(T)}{2\sigma^2}}.
\end{align*}

{\noindent\bf 2. Instance-dependent scenario.} We have
\begin{align*}
    & \quad\;\mbP\left(R_\theta^\pi(T)\geq x\right) \\
    & = \mbP\left(\sum_{k: \Delta_k>0}n_k\Delta_k \geq x\right) \\
    & \leq \mbP\left(\sum_{k: \Delta_k>0}(n_k-1)\Delta_k\geq x - K\right) \\
    & \leq \mbP\left(\bigcup_{k: \Delta_k>0}\left\{(n_k-1)\Delta_k\geq\frac{(x-K)/\Delta_k}{\sum_{k': \Delta_{k'}>0}1/\Delta_{k'}}\right\}\right) \\
    & \leq \sum_{k: \Delta_k>0}\mbP\left((n_k-1)\Delta_k\geq\frac{(x-K)/\Delta_k}{\sum_{k': \Delta_{k'}>0}1/\Delta_{k'}}\right).
\end{align*}

Denote 
\begin{align*}
    \Delta_0 = \frac{1}{\sum_{k': \Delta_{k'}>0}1/\Delta_{k'}}.
\end{align*}
Fix $k: \Delta_k>0$. Now for each $k$, we let
\begin{align*}
    m_k = (x-K)\Delta_0/\Delta_k^2 \leq n_k-1.
\end{align*}
Consider the time we pull arm $k$ for the $\lceil m_k\rceil+1$th time (which is $t_k(\lceil m_k\rceil+1)$; for simplicity we write it as $t_k$) We know that the following happens w.p. 1:
\begin{align*}
    \hat\mu_{t_k-1, 1} + \rad(n_{t_k-1, 1})\leq \hat\mu_{t_k-1, k} + \rad(\lceil m_k\rceil).
\end{align*}
We have
\begin{align*}
    & \quad\; \mbP\left((n_k-1)\Delta_k\geq(x-K)\Delta_0/\Delta_k\right) \nonumber\\
    & \leq \mbP\brs{\hat\mu_{t_k-1, 1} + \rad(n_{t_k-1, 1})\leq \hat\mu_{t_k-1, k} + \rad(\lceil m_k\rceil)} \nonumber\\
    & = \mbP\left(\mu_1 + \frac{\sum_{\ell=1}^{n_{t_k-1, 1}}\epsilon_{t_1(\ell), 1}}{n_{t_k-1, 1}} + \rad(n_{t_k-1, 1})\leq \mu_k + \frac{\sum_{\ell=1}^{\lceil m_k\rceil}\epsilon_{t_k(\ell), k}}{\lceil m_k\rceil} + \rad(\lceil m_k\rceil)\right) \nonumber\\
    & \leq \mbP\brs{\frac{\sum_{\ell=1}^{\lceil m_k\rceil}\epsilon_{t_k(\ell), k}}{\lceil m_k\rceil} \geq \frac{\Delta_k}{2} - \rad(\lceil m_k\rceil)} + \mbP\brs{\frac{\sum_{\ell=1}^{n_{t_k-1, 1}}\epsilon_{t_1(\ell), 1}}{n_{t_k-1, 1}} \leq - \frac{\Delta_k}{2} - \rad(n_{t_k-1, 1})} \nonumber\\
    & \leq \mbP\brs{\frac{\sum_{\ell=1}^{\lceil m_k\rceil}\epsilon_{t_k(\ell), k}}{\lceil m_k\rceil} \geq \frac{\Delta_k}{2} - \rad(\lceil m_k\rceil)} + \mbP\brs{\exists n: \frac{\sum_{\ell=1}^{n}\epsilon_{t_1(\ell), 1}}{n} \leq - \frac{\Delta_k}{2} - \rad(n)} \nonumber\\
    & \triangleq \mbP\brs{\cE_k} + \mbP\brs{\cF_k}.
\end{align*}

We have the following bounds on the two tail probabilities.
\begin{align*}
    \mbP\brs{\cE_k} 
    & \leq \exp\brs{-\lceil m_k\rceil\frac{\brs{\frac{\Delta_k}{2} - \rad(\lceil m_k\rceil)}^2}{2\sigma^2}} \\
    & \leq \exp\brs{-m_k\frac{\brs{\frac{\Delta_k}{2} - \rad(m_k)}^2}{2\sigma^2}} \\
    & = \exp\brs{-\frac{(x-K)\Delta_0}{\Delta_k^2}\cdot\frac{\brs{\frac{\Delta_k}{2} - \sqrt{f(T)\frac{\Delta_k^2}{(x-K)\Delta_0}}}^2}{2\sigma^2}} \\
    & = \exp\brs{\frac{\brs{\frac{\sqrt{(x-K)\Delta_0}}{2} - \sqrt{f(T)}}^2}{2\sigma^2}} \\
    & \leq \exp\brs{-\frac{\brs{(x-K)\Delta_0-8f(T)}_+}{16\sigma^2}}.
\end{align*}
Meanwhile, 
\begin{align*}
    \mbP(\cF_k) & \leq \mbP\brs{\exists n: \frac{\sum_{\ell=1}^{n}\epsilon_{t_1(\ell), 1}}{n} \leq - \frac{\Delta_k}{2} - \eta\frac{(T/K)^\alpha\sqrt{\ln K}}{n}} + \\
    & \quad\quad \mbP\brs{\exists n: \frac{\sum_{\ell=1}^{n}\epsilon_{t_1(\ell), 1}}{n} \leq - \frac{\Delta_k}{2} - \sqrt{\frac{f(T)}{n}}} \\
    & \leq \exp\brs{-2\frac{\eta}{\sigma^2}\frac{\Delta_k}{2}(T/K)^\alpha\sqrt{\ln K}} + \sum_{n=1}^T\exp\brs{-\frac{f(T)}{2\sigma^2}} \\
    & \leq \exp\brs{-\frac{\eta\Delta_kT^\alpha\sqrt{\ln K}}{\sigma^2K^{\alpha}}} + T\exp\brs{-\frac{f(T)}{2\sigma^2}}.
\end{align*}

Combining all the equations above yields
\begin{align*}
    & \quad\; \mbP(R_\theta^\pi(T)\geq x) \\
    & \leq K\exp\brs{-\frac{\brs{(x-K)\Delta_0-8f(T)}_+}{16\sigma^2}} + \sum_{k:\Delta_k>0}\exp\brs{-\frac{\eta\Delta_kT^\alpha\sqrt{\ln K}}{\sigma^2K^{\alpha}}} + KT\exp\brs{-\frac{f(T)}{2\sigma^2}}.
\end{align*}

$\hfill\Box$

{\noindent\bf Proof of Theorem \ref{thm:upper-bound-any}.} Fix a time horizon of $T$. We write $t_k = t_k(n_{T, k})$ as the last time that arm $k$ is pulled throughout the $T$ time periods. Without loss of generality, we assume $\theta_1  = \theta_*$. Let $\pi = \UCB$ and $x\geq K$. We prove the bounds for two scenarios separately.

{\noindent\bf 1. Worst-case scenario.} Define $c_\alpha = \frac{1-\alpha}{2-\alpha}$ and
\begin{align*}
    \cA' = \left\{k\neq 1: n_k\leq 1 + \frac{t_k^\alpha T^{1-\alpha}}{K}\right\}.
\end{align*}
We have
\begin{align*}
    & \quad\;\mbP\left(R_\theta^\pi(T)\geq x\right) \\
    & = \mbP\left(\sum_{k\in \cA'}n_k\Delta_k + \sum_{k\notin \cA'}n_k\Delta_k\geq x\right) \\
    & \leq \mbP\left(\sum_{k\in \cA'}(n_k-1)\Delta_k + \sum_{k\notin \cA'}(n_k-1)\Delta_k\geq x - K\right) \\
    & \leq \mbP\left(\left(\bigcup_{k\in\cA'}\left\{(n_k-1)\Delta_k\geq (1-c_\alpha)\frac{x-K}{K}\right\}\right)\bigcup\left(\bigcup_{k\notin\cA'}\left\{\Delta_k\geq c_\alpha\frac{x-K}{t_k^\alpha T^{1-\alpha}}\right\}\right)\right) \\
    & \leq \sum_{k\neq 1}\mbP\left(\left\{(n_k-1)\Delta_k\geq c_\alpha\frac{x-K}{K}, \ k\in\cA'\right\} \bigcup \left\{\Delta_k\geq c_\alpha\frac{x-K}{t_k^\alpha T^{1-\alpha}}, \ k\notin\cA'\right\}\right) \\
    & \triangleq \sum_{k\neq 1}\mbP\left(\cB_k\cup\cC_k\right).
\end{align*}
To prove the second inequality, we only need to show that the following cannot hold simultaneously:
\begin{align*}
    (n_k-1)\Delta_k < c_\alpha\frac{x-K}{K}, \quad \forall k\in\cA'; \quad \quad (n_k-1)\Delta_k < c_\alpha\frac{(n_k-1)(x-2K)}{t_k^\alpha T^{1-\alpha}}, \quad \forall k\notin\cA'.
\end{align*}
If not, then we have
\begin{align*}
    \sum_{k\neq 1} (n_k-1)\Delta_k & = \sum_{k\in\cA'}(n_k-1)\Delta_k + \sum_{k\notin\cA'}(n_k-1)\Delta_k \\
    & < c_\alpha\frac{(x-K)|\cA'|}{K} + c_\alpha\frac{x-K}{T^{1-\alpha}}\sum_{k\notin\cA'}\frac{n_k}{t_k^\alpha} \\
    & \leq c_\alpha(x-K) + c_\alpha\frac{x-K}{T^{1-\alpha}}\sum_{k\notin\cA'}\frac{n_k}{t_k^\alpha} \\
    & \leq x - K.
\end{align*}
In fact, to bound $\sum_{k\notin\cA'}\frac{n_k}{t_k}$, we can assume $0=t_{k_0} < t_{k_1} < t_{k_2} < \cdots$. Then we have
\begin{align*}
    t_{k_i} \geq n_{k_1} + \cdots + n_{k_i}
\end{align*}
because before up to time $t_{k_i}$, arms $k_1, \cdots, k_i$ have been pulled completely, and after time $t_{k_i}$ none of them will be pulled. Thus, 
\begin{align*}
    \sum_{k\notin\cA'}\frac{n_k}{t_k^\alpha} & = \sum_{i=1}^{|\cA'^c|}\frac{n_{k_i}}{\brs{\sum_{j=1}^i n_{k_i}}^\alpha} \leq 1+\int_1^T \frac{1}{t^\alpha}dt= 1+\frac{T^{1-\alpha}-1}{1-\alpha} \leq \frac{T^{1-\alpha}}{1-\alpha}.
\end{align*}

Fix $k\neq 1$. We let $m_k = c_\alpha\frac{x-K}{K\Delta_k}$. 
\begin{itemize}
\item $\cB_k$ happens. Then
\begin{align*}
    \frac{t_k^\alpha T^{1-\alpha}}{K}\geq n_k - 1\geq m_k \quad \Longrightarrow \quad \frac{t_k^\alpha T^{1-\alpha}}{K} \wedge (n_k-1) \geq m_k.
\end{align*}
\item $\cC_k$ happens. Then
\begin{align*}
    n_k - 1 \geq \frac{t_k^\alpha T^{1-\alpha}}{K} \geq m_k \quad \Longrightarrow \quad \frac{t_k^\alpha T^{1-\alpha}}{K} \wedge (n_k-1) \geq m_k.
\end{align*}
\end{itemize}
We additionally define $T_k$ such that $T_k^\alpha T^{1-\alpha} = Km_k$. Then $t_k\geq T_k$ if $\cB_k$ or $\cC_k$ happens.

Consider the time we pull arm $k$ for the last ($n_k$th) time. We know that the following happens w.p. 1:
\begin{align*}
    \hat\mu_{t_k-1, 1} + \rad_{t_k}(n_{t_k-1, 1})\leq \hat\mu_{t_k-1, k} + \rad_{t_k}(n_k-1).
\end{align*}
We have
\begin{align*}
    & \quad\;\mbP\brs{\cB_k\cup\cC_k} \\
    & \leq \mbP\brs{\hat\mu_{t_k-1, 1} + \rad_{t_k}(n_{t_k-1, 1})\leq \hat\mu_{t_k-1, k} + \rad_{t_k}(n_k-1),\ \frac{t_k^\alpha T^{1-\alpha}}{K} \wedge (n_k-1) \geq m_k} \\
    & = \mbP\brs{\mu_1 + \frac{\sum_{\ell=1}^{n_{t_k-1, 1}}\epsilon_{t_1(\ell), 1}}{n_{t_k-1, 1}} + \rad_{t_k}(n_{t_k-1, 1})\leq \mu_k + \frac{\sum_{\ell=1}^{n_k-1}\epsilon_{t_k(\ell), k}}{n_k-1} + \rad_{t_k}(n_k-1),\ \frac{t_k^\alpha T^{1-\alpha}}{K} \wedge (n_k-1) \geq m_k} \\
    & \leq \mbP\brs{\frac{\sum_{\ell=1}^{n_k-1}\epsilon_{t_k(\ell), k}}{n_k-1} \geq \frac{\Delta_k}{2} - \rad_{t_k}(n_k-1), \ \frac{t_k^\alpha T^{1-\alpha}}{K} \wedge (n_k-1) \geq m_k} + \\ 
    & \quad\quad \mbP\brs{\frac{\sum_{\ell=1}^{n_{t_k-1, 1}}\epsilon_{t_1(\ell), 1}}{n_{t_k-1, 1}} \leq - \frac{\Delta_k}{2} - \rad_{t_k}(n_{t_k-1, 1}), \ \frac{t_k^\alpha T^{1-\alpha}}{K} \wedge (n_k-1) \geq m_k} \\
    & \leq \mbP\brs{\frac{\sum_{\ell=1}^{n_k-1}\epsilon_{t_k(\ell), k}}{n_k-1} \geq \frac{\Delta_k}{2} - \rad_{T}(n_k-1), \ n_k-1 \geq m_k} + \mbP\brs{\exists n: \frac{\sum_{\ell=1}^{n}\epsilon_{t_1(\ell), 1}}{n} \leq - \frac{\Delta_k}{2} - \rad_{t_k}(n)} \\
    & \triangleq \mbP\brs{\cE_k} + \mbP\brs{\cF_k}.
\end{align*}

We have the following bounds on the two tail probabilities.
\begin{align*}
    & \quad\; \mbP\brs{\cE_k} \\
    & \leq \mbP\brs{\exists n\geq m_k: \sum_{\ell=1}^{n}\epsilon_{t_k(\ell), k}\geq \frac{\Delta_k}{2}n - \eta(T/K)^{\alpha}\sqrt{\ln K}} \\
    & \leq \mbP\brs{\exists n\geq 0: \sum_{\ell=\lceil m_k\rceil+1}^{\lceil m_k\rceil + n}\epsilon_{t_k(\ell), k}\geq \frac{\Delta_k}{2}n + \frac{\lceil m_k\rceil\Delta_k}{4}} + \mbP\brs{\sum_{\ell=1}^{\lceil m_k\rceil}\epsilon_{t_k(\ell), k}\geq \frac{\lceil m_k\rceil\Delta_k}{4} -\frac{\eta(T/K)^{\alpha}\sqrt{\ln K}}{2}} \\
    & \leq \exp\brs{-2\frac{m_k\Delta_k^2}{8\sigma^2}} + \exp\brs{-\lceil m_k\rceil\frac{\brs{\frac{\Delta_k}{4} - \frac{\eta(T/K)^{\alpha}\sqrt{\ln K}}{2m_k}}^2}{2\sigma^2}} \\
    & \leq 2\exp\brs{-m_k\frac{\brs{\frac{\Delta_k}{4} - \frac{\eta(T/K)^{\alpha}\sqrt{\ln K}}{2m_k}}^2}{2\sigma^2}} \\
    & \leq 2\exp\brs{-c_{\alpha}\frac{x-K}{K\Delta_k}\cdot\frac{\brs{\frac{\Delta_k}{4} - \eta\frac{(T/K)^\alpha\sqrt{\ln K}}{2c_\alpha(x-K)}(2K\Delta_k)}^2}{2\sigma^2}} \\
    & = 2\exp\brs{-c_\alpha\Delta_k\frac{(x-K)}{2K\sigma^2}\brs{\frac{1}{4}-\eta\frac{K^{1-\alpha}T^{\alpha}\sqrt{\ln K}}{c_\alpha(x-K)}}^2} \\
    & \leq 2\exp\brs{-\frac{\brs{c_\alpha(x-K)-4\eta K^{1-\alpha}T^{\alpha}\sqrt{\ln K}}_+^2}{32\sigma^2KT}}.
\end{align*}
Meanwhile, 
\begin{align*}
    \mbP(\cF_k) & \leq \mbP\brs{\exists n: \frac{\sum_{\ell=1}^{n}\epsilon_{t_1(\ell), 1}}{n} \leq - \frac{\Delta_k}{2} - \eta\frac{(t_k/K)^\alpha\sqrt{\ln K}}{n}} + \\
    & \quad\quad \mbP\brs{\exists n: \frac{\sum_{\ell=1}^{n}\epsilon_{t_1(\ell), 1}}{n} \leq - \frac{\Delta_k}{2} - \sqrt{\frac{f(t_k)}{n}}} \\
    & \leq \exp\brs{-2\frac{\eta}{\sigma^2}\frac{\Delta_k}{2}(x/K)^\alpha\sqrt{\ln K}} + \sum_{n=1}^{\lfloor x\rfloor}\exp\brs{-\frac{f(x)}{2\sigma^2}} + \sum_{\lfloor x\rfloor +1}^T\exp\brs{-\frac{f(n)}{2\sigma^2}} \\
    & \leq \exp\brs{-c_\alpha\frac{\eta x\sqrt{\ln K}}{\sigma^2K^{\alpha} T^{1-\alpha}}} + \int_{0}^T\exp\brs{-\frac{f(x\vee y)}{2\sigma^2}}dy.
\end{align*}
Note that we have utilized the fact that $t_k \geq x$ and $t_k\geq n$.

Note that the equations above hold for any instance $\theta$. Combining all the equations above yields
\begin{align*}
    & \quad\sup_{\theta}\mbP(R_\theta^\pi(T)\geq x) \\
    & \leq K\exp\left(-\frac{\left(c_\alpha(x-K)-4\eta K^{1-\alpha}T^\alpha\sqrt{\ln K}\right)_+^2}{32\sigma^2KT}\right) + K\exp\brs{-c_\alpha\frac{\eta x\sqrt{\ln K}}{2\sigma^2K^{\alpha} T^{1-\alpha}}} + K\int_{0}^T\exp\brs{-\frac{f(x\vee y)}{2\sigma^2}}dy.
\end{align*}

{\noindent\bf 2. Instance-dependent scenario.}  We have
\begin{align*}
    & \quad\;\mbP\left(R_\theta^\pi(T)\geq x\right) \\
    & = \mbP\left(\sum_{k: \Delta_k>0}n_k\Delta_k \geq x\right) \\
    & \leq \mbP\left(\sum_{k: \Delta_k>0}(n_k-1)\Delta_k\geq x - K\right) \\
    & \leq \mbP\left(\bigcup_{k: \Delta_k>0}\left\{(n_k-1)\Delta_k\geq\frac{(x-K)/\Delta_k}{\sum_{k': \Delta_{k'}>0}1/\Delta_{k'}}\right\}\right) \\
    & \leq \sum_{k: \Delta_k>0}\mbP\left((n_k-1)\Delta_k\geq\frac{(x-K)/\Delta_k}{\sum_{k': \Delta_{k'}>0}1/\Delta_{k'}}\right).
\end{align*}
Denote 
\begin{align*}
    \Delta_0 = \frac{1}{\sum_{k': \Delta_{k'}>0}1/\Delta_{k'}}.
\end{align*}
Fix $k: \Delta_k>0$. Now for each $k$, we let
\begin{align*}
    m_k = (x-K)\Delta_0/\Delta_k^2\leq n_k-1. 
\end{align*}
Consider the time we pull arm $k$ for the $n_k$th time. We know that the following happens w.p. 1:
\begin{align*}
    \hat\mu_{t_k-1, 1} + \rad_{t_k}(n_{t_k-1, 1})\leq \hat\mu_{t_k-1, k} + \rad_{t_k}(n_k-1).
\end{align*}
We have
\begin{align*}
    & \quad\; \mbP\left((n_k-1)\Delta_k\geq(x-K)\Delta_0/\Delta_k\right) \\
    & \leq \mbP\brs{\hat\mu_{t_k-1, 1} + \rad_{t_k}(n_{t_k-1, 1})\leq \hat\mu_{t_k-1, k} + \rad_{t_k}(n_k-1)} \\
    & = \mbP\left(\mu_1 + \frac{\sum_{\ell=1}^{n_{t_k-1, 1}}\epsilon_{t_1(\ell), 1}}{n_{t_k-1, 1}} + \rad_{t_k}(n_{t_k-1, 1})\leq \mu_k + \frac{\sum_{\ell=1}^{n_k-1}\epsilon_{t_k(\ell), k}}{n_k-1} + \rad_{t_k}(n_k-1)\right) \\
    & \leq \mbP\brs{\frac{\sum_{\ell=1}^{n_k-1}\epsilon_{t_k(\ell), k}}{n_k-1} \geq \frac{\Delta_k}{2} - \rad_{t_k}(n_k-1)} + \mbP\brs{\frac{\sum_{\ell=1}^{n_{t_k-1, 1}}\epsilon_{t_1(\ell), 1}}{n_{t_k-1, 1}} \leq - \frac{\Delta_k}{2} - \rad_{t_k}(n_{t_k-1, 1})} \\
    & \leq \mbP\brs{\frac{\sum_{\ell=1}^{n_k-1}\epsilon_{t_k(\ell), k}}{n_k-1} \geq \frac{\Delta_k}{2} - \rad_{t_k}(n_k-1)} + \mbP\brs{\exists n: \frac{\sum_{\ell=1}^{n}\epsilon_{t_1(\ell), 1}}{n} \leq - \frac{\Delta_k}{2} - \rad_{t_k}(n)} \\
    & \leq \mbP\brs{\frac{\sum_{\ell=1}^{n_k-1}\epsilon_{t_k(\ell), k}}{n_k-1} \geq \frac{\Delta_k}{2} - \rad_{T}(n_k-1)} + \mbP\brs{\exists n: \frac{\sum_{\ell=1}^{n}\epsilon_{t_1(\ell), 1}}{n} \leq - \frac{\Delta_k}{2} - \rad_{x}(n)} \\
    & \triangleq \mbP\brs{\cE_k} + \mbP\brs{\cF_k}.
\end{align*}

We have the following bounds on the two tail probabilities.
\begin{align*}
    & \quad\; \mbP\brs{\cE_k} \\
    & \leq \mbP\brs{\exists n\geq m_k: \sum_{\ell=1}^{n}\epsilon_{t_k(\ell), k}\geq \frac{\Delta_k}{2}n - \sqrt{f(T)n}} \\
    & \leq \mbP\brs{\exists n\geq m_k: \sum_{\ell=1}^{n}\epsilon_{t_k(\ell), k}\geq \frac{\Delta_k}{2}n - \sqrt{\frac{f(T)}{m_k}}n} \\
    & \leq \mbP\brs{\exists n\geq 0: \sum_{\ell=\lceil m_k\rceil+1}^{\lceil m_k\rceil + n}\epsilon_{t_k(\ell), k}\geq \brs{\frac{\Delta_k}{2}-\sqrt{\frac{f(T)}{m_k}}}n + \frac{\lceil m_k\rceil}{4}\brs{\frac{\Delta_k}{2}-\sqrt{\frac{f(T)}{m_k}}}} + \\
    & \quad\quad \mbP\brs{\sum_{\ell=1}^{\lceil m_k\rceil}\epsilon_{t_k(\ell), k}\geq \frac{\lceil m_k\rceil}{4}\brs{\frac{\Delta_k}{2}-\sqrt{\frac{f(T)}{m_k}}}} \\
    & \leq 2\exp\brs{-2\frac{m_k}{4\sigma^2}\brs{\frac{\Delta_k}{2}-\sqrt{\frac{f(T)}{m_k}}}^2} + \exp\brs{-\frac{m_k}{32\sigma^2}\brs{\frac{\Delta_k}{2} - \sqrt{\frac{f(T)}{m_k}}}^2} \\
    & = 2\exp\brs{-\frac{(x-K)\Delta_0}{\Delta_k^2}\cdot\frac{\brs{\frac{\Delta_k}{2} - \sqrt{f(T)\frac{\Delta_k^2}{(x-K)\Delta_0}}}^2}{32\sigma^2}} \\
    & = 2\exp\brs{\frac{\brs{\frac{\sqrt{(x-K)\Delta_0}}{2} - \sqrt{f(T)}}^2}{32\sigma^2}} \\
    & \leq 2\exp\brs{-\frac{\brs{(x-K)\Delta_0-8f(T)}_+}{256\sigma^2}}.
\end{align*}
Meanwhile, 
\begin{align*}
    \mbP(\cF_k) & \leq \mbP\brs{\exists n: \frac{\sum_{\ell=1}^{n}\epsilon_{t_1(\ell), 1}}{n} \leq - \frac{\Delta_k}{2} - \eta\frac{(x/K)^\alpha\sqrt{\ln K}}{n}} + \\
    & \quad\quad \mbP\brs{\exists n: \frac{\sum_{\ell=1}^{n}\epsilon_{t_1(\ell), 1}}{n} \leq - \frac{\Delta_k}{2} - \sqrt{\frac{f(x)}{n}}} \\
    & \leq \exp\brs{-2\frac{\eta}{\sigma^2}\frac{\Delta_k}{2}(x/K)^\alpha\sqrt{\ln K}} + \sum_{n=1}^{\lfloor x\rfloor}\exp\brs{-\frac{f(x)}{2\sigma^2}} + \sum_{\lfloor x\rfloor +1}^T\exp\brs{-\frac{f(n)}{2\sigma^2}} \\
    & \leq \exp\brs{-\frac{\eta\Delta_kx^\alpha\sqrt{\ln K}}{\sigma^2K^{\alpha}}} + \int_{0}^T\exp\brs{-\frac{f(x\vee y)}{2\sigma^2}}dy.
\end{align*}

Combining all the equations above yields
\begin{align*}
    & \quad\; \mbP(R_\theta^\pi(T)\geq x) \\
    & \leq 2K\exp\brs{-\frac{\brs{(x-K)\Delta_0-8f(T)}_+}{256\sigma^2}} + \sum_{k:\Delta_k>0}\exp\brs{-\frac{\eta\Delta_kx^\alpha\sqrt{\ln K}}{\sigma^2K^{\alpha}}} + K\int_{0}^T\exp\brs{-\frac{f(x\vee y)}{2\sigma^2}}dy.
\end{align*}

$\hfill\Box$

\section{Proofs for Section \ref{sec:extension}}

We introduce the following lemma.

\begin{lemma} \label{lemma:large_deviation-subexp}
    Assume $\xi_1, \dots, \xi_t, \cdots\in \SE(\sigma^2, \nu)$ are i.i.d. sub-exponential random variables with zero mean (i.e. $\ex{}{\xi_1}=0$) Then for any $\lambda >0, B\geq 0$, we have 
    \begin{equation}
        \begin{aligned}
            \mbP\brs{\exists t\geq 1\text{ s.t. }\sum_{s=1}^t\xi_s \ge B+\lambda t} & \le \exp\brs{-2B\frac{\lambda}{\sigma^2}\wedge \frac{B}{\nu}},\\
            \mbP\brs{\exists t\geq 1\text{ s.t. }\sum_{s=1}^t\xi_s \le -B-\lambda t} & \le \exp\brs{-2B\frac{\lambda}{\sigma^2}\wedge \frac{B}{\nu}}.
        \end{aligned}
    \end{equation}
\end{lemma}

{\noindent\bf Proof of Theorem \ref{thm:upper-bound-fixed-subexp}.} Without loss of generality, we assume $\theta_1  = \theta_*$. Let $\pi = \UCB$ and $x\geq K$. We prove the bounds for two scenarios separately. To avoid repetition, We only highlight the difference compared to the proof of Theorem \ref{thm:upper-bound-fixed}: bounding $\mbP\brs{\cE_k}$ and $\mbP\brs{\cE_k}$

{\noindent\bf 1. Worst-case scenario.} We have the following bounds on the two tail probabilities.
\begin{align*}
    \mbP\brs{\cE_k} 
    & \leq \exp\brs{-\lceil m_k\rceil\frac{\brs{\frac{\Delta_k}{2} - \rad(\lceil m_k\rceil)}^2}{2\sigma^2}\wedge\lceil m_k\rceil\frac{\brs{\frac{\Delta_k}{2} - \rad(\lceil m_k\rceil)}_+}{2\nu}} \\
    & \leq \exp\brs{-m_k\frac{\brs{\frac{\Delta_k}{2} - \rad(m_k)}^2}{2\sigma^2}\wedge m_k\frac{\brs{\frac{\Delta_k}{2} - \rad(m_k)}_+}{2\nu}} \\
    & \leq \exp\brs{-\frac{x-K}{2K\Delta_k}\cdot\frac{\brs{\frac{\Delta_k}{2} - \eta\frac{(T/K)^\alpha\sqrt{\ln K}}{x-K}(2K\Delta_k)}^2}{2\sigma^2}\wedge\frac{\brs{\frac{x-K}{2K}-\eta (T/K)^\alpha\sqrt{\ln K}}_+}{2\nu}} \\
    & = \exp\brs{-\Delta_k\frac{(x-K)}{4K\sigma^2}\brs{\frac{1}{2}-\eta\frac{2K^{1-\alpha}T^{\alpha}\sqrt{\ln K}}{x-K}}^2\wedge \frac{\brs{x-K-2\eta K^{1-\alpha}T^{\alpha}\sqrt{\ln K}}_+}{4\nu K}} \\
    & \leq \exp\brs{-\frac{\brs{x-K-4\eta K^{1-\alpha}T^{\alpha}\sqrt{\ln K}}_+^2}{32\sigma^2KT}\wedge \frac{\brs{x-K-2\eta K^{1-\alpha}T^{\alpha}\sqrt{\ln K}}_+}{4\nu K}} \\
    & \leq \exp\brs{-\frac{\brs{x-K-4\eta K^{1-\alpha}T^{\alpha}\sqrt{\ln K}}_+^2}{(32\sigma^2\vee 4\nu)KT}}.
\end{align*}
Meanwhile, 
\begin{align*}
    \mbP(\cF_k) & \leq \mbP\brs{\exists n: \frac{\sum_{\ell=1}^{n}\epsilon_{t_1(\ell), 1}}{n} \leq - \frac{\Delta_k}{2} - \eta\frac{(T/K)^\alpha\sqrt{\ln K}}{n}, \ \frac{T}{K} \geq m_k} + \\
    & \quad\quad \mbP\brs{\exists n: \frac{\sum_{\ell=1}^{n}\epsilon_{t_1(\ell), 1}}{n} \leq - \frac{\Delta_k}{2} - \sqrt{\frac{f(T)}{n}} \vee \kappa\frac{f(T)}{n}, \ \frac{T}{K} \geq m_k} \\
    & \leq \exp\brs{-2\frac{\eta}{\sigma^2}\frac{x-K}{4T}(T/K)^\alpha\sqrt{\ln K}\wedge \eta\frac{(T/K)^\alpha\sqrt{\ln K}}{\nu}} + \sum_{n=1}^T\exp\brs{-\frac{f(T)}{2\sigma^2}\wedge\kappa\frac{f(T)}{2\nu}} \\
    & \leq \exp\brs{-\frac{\eta(x-K)_+\sqrt{\ln K}}{(2\sigma^2\vee\nu)K^{\alpha} T^{1-\alpha}}} + T\exp\brs{-\frac{f(T)}{2\sigma^2\vee 2\nu\kappa^{-1}}}.
\end{align*}

Note that the equations above hold for any instance $\theta$. Combining all the equations above yields
\begin{align*}
    & \quad\; \sup_{\theta}\mbP(R_\theta^\pi(T)\geq x) \\
    & \leq K\exp\left(\frac{\left(x-K-4\eta K^{1-\alpha}T^\alpha\sqrt{\ln K}\right)_+^2}{(32\sigma^2\vee 4\nu)KT}\right) + K\exp\brs{-\frac{\eta(x-K)_+\sqrt{\ln K}}{(2\sigma^2\vee\nu)K^{\alpha} T^{1-\alpha}}} + KT\exp\brs{-\frac{f(T)}{2\sigma^2\vee 2\nu\kappa^{-1}}}.
\end{align*}

{\noindent\bf 2. Instance-dependent scenario.} Let $x \geq K + (4\vee\kappa^2)\cdot f(T)\frac{1}{\Delta_0}$, then
\begin{align*}
    \kappa\frac{f(T)}{m_k} \leq \kappa \sqrt{\frac{f(T)}{m_k}} \cdot \sqrt{\frac{f(T)}{(x-K)\Delta_k}} \leq \sqrt{\frac{f(T)}{m_k}}.
\end{align*}
We have the following bounds on the two tail probabilities.
\begin{align*}
    \mbP\brs{\cE_k} 
    & \leq \exp\brs{-\lceil m_k\rceil\frac{\brs{\frac{\Delta_k}{2} - \rad(\lceil m_k\rceil)}^2}{2\sigma^2}\wedge \lceil m_k\rceil\frac{\brs{\frac{\Delta_k}{2} - \rad(\lceil m_k\rceil)}_+}{2\nu}} \\
    & \leq \exp\brs{-m_k\frac{\brs{\frac{\Delta_k}{2} - \rad(m_k)}^2}{2\sigma^2}\wedge m_k\frac{\brs{\frac{\Delta_k}{2} - \rad(m_k)}_+}{2\nu}} \\
    & = \exp\brs{-m_k\brs{\frac{\brs{\frac{\Delta_k}{2} - \sqrt{\frac{f(T)}{m_k}}}^2}{2\sigma^2}\wedge\frac{\brs{\frac{\Delta_k}{2} - \sqrt{\frac{f(T)}{m_k}}}_+}{2\nu}}} \\
    & = \exp\brs{-\frac{\brs{\frac{\sqrt{(x-K)\Delta_0}}{2} - \sqrt{f(T)}}^2}{2\sigma^2}\wedge \frac{\brs{\frac{(x-K)\Delta_0}{2\Delta_k}-\sqrt{f(T)}\frac{\sqrt{(x-K)\Delta_0}}{\Delta_k}}}{2\nu}} \\
    & \leq \exp\brs{-\frac{\brs{(x-K)\Delta_0/8-f(T)}_+}{2\sigma^2}\wedge \frac{\brs{\frac{(x-K)\Delta_0}{4}-f(T)}_+}{2\nu}} \\
    & \leq \exp\brs{-\frac{\brs{(x-K)\Delta_0-8f(T)}_+}{16\sigma^2\vee 8\nu}}.
\end{align*}
Meanwhile, 
\begin{align*}
    \mbP(\cF_k) & \leq \mbP\brs{\exists n: \frac{\sum_{\ell=1}^{n}\epsilon_{t_1(\ell), 1}}{n} \leq - \frac{\Delta_k}{2} - \eta\frac{(T/K)^\alpha\sqrt{\ln K}}{n}} + \\
    & \quad\quad \mbP\brs{\exists n: \frac{\sum_{\ell=1}^{n}\epsilon_{t_1(\ell), 1}}{n} \leq - \frac{\Delta_k}{2} - \sqrt{\frac{f(T)}{n}}\vee\kappa\frac{f(T)}{n}} \\
    & \leq \exp\brs{-2\frac{\eta}{\sigma^2}\frac{\Delta_k}{2}(T/K)^\alpha\sqrt{\ln K}\wedge \eta\frac{(T/K)^\alpha\sqrt{\ln K}}{\nu}} + \sum_{n=1}^T\exp\brs{-\frac{f(T)}{2\sigma^2}\wedge\kappa\frac{f(T)}{2\nu}} \\
    & \leq \exp\brs{-\frac{\eta\Delta_kT^\alpha\sqrt{\ln K}}{(\sigma^2\vee\nu)K^{\alpha}}} + T\exp\brs{-\frac{f(T)}{2\sigma^2\vee2\nu\kappa^{-1}}}.
\end{align*}

Combining all the equations above yields
\begin{align*}
    & \quad\; \mbP(R_\theta^\pi(T)\geq x) \\
    & \leq K\exp\brs{-\frac{\brs{(x-K)\Delta_0-(8\vee\kappa^2)f(T)}_+}{16\sigma^2\vee 8\nu}} + \sum_{k:\Delta_k>0}\exp\brs{-\frac{\eta\Delta_k(T/K)^\alpha\sqrt{\ln K}}{\sigma^2\vee\nu}} + \\
    & \quad\quad KT\exp\brs{-\frac{f(T)}{2\sigma^2\vee 2\nu\kappa^{-1}}}.
\end{align*}

$\hfill\Box$

{\noindent\bf Proof of Theorem \ref{thm:upper-bound-any-subexp}.} Without loss of generality, we assume $\theta_1  = \theta_*$. Let $\pi = \UCB$ and $x\geq K$. We prove the bounds for two scenarios separately. To avoid repetition, We only highlight the difference compared to the proof of Theorem \ref{thm:upper-bound-fixed}: bounding $\mbP\brs{\cE_k}$ and $\mbP\brs{\cF_k}$.

{\noindent\bf 1. Worst-case scenario.} We have the following bounds on the two tail probabilities.
\begin{align*}
    & \quad\; \mbP\brs{\cE_k} \\
    & \leq \mbP\brs{\exists n\geq m_k: \sum_{\ell=1}^{n}\epsilon_{t_k(\ell), k}\geq \frac{\Delta_k}{2}n - \eta(T/K)^{\alpha}\sqrt{\ln K}} \\
    & \leq \mbP\brs{\exists n\geq 0: \sum_{\ell=\lceil m_k\rceil+1}^{\lceil m_k\rceil + n}\epsilon_{t_k(\ell), k}\geq \frac{\Delta_k}{2}n + \frac{\lceil m_k\rceil\Delta_k}{4}} + \mbP\brs{\sum_{\ell=1}^{\lceil m_k\rceil}\epsilon_{t_k(\ell), k}\geq \frac{\lceil m_k\rceil\Delta_k}{4} -\frac{\eta(T/K)^{\alpha}\sqrt{\ln K}}{2}} \\
    & \leq \exp\brs{-2\frac{m_k\Delta_k^2}{8\sigma^2}\wedge\frac{m_k\Delta_k}{4\nu}} + \exp\brs{-\lceil m_k\rceil\frac{\brs{\frac{\Delta_k}{4} - \frac{\eta(T/K)^{\alpha}\sqrt{\ln K}}{2m_k}}^2}{2\sigma^2}\wedge \lceil m_k\rceil\frac{\brs{\frac{\Delta_k}{4} - \frac{\eta(T/K)^{\alpha}\sqrt{\ln K}}{2m_k}}_+}{2\nu}} \\
    & \leq \exp\brs{-m_k\frac{\brs{\frac{\Delta_k}{4} - \frac{\eta(T/K)^{\alpha}\sqrt{\ln K}}{2m_k}}^2}{2\sigma^2\vee 2\nu}\wedge\frac{\frac{m_k\Delta_k}{4}-\eta(T/K)^\alpha\sqrt{\ln K}}{2\nu}} \\
    & \leq 2\exp\brs{-c_{\alpha}\frac{x-K}{K\Delta_k}\cdot\frac{\brs{\frac{\Delta_k}{4} - \eta\frac{(T/K)^\alpha\sqrt{\ln K}}{2c_{\alpha}(x-K)}(2K\Delta_k)}^2}{2\sigma^2}\wedge \frac{\frac{c_\alpha(x-K)}{4K}-\eta(T/K)^\alpha\sqrt{\ln K}}{2\nu}} \\
    & = 2\exp\brs{-c_\alpha\Delta_k\frac{(x-K)}{2K\sigma^2}\brs{\frac{1}{4}-\eta\frac{K^{1-\alpha}T^{\alpha}\sqrt{\ln K}}{c_\alpha(x-K)}}^2\wedge \frac{c_\alpha(x-K)-4\eta K^{1-\alpha} T^{\alpha}\sqrt{\ln K}}{8\nu K}} \\
    & \leq 2\exp\brs{-\frac{\brs{c_\alpha(x-K)-4\eta K^{1-\alpha}T^{\alpha}\sqrt{\ln K}}_+^2}{(32\sigma^2\vee 8\nu)KT}}.
\end{align*}
Meanwhile, 
\begin{align*}
    \mbP(\cF_k) & \leq \mbP\brs{\exists n: \frac{\sum_{\ell=1}^{n}\epsilon_{t_1(\ell), 1}}{n} \leq - \frac{\Delta_k}{2} - \eta\frac{(t_k/K)^\alpha\sqrt{\ln K}}{n}} + \\
    & \quad\quad \mbP\brs{\exists n: \frac{\sum_{\ell=1}^{n}\epsilon_{t_1(\ell), 1}}{n} \leq - \frac{\Delta_k}{2} - \sqrt{\frac{f(t_k)}{n}}\vee \kappa\frac{f(t_k)}{n}} \\
    & \leq \exp\brs{-2\frac{\eta}{\sigma^2}\frac{\Delta_k}{2}(x/K)^\alpha\sqrt{\ln K}\wedge\eta\frac{(x/K)^\alpha\sqrt{\ln K}}{\nu}} + \sum_{n=1}^{\lfloor x\rfloor}\exp\brs{-\frac{f(x)}{2\sigma^2}\wedge \kappa\frac{f(x)}{2\nu}} + \nonumber\\
    & \quad\quad \sum_{n=\lfloor x\rfloor + 1}^T \exp\brs{-\frac{f(n)}{2\sigma^2}\wedge \kappa\frac{f(n)}{2\nu}} \\
    & \leq \exp\brs{-c_\alpha\frac{\eta x\sqrt{\ln K}}{(\sigma^2\vee\nu)K^{\alpha} T^{1-\alpha}}} + \int_{0}^T\exp\brs{-\frac{f(x\vee y)}{2\sigma^2\vee 2\nu\kappa^{-1}}}dy.
\end{align*}
Note that we have utilized the fact that $t_k \geq x$ and $t_k\geq n$.

Note that the equations above hold for any instance $\theta$. Combining all the equations above yields
\begin{align*}
    & \quad\sup_{\theta}\mbP(R_\theta^\pi(T)\geq x) \\
    & \leq K\exp\left(-\frac{\left(c_{\alpha}(x-K)-4\eta K^{1-\alpha}T^\alpha\sqrt{\ln K}\right)_+^2}{(32\sigma^2\vee 8\nu)KT}\right) + K\exp\brs{-\frac{c_\alpha\eta x\sqrt{\ln K}}{(\sigma^2\vee\nu)K^{\alpha} T^{1-\alpha}}} + \\
    & \quad\quad K\int_{0}^T\exp\brs{-\frac{f(x\vee y)}{2\sigma^2\vee 2\nu\kappa^{-1}}}dy.
\end{align*}

{\noindent\bf 2. Instance-dependent scenario.} Let $x \geq K + (4\vee\kappa^2)\cdot f(T)\frac{1}{\Delta_0}$, then for any $n\geq m_k$, 
\begin{align*}
    \kappa\frac{f(T)}{n} \leq \kappa \sqrt{\frac{f(T)}{n}} \cdot \sqrt{\frac{f(T)}{(x-K)\Delta_k}} \leq \sqrt{\frac{f(T)}{n}},
\end{align*}
\begin{align*}
    m_k = (x-K)\Delta_0/\Delta_k^2\leq n_k-1. 
\end{align*}
We have the following bounds on the two tail probabilities.
\begin{align*}
    & \quad\; \mbP\brs{\cE_k} \\
    & \leq \mbP\brs{\exists n\geq m_k: \sum_{\ell=1}^{n}\epsilon_{t_k(\ell), k}\geq \frac{\Delta_k}{2}n - \sqrt{f(T)n}} \\
    & \leq \mbP\brs{\exists n\geq m_k: \sum_{\ell=1}^{n}\epsilon_{t_k(\ell), k}\geq \frac{\Delta_k}{2}n - \sqrt{\frac{f(T)}{m_k}}n} \\
    & \leq \mbP\brs{\exists n\geq 0: \sum_{\ell=\lceil m_k\rceil+1}^{\lceil m_k\rceil + n}\epsilon_{t_k(\ell), k}\geq \brs{\frac{\Delta_k}{2}-\sqrt{\frac{f(T)}{m_k}}}n + \frac{\lceil m_k\rceil}{4}\brs{\frac{\Delta_k}{2}-\sqrt{\frac{f(T)}{m_k}}}} + \\
    & \quad\quad \mbP\brs{\sum_{\ell=1}^{\lceil m_k\rceil}\epsilon_{t_k(\ell), k}\geq \frac{\lceil m_k\rceil}{4}\brs{\frac{\Delta_k}{2}-\sqrt{\frac{f(T)}{m_k}}}} \\
    & \leq \exp\brs{-2\frac{m_k}{4\sigma^2}\brs{\frac{\Delta_k}{2}-\sqrt{\frac{f(T)}{m_k}}}^2 \wedge \frac{m_k}{4\nu}\brs{\frac{\Delta_k}{2}-\sqrt{\frac{f(T)}{m_k}}}} + \\
    & \quad\quad \exp\brs{-\frac{m_k}{32\sigma^2}\brs{\frac{\Delta_k}{2} - \sqrt{\frac{f(T)}{m_k}}}^2\wedge \frac{m_k}{8\nu}\brs{\frac{\Delta_k}{2}-\sqrt{\frac{f(T)}{m_k}}}} \\
    & = 2\exp\brs{-\frac{(x-K)\Delta_0}{\Delta_k^2}\cdot\frac{\brs{\frac{\Delta_k}{2} - \sqrt{f(T)\frac{\Delta_k^2}{(x-K)\Delta_0}}}^2}{32\sigma^2}\wedge \frac{(x-K)\Delta_0}{\Delta_k^2}\cdot\frac{\brs{\frac{\Delta_k}{2} - \sqrt{f(T)\frac{\Delta_k^2}{(x-K)\Delta_0}}}_+}{8\nu}} \\
    & = 2\exp\brs{\frac{\brs{\frac{\sqrt{(x-K)\Delta_0}}{2} - \sqrt{f(T)}}^2}{32\sigma^2}\wedge\frac{\brs{\frac{(x-K)\Delta_0}{2} - \sqrt{(x-K)\Delta_0f(T)}}_+}{8\nu}} \\
    & \leq 2\exp\brs{-\frac{\brs{(x-K)\Delta_0-8f(T)}_+}{256\sigma^2\vee 32\nu}}.
\end{align*}
Meanwhile, 
\begin{align*}
    \mbP(\cF_k) & \leq \mbP\brs{\exists n: \frac{\sum_{\ell=1}^{n}\epsilon_{t_1(\ell), 1}}{n} \leq - \frac{\Delta_k}{2} - \eta\frac{(x/K)^\alpha\sqrt{\ln K}}{n}} + \\
    & \quad\quad \mbP\brs{\exists n: \frac{\sum_{\ell=1}^{n}\epsilon_{t_1(\ell), 1}}{n} \leq - \frac{\Delta_k}{2} - \sqrt{\frac{f(x)}{n}}} \\
    & \leq \exp\brs{-2\frac{\eta}{\sigma^2}\frac{\Delta_k}{2}(x/K)^\alpha\sqrt{\ln K}\wedge \eta\frac{(x/K)^\alpha\sqrt{\ln K}}{\nu}} + \sum_{n=1}^{\lfloor x\rfloor}\exp\brs{-\frac{f(x)}{2\sigma^2}\wedge \kappa\frac{f(x)}{2\nu}} + \nonumber\\
    & \quad\quad \sum_{n=\lfloor x\rfloor + 1}^T \exp\brs{-\frac{f(n)}{2\sigma^2}\wedge \kappa\frac{f(n)}{2\nu}} \\
    & \leq \exp\brs{-\frac{\eta\Delta_k(x/K)^\alpha\sqrt{\ln K}}{\sigma^2\vee\nu}} + T\exp\brs{-\frac{f(x)}{2\sigma^2\vee2\nu\kappa^{-1}}} + \int_{0}^T\exp\brs{-\frac{f(x\vee y)}{2\sigma^2\vee 2\nu\kappa^{-1}}}dy.
\end{align*}

Combining all the equations above yields
\begin{align*}
    & \quad\; \mbP(R_\theta^\pi(T)\geq x) \\
    & \leq 2K\exp\brs{-\frac{\brs{(x-K)\Delta_0-(8\vee\kappa^2)f(T)}_+}{256\sigma^2\vee 32\nu}} + \sum_{k:\Delta_k>0}\exp\brs{-\frac{\eta\Delta_k(x/K)^\alpha\sqrt{\ln K}}{\sigma^2\vee\nu}} + \\
    & \quad\quad K\int_{0}^T\exp\brs{-\frac{f(x\vee y)}{2\sigma^2\vee 2\nu\kappa^{-1}}}dy.
\end{align*}

$\hfill\Box$

{\noindent\bf Proof of Theorem \ref{thm:linear}.} To simplify notations, we write $\Delta_t\triangleq \theta^\top(a_t^*-a_t)\in[0, 1]$. Also, we write
\begin{align*}
    A_t = [a_1, \cdots, a_t], \quad R_t = [r_1, \cdots, r_t]^\top, \quad \mathcal E_t = [\epsilon_{1, a_1}, \cdots, \epsilon_{t, a_t}]^\top.
\end{align*}
Meanwhile, for any vector $z$, denote
\begin{align*}
    \|z\|_t = \|z\|_{V_{t-1}^{-1}} = \sqrt{z^\top V_{t-1}^{-1}z}.
\end{align*}
Then
\begin{align*}
    \hat\theta_t & = V_t^{-1}A_tR_t = V_t^{-1}A_t(A_t^\top\theta+\mathcal E_t) = \theta - V_t^{-1}\theta + V_t^{-1}A_t\mathcal E_t.
\end{align*}
Note that
\begin{align*}
    R_\theta^\pi(T) = \sum_{t}\Delta_t = \sum_{t}\frac{\Delta_t}{\|a_t\|_t^2}\cdot \|a_t\|_t^2
\end{align*}
and from Lemma 11 in \cite{abbasi2011improved}, 
\begin{align*}
    \sum_{t}\|a_t\|_t^2 \leq 2\ln\det V_{T-1} - 2\ln\det V_1 \leq 2d\ln\left(\frac{tr(V_{T-1})}{d}\right) \leq 2d\ln\frac{T}{d} \leq 2d\ln T.
\end{align*}
Another fact we will be using in the proof is from Theorem 1 in \cite{abbasi2011improved}, where it is shown that for any $\delta>0$, w.p. at least $1-\delta$, the following holds:
\begin{align*}
    (A_{t-1}\mathcal E_{t-1})^\top  V_{t-1}^{-1}A_{t-1}\mathcal E_{t-1} \leq 2\sigma^2\log\left(\frac{\det(V_{t-1})/\det(V_0)}{\delta}\right) \leq 2\sigma^2\log\left(\frac{(T/d)^{2d}}{\delta}\right).
\end{align*}
Thus, for any $y\geq 0$, we have
\begin{align*}
    \mbP\left(\sqrt{(A_{t-1}\mathcal E_{t-1})^\top  V_{t-1}^{-1}A_{t-1}\mathcal E_{t-1}} \geq x\right) \leq (T/d)^{2d}\exp\left(-\frac{x^2}{2\sigma^2}\right).
\end{align*}
We have, for any $\theta$, 
\begin{align} \label{eq:linear-decompose}
    \mbP(R_\theta^\pi(T) \geq x) \leq \mbP\left(\bigcup_{t\geq 2}\left\{\Delta_t \geq \frac{x-1}{2T},\ \|a_t\|_t^2\leq \frac{d}{T} \right\}\right) + \mbP\left(\bigcup_{t\geq 2}\left\{\frac{\Delta_t}{\|a_t\|_t^2}\geq \frac{x-1}{4d\ln T}, \ \|a_t\|_t^2 > \frac{d}{T}\right\}\right).
\end{align}
The reason that \eqref{eq:linear-decompose} holds is as follows. To prove it, we only need to show that the following events cannot hold simultaneously:
\begin{align*}
    \Delta_t < \frac{x-1}{2T},\ \text{ if }\|a_t\|_t^2\leq d/T; \quad \frac{\Delta_t}{\|a_t\|_t^2} < \frac{x-1}{4d\ln T}, \ \text{ if }\|a_t\|_t^2 > d/T.
\end{align*}
If not, then
\begin{align*}
    R_{\theta}^\pi(T) & = \theta^\top(a_1^*-a_1) + \sum_{t\geq 2} \Delta_t\mathds 1\{\|a_t\|_t^2\leq \frac{d}{T}\} + \frac{\Delta_t}{\|a_t\|_t^2}\cdot \|a_t\|_t^2\mathds 1\{\|a_t\|_t^2 > \frac{d}{T}\} \\
    & < 1 + \sum_{t\geq 2}\frac{x-1}{2T} + \sum_{t\geq 2}\frac{x-1}{4d\ln T}\|a_t\|_t^2 \\
    & \leq 1 + \frac{x-1}{2} + \frac{x-1}{2} = x.
\end{align*}
This is a contradiction.
At time $t$, the policy takes action $a_t$, which means
\begin{align}
    & \quad\ \quad\ \hat\theta_{t-1}^\top a_t + \eta(T/d)^\alpha\sqrt{d}\|a_t\|_t^2\wedge\sqrt{f(T)}\|a_t\|_t + \sqrt{d\|a_t\|_t^2} \geq \nonumber\\
    & \quad\quad\quad \hat\theta_{t-1}^\top a_t^* + \eta(T/d)^\alpha\sqrt{d}\|a_t^*\|_t^2\wedge\sqrt{f(T)}\|a_t^*\|_t + \sqrt{d\|a_t^*\|_t^2} \nonumber\\
    & \Leftrightarrow\quad \theta^\top a_t - \theta^\top V_{t-1}^{-1}a_t + (V_{t-1}^{-1}A_{t-1}\mathcal E_{t-1})^\top a_t + \eta(T/d)^\alpha\sqrt{d}\|a_t\|_t^2\wedge\sqrt{f(T)}\|a_t\|_t + \sqrt{d\|a_t\|_t^2} \geq \nonumber\\
    & \quad\quad\quad \theta^\top a_t^* - \theta^\top V_{t-1}^{-1}a_t^* + (V_{t-1}^{-1}A_{t-1}\mathcal E_{t-1})^\top a_t^* + \eta(T/d)^\alpha\sqrt{d}\|a_t^*\|_t^2\wedge\sqrt{f(T)}\|a_t^*\|_t + \sqrt{d\|a_t^*\|_t^2} \nonumber\\
    & \Leftrightarrow\quad (V_{t-1}^{-1}A_{t-1}\mathcal E_{t-1})^\top a_t + \eta(T/d)^\alpha\sqrt{d}\|a_t\|_t^2\wedge\sqrt{f(T)}\|a_t\|_t + \sqrt{d\|a_t\|_t^2} - \theta^\top V_{t-1}^{-1}a_t \geq \nonumber\\
    & \quad\quad\quad \Delta_t + (V_{t-1}^{-1}A_{t-1}\mathcal E_{t-1})^\top a_t^* + \eta(T/d)^\alpha\sqrt{d}\|a_t^*\|_t^2\wedge\sqrt{f(T)}\|a_t^*\|_t + \sqrt{d\|a_t^*\|_t^2} - \theta^\top V_{t-1}^{-1}a_t^* \nonumber\\
    & \Rightarrow\quad a_t^\top V_{t-1}^{-1}A_{t-1}\mathcal E_{t-1} + \eta(T/d)^\alpha\sqrt{d}\|a_t\|_t^2\wedge\sqrt{f(T)}\|a_t\|_t + 2\sqrt{d\|a_t\|_t^2} \geq \nonumber\\
    & \quad\quad\quad \Delta_t + a_t^{*\top}V_{t-1}^{-1}A_{t-1}\mathcal E_{t-1} + \eta(T/d)^\alpha\sqrt{d}\|a_t^*\|_t^2\wedge\sqrt{f(T)}\|a_t^*\|_t \nonumber\\
    & \Rightarrow\quad a_t^\top V_{t-1}^{-1}A_{t-1}\mathcal E_{t-1} \geq \frac{\Delta_t}{2} - \eta(T/d)^\alpha\sqrt{d}\|a_t\|_t^2\wedge\sqrt{f(T)}\|a_t\|_t - 2\sqrt{d\|a_t\|_t^2} \quad \text{or} \nonumber\\
    & \quad\quad\quad -a_t^{*\top}V_{t-1}^{-1}A_{t-1}\mathcal E_{t-1} \geq \frac{\Delta_t}{2} + \eta(T/d)^\alpha\sqrt{d}\|a_t^*\|_t^2\wedge\sqrt{f(T)}\|a_t^*\|_t. \label{eq:linear-action}
\end{align}
Note that in \eqref{eq:linear-action} we use the following inequality: for any $a\in\cA_t$, 
\begin{align*}
    |\theta^\top V_{t-1}^{-1}a| \leq \sqrt{\theta^\top V_{t-1}^{-1}\theta}\sqrt{a^\top V_{t-1}^{-1}a} \leq \sqrt{d(a^\top V_{t-1}^{-1}a)}.
\end{align*}
Combining \eqref{eq:linear-decompose} and \eqref{eq:linear-action} yields {\footnotesize
\begin{align*}
    & \quad\; \mbP\left(R_\theta^\pi(T) \geq x/2\right) \\
    & \leq \sum_{t}\mbP\left(\Delta_t \geq \frac{x-1}{2T},\ \|a_t\|_t^2\leq \frac{d}{T},\ a_t^\top V_{t-1}^{-1}A_{t-1}\mathcal E_{t-1} \geq \frac{\Delta_t}{2} - \eta(T/d)^\alpha\sqrt{d}\|a_t\|_t^2\wedge\sqrt{f(T)}\|a_t\|_t - 2\sqrt{d\|a_t\|_t^2}\right) \\
    & + \sum_t\mbP\left(\Delta_t \geq \frac{x-1}{2T},\ \|a_t\|_t^2\leq \frac{d}{T},\ -a_t^{*\top}V_{t-1}^{-1}A_{t-1}\mathcal E_{t-1} \geq \frac{\Delta_t}{2} + \eta(T/d)^\alpha\sqrt{d}\|a_t^*\|_t^2\wedge\sqrt{f(T)}\|a_t^*\|_t\right) \\
    & + \sum_{t}\mbP\left(\frac{\Delta_t}{\|a_t\|_t^2}\geq \frac{x-1}{4d\ln T},\ \|a_t\|_t^2 > \frac{d}{T},  \ a_t^\top V_{t-1}^{-1}A_{t-1}\mathcal E_{t-1} \geq \frac{\Delta_t}{2} - \eta(T/d)^\alpha\sqrt{d}\|a_t\|_t^2\wedge\sqrt{f(T)}\|a_t\|_t - 2\sqrt{d\|a_t\|_t^2}\right) \\
    & + \sum_t\mbP\left(\frac{\Delta_t}{\|a_t\|_t^2}\geq \frac{x-1}{4d\ln T},\ \|a_t\|_t^2 > \frac{d}{T},\ -a_t^{*\top}V_{t-1}^{-1}A_{t-1}\mathcal E_{t-1} \geq \frac{\Delta_t}{2} + \eta(T/d)^\alpha\sqrt{d}\|a_t^*\|_t^2\wedge\sqrt{f(T)}\|a_t^*\|_t\right).
\end{align*}}
We bound each term separately. 

{\noindent\bf 1. Worst-case scenario.} We have
\begin{align*}
    & \quad\; \mbP\left(\Delta_t \geq \frac{x-1}{2T},\ \|a_t\|_t^2\leq \frac{d}{T},\ a_t^\top V_{t-1}^{-1}A_{t-1}\mathcal E_{t-1} \geq \frac{\Delta_t}{2} - \eta(T/d)^\alpha\sqrt{d}\|a_t\|_t^2\wedge\sqrt{f(T)}\|a_t\|_t - 2\sqrt{d\|a_t\|_t^2}\right) \\
    & \leq \mbP\left(\|a_t\|_t^2\leq \frac{d}{T},\ a_t^\top V_{t-1}^{-1}A_{t-1}\mathcal E_{t-1} \geq \frac{x-1}{4T} - \eta(d/T)^{1-\alpha}(T/d)^\alpha - 2d/\sqrt{T}\right) \\
    & \leq \mbP\left(\frac{|a_t^\top V_{t-1}^{-1}A_{t-1}\mathcal E_{t-1}|}{\sqrt{\|a_t\|_t^2}}\geq \frac{\left(\frac{x-1}{4T} - \eta(d/T)^{1-\alpha}\sqrt{d}-2d/\sqrt{T}\right)_+}{\sqrt{d/T}\ln T}\right) \\
    & \leq \mbP\left(\sqrt{(A_{t-1}\mathcal E_{t-1})^\top  V_{t-1}^{-1}A_{t-1}\mathcal E_{t-1}} \geq \frac{\left(x-1 - 8d\sqrt{T}\ln T - 4\eta d^{\frac{3}{2}-\alpha}T^\alpha\ln T\right)_+}{4\sqrt{dT}}\right) \\
    & \leq (T/d)^{2d}\exp\left(-\frac{\left(x-1 - 8d\sqrt{T}\ln T - 4\eta d^{\frac{3}{2}-\alpha}T^\alpha\ln T\right)_+^2}{32\sigma^2dT\ln^2T}\right)
\end{align*}
and
\begin{align*}
    & \quad\; \mbP\left(\Delta_t \geq \frac{x-1}{2T},\ \|a_t\|_t^2\leq \frac{d}{T},\ -a_t^{*\top}V_{t-1}^{-1}A_{t-1}\mathcal E_{t-1} \geq \frac{\Delta_t}{2} + \eta(T/d)^\alpha\sqrt{d}\|a_t^*\|_t^2\wedge\sqrt{f(T)}\|a_t^*\|_t\right) \\
    & \leq \mbP\left(t\geq \frac{x-1}{2\Delta_t\ln T},\ -a_t^{*\top}V_{t-1}^{-1}A_{t-1}\mathcal E_{t-1} \geq \sqrt{2\frac{x-1}{2T}\eta(T/d)^\alpha\sqrt{d}\|a_t^*\|_t^2}\wedge\sqrt{f(T)}\|a_t^*\|_t\right) \\
    & \leq \mbP\left(\frac{|a_t^{*\top} V_{t-1}^{-1}A_{t-1}\mathcal E_{t-1}|}{\sqrt{\|a_t^*\|_t^2}} \geq \sqrt{\frac{(x-1)_+\eta\sqrt{d}}{2d^{\alpha}T^{1-\alpha}\ln T}}\wedge\sqrt{f(T)}\right) \\
    & \leq \mbP\left(\sqrt{(A_{t-1}\mathcal E_{t-1})^\top  V_{t-1}^{-1}A_{t-1}\mathcal E_{t-1}} \geq \frac{\sqrt{(x-1)_+\eta\sqrt{d}}}{\sqrt{2d^{\alpha}T^{1-\alpha}\ln T}}\wedge\sqrt{f(T)}\right) \\
    & \leq (T/d)^{2d}\exp\left(-\frac{\eta(x-1)_+}{4\sigma^2 d^{\alpha-\frac{1}{2}}T^{1-\alpha}\ln T}\wedge\frac{f(T)}{2\sigma^2}\right)
\end{align*}
and
\begin{align*}
    & \quad\; \mbP\left(\frac{\Delta_t}{\|a_t\|_t^2}\geq \frac{x-1}{4d\ln T}, \ \|a_t\|_t^2 > \frac{d}{T}, \ a_t^\top V_{t-1}^{-1}A_{t-1}\mathcal E_{t-1} \geq \frac{\Delta_t}{2} - \eta(T/d)^\alpha\sqrt{d}\|a_t\|_t^2\wedge\sqrt{f(T)}\|a_t\|_t - 2\sqrt{d\|a_t\|_t^2}\right) \\
    & = \mbP\left(\frac{\Delta_t}{\|a_t\|_t^2}\geq \frac{x-1}{4d\ln T}, \ \|a_t\|_t^2 > \frac{d}{T}, \ \frac{a_t^\top V_{t-1}^{-1}A_{t-1}\mathcal E_{t-1}}{\sqrt{\|a_t\|_t^2}\sqrt{\|a_t\|_t^2}} \geq \frac{\Delta_t}{2\|a_t\|_t^2} - \eta(T/d)^\alpha\sqrt{d} - 2\sqrt{t}\right) \\
    & \leq \mbP\left(\|a_t\|_t^2 > \frac{d}{T}, \ \frac{a_t^\top V_{t-1}^{-1}A_{t-1}\mathcal E_{t-1}}{\sqrt{\|a_t\|_t^2}\sqrt{\|a_t\|_t^2}} \geq \frac{x-1}{8d\ln T} - \eta(T/d)^\alpha\sqrt{d} - 2\sqrt{t}\right) \\
    & \leq \mbP\left(\frac{|a_t^\top V_{t-1}^{-1}A_{t-1}\mathcal E_{t-1}|}{\sqrt{\|a_t\|_t^2}} \geq \left(\frac{x-1}{8d\ln T} - \eta(T/d)^\alpha\sqrt{d} - 2\sqrt{t}\right)_+\sqrt{\frac{d}{T}}\right) \\
    & \leq \mbP\left(\sqrt{(A_{t-1}\mathcal E_{t-1})^\top  V_{t-1}^{-1}A_{t-1}\mathcal E_{t-1}} \geq \frac{\left(x-1-16d\sqrt{T}\ln T - 8\eta d^{\frac{3}{2}-\alpha}T^\alpha\ln T\right)_+}{8\sqrt{dT}\ln T}\right) \\
    & \leq (T/d)^{2d}\exp\left(-\frac{\left(x-1-16d\sqrt{T}\ln T - 8\eta d^{\frac{3}{2}-\alpha}T^\alpha\ln T\right)_+^2}{128\sigma^2dT\ln^2T}\right)
\end{align*}
and
\begin{align*}
    & \quad\; \mbP\left(\frac{\Delta_t}{\|a_t\|_t^2}\geq \frac{x-1}{4d\ln T},\ \|a_t\|_t^2 > \frac{d}{T}, -a_t^{*\top}V_{t-1}^{-1}A_{t-1}\mathcal E_{t-1}\geq \frac{\Delta_t}{2} + \eta(T/d)^\alpha\sqrt{d}\|a_t^*\|_t^2\wedge\sqrt{f(T)}\|a_t^*\|_t\right) \\
    & \leq \mbP\left(-a_t^{*\top}V_{t-1}^{-1}A_{t-1}\mathcal E_{t-1}\geq \sqrt{2\frac{x-1}{4T}\eta(T/d)^\alpha\sqrt{d}\|a_t^*\|_t^2}\wedge\sqrt{f(T)}\|a_t^*\|_t\right) \\
    & \leq \mbP\left(\frac{|a_t^{*\top} V_{t-1}^{-1}A_{t-1}\mathcal E_{t-1}|}{\|a_t^*\|_t} \geq \sqrt{\frac{(x-1)\eta}{2d^\alpha T^{1-\alpha}\ln T}}\wedge \sqrt{f(T)}\right) \\
    & \leq \mbP\left(\sqrt{(A_{t-1}\mathcal E_{t-1})^\top  V_{t-1}^{-1}A_{t-1}\mathcal E_{t-1}} \geq \frac{\sqrt{(x-1)_+\eta\sqrt{d}}}{\sqrt{2d^\alpha T^{1-\alpha}\ln T}}\wedge\sqrt{f(T)}\right) \\
    & \leq (T/d)^{2d}\exp\left(-\frac{\eta(x-1)_+}{4\sigma^2 d^{\alpha-\frac{1}{2}} T^{1-\alpha}\ln T}\wedge\frac{f(T)}{2\sigma^2}\right).
\end{align*}

Plugging the four bounds above into \eqref{eq:linear-decompose} yields the final result
\begin{align*}
    \sup_{\theta}\mbP(R_{\theta}^\pi(T)\geq x) & \leq 2d(T/d)^{2d+1}\exp\left(-\frac{\left(x-1-16d\sqrt{T}\ln T - 8\eta d^{\frac{3}{2}-\alpha}T^\alpha\ln T\right)_+^2}{128\sigma^2dT\ln^2T}\right) \\
    & \quad\quad + 2d(T/d)^{2d+1}\exp\left(-\frac{\eta(x-1)_+}{4\sigma^2 d^{\alpha-\frac{1}{2}}T^{1-\alpha}\ln T}\right) + 2(T/d)^{2d}\int_{0}^T\exp\brs{-\frac{f(x\vee y)}{2\sigma^2}}dy.
\end{align*}

{\noindent\bf 2. Instance-dependent scenario.} We have
\begin{align*}
    & \quad\; \mbP\left(\Delta_t \geq \frac{x-1}{2T},\ \|a_t\|_t^2\leq \frac{d}{T},\ a_t^\top V_{t-1}^{-1}A_{t-1}\mathcal E_{t-1} \geq \frac{\Delta_t}{2} - \eta(T/d)^\alpha\sqrt{d}\|a_t\|_t^2\wedge\sqrt{f(T)}\|a_t\|_t - 2\sqrt{d\|a_t\|_t^2}\right) \\
    & \leq \mbP\left(\Delta_t \geq \frac{x-1}{2T},\ \|a_t\|_t^2\leq \frac{d}{T},\ \frac{a_t^\top V_{t-1}^{-1}A_{t-1}\mathcal E_{t-1}}{\|a_t\|_t} \geq \frac{1}{2}\sqrt{\Delta_t}\cdot\sqrt{\frac{\Delta_t}{\|a_t\|_t^2}} - \sqrt{f(T)} - 2\sqrt{d}\right) \\
    & \leq \mbP\left(\frac{|a_t^\top V_{t-1}^{-1}A_{t-1}\mathcal E_{t-1}|}{\sqrt{\|a_t\|_t^2}}\geq \left(\frac{1}{2}\sqrt{\Delta}\cdot\sqrt{\frac{x-1}{2d\ln T}} - \sqrt{f(T)} - 2\sqrt{d}\right)_+\right) \\
    & \leq \mbP\left(\sqrt{(A_{t-1}\mathcal E_{t-1})^\top  V_{t-1}^{-1}A_{t-1}\mathcal E_{t-1}} \geq \frac{\sqrt{\Delta(x-1)} - 2\sqrt{2}\sqrt{f(T)} - 4\sqrt{2}\sqrt{d}}{2\sqrt{2d\ln T}}\right) \\
    & \leq (T/d)^{2d}\exp\left(-\frac{\left(\Delta(x-1)/4 - 64d - 16f(T)\right)_+}{16\sigma^2d\ln T}\right)
\end{align*}
and
\begin{align*}
    & \quad\; \mbP\left(\Delta_t \geq \frac{x-1}{2T},\ \|a_t\|_t^2\leq \frac{d}{T},\ -a_t^{*\top}V_{t-1}^{-1}A_{t-1}\mathcal E_{t-1} \geq \frac{\Delta_t}{2} + \eta(T/d)^\alpha\sqrt{d}\|a_t^*\|_t^2\wedge\sqrt{f(T)}\|a_t^*\|_t\right) \\
    & \leq \mbP\left(-a_t^{*\top}V_{t-1}^{-1}A_{t-1}\mathcal E_{t-1} \geq \sqrt{\Delta\eta(T/d)^\alpha\sqrt{d}\|a_t^*\|_t^2}\wedge\sqrt{f(T)}\|a_t^*\|_t\right) \\
    & \leq \mbP\left(\frac{|a_t^{*\top} V_{t-1}^{-1}A_{t-1}\mathcal E_{t-1}|}{\sqrt{\|a_t^*\|_t^2}} \geq \sqrt{\Delta\eta(T/d)^\alpha\sqrt{d}}\wedge\sqrt{f(T)}\right) \\
    & \leq \mbP\left(\sqrt{(A_{t-1}\mathcal E_{t-1})^\top  V_{t-1}^{-1}A_{t-1}\mathcal E_{t-1}} \geq \frac{\sqrt{\eta\Delta T^\alpha\sqrt{d}}}{d^{\frac{\alpha}{2}}}\wedge\sqrt{f(T)}\right) \\
    & \leq (T/d)^{2d}\exp\left(-\frac{\eta\Delta T^\alpha}{2\sigma^2 d^{\alpha-\frac{1}{2}}}\wedge\frac{f(T)}{2\sigma^2}\right)
\end{align*}
and
\begin{align*}
    & \quad\; \mbP\left(\frac{\Delta_t}{\|a_t\|_t^2}\geq \frac{x-1}{4d\ln T}, \ \|a_t\|_t^2 > \frac{d}{T}, \ a_t^\top V_{t-1}^{-1}A_{t-1}\mathcal E_{t-1} \geq \frac{\Delta_t}{2} - \eta(T/d)^\alpha\sqrt{d}\|a_t\|_t^2\wedge\sqrt{f(T)}\|a_t\|_t - 2\sqrt{d\|a_t\|_t^2}\right) \\
    & \leq \mbP\left(\frac{\Delta_t}{\|a_t\|_t^2}\geq \frac{x-1}{4d\ln T},\ \|a_t\|_t^2>\frac{d}{T},\ \frac{a_t^\top V_{t-1}^{-1}A_{t-1}\mathcal E_{t-1}}{\|a_t\|_t} \geq \frac{1}{2}\sqrt{\Delta_t}\cdot\sqrt{\frac{\Delta_t}{\|a_t\|_t^2}} - \sqrt{f(T)} - 2\sqrt{d}\right) \\
    & \leq \mbP\left(\frac{|a_t^\top V_{t-1}^{-1}A_{t-1}\mathcal E_{t-1}|}{\sqrt{\|a_t\|_t^2}}\geq \left(\frac{1}{2}\sqrt{\Delta}\cdot\sqrt{\frac{x-1}{4d\ln T}} - \sqrt{f(T)} - 2\sqrt{d}\right)_+\right) \\
    & \leq \mbP\left(\sqrt{(A_{t-1}\mathcal E_{t-1})^\top  V_{t-1}^{-1}A_{t-1}\mathcal E_{t-1}} \geq \frac{\sqrt{\Delta(x-1)} - 4\sqrt{f(T)} - 8\sqrt{d}}{4\sqrt{d\ln T}}\right) \\
    & \leq (T/d)^{2d}\exp\left(-\frac{\left(\Delta(x-1)/4 - 128d - 32f(T)\right)_+}{32\sigma^2d\ln T}\right)
\end{align*}
and
\begin{align*}
    & \quad\; \mbP\left(\frac{\Delta_t}{\|a_t\|_t^2}\geq \frac{x-1}{4d\ln T},\ \|a_t\|_t^2 > \frac{d}{T}, -a_t^{*\top}V_{t-1}^{-1}A_{t-1}\mathcal E_{t-1}\geq \frac{\Delta_t}{2} + \eta(T/d)^\alpha\sqrt{d}\|a_t^*\|_t^2\wedge\sqrt{f(T)}\|a_t^*\|_t\right) \\
    & \leq \mbP\left(-a_t^{*\top}V_{t-1}^{-1}A_{t-1}\mathcal E_{t-1}\geq \sqrt{\Delta_t\eta(T/d)^\alpha\sqrt{d}\|a_t^*\|_t^2}\wedge\sqrt{f(T)}\|a_t^*\|_t, \ \Delta_t\geq \frac{x-1}{4T}\right) \\
    & \leq \mbP\left(\frac{|a_t^{*\top} V_{t-1}^{-1}A_{t-1}\mathcal E_{t-1}|}{\sqrt{\|a_t^*\|_t^2}} \geq \sqrt{\Delta\eta(T/d)^\alpha\sqrt{d}}\wedge\sqrt{f(T)}, \ t\geq \frac{x-1}{8\sqrt{d}\ln T}\right) \\
    & \leq \mbP\left(\sqrt{(A_{t-1}\mathcal E_{t-1})^\top  V_{t-1}^{-1}A_{t-1}\mathcal E_{t-1}} \geq \frac{\sqrt{\eta\Delta T^\alpha\sqrt{d}}}{(8d\sqrt{d}\ln T)^{\frac{\alpha}{2}}}\wedge\sqrt{f(T)}\right) \\
    & \leq (T/d)^{2d}\exp\left(-\frac{\eta\Delta T^\alpha}{2\sigma^2 d^{\alpha-\frac{1}{2}}}\wedge\frac{f(T)}{2\sigma^2}\right).
\end{align*}

Plugging the four bounds above into \eqref{eq:linear-decompose} yields the final result
\begin{align*}
    \sup_{\theta}\mbP(R_{\theta}^\pi(T)\geq x) & \leq 2d(T/d)^{2d+1}\exp\left(-\frac{\left(\Delta(x-1)/4 - 128d - 32f(T)\right)_+}{32\sigma^2d\ln T}\right) \\
    & \quad\quad + 2d(T/d)^{2d+1}\exp\left(-\frac{\eta\Delta T^\alpha}{2\sigma^2 d^{\alpha-\frac{1}{2}}}\wedge\frac{f(T)}{2\sigma^2}\right)
\end{align*}

$\hfill\Box$

{\noindent\bf Proof of Theorem \ref{thm:linear-any}.} The proof follows similarly to that of Theorem \ref{thm:linear}. For completeness, we present the proof in detail as follows. To simplify notations, we write $\Delta_t\triangleq \theta^\top(a_t^*-a_t)\in[0, 1]$. Also, we write
\begin{align*}
    A_t = [a_1, \cdots, a_t], \quad R_t = [r_1, \cdots, r_t]^\top, \quad \mathcal E_t = [\epsilon_{1, a_1}, \cdots, \epsilon_{t, a_t}]^\top.
\end{align*}
Meanwhile, for any vector $z$, denote
\begin{align*}
    \|z\|_t = \|z\|_{V_{t-1}^{-1}} = \sqrt{z^\top V_{t-1}^{-1}z}.
\end{align*}
Then
\begin{align*}
    \hat\theta_t & = V_t^{-1}A_tR_t = V_t^{-1}A_t(A_t^\top\theta+\mathcal E_t) = \theta - V_t^{-1}\theta + V_t^{-1}A_t\mathcal E_t.
\end{align*}
Note that
\begin{align*}
    R_\theta^\pi(T) = \sum_{t}\Delta_t = \sum_{t}\frac{\Delta_t}{\|a_t\|_t^2}\cdot \|a_t\|_t^2.
\end{align*}
and from Lemma 11 in \cite{abbasi2011improved}, 
\begin{align*}
    \sum_{t}\|a_t\|_t^2 \leq 2\ln\det V_{T-1} - 2\ln\det V_1 \leq 2d\ln\left(\frac{tr(V_{T-1})}{d}\right) \leq 2d\ln\frac{T}{d} \leq 2d\ln T.
\end{align*}
Another fact we will be using in the proof is from Theorem 1 in \cite{abbasi2011improved}, where it is shown that for any $\delta>0$, w.p. at least $1-\delta$, the following holds:
\begin{align*}
    (A_{t-1}\mathcal E_{t-1})^\top  V_{t-1}^{-1}A_{t-1}\mathcal E_{t-1} \leq 2\sigma^2\log\left(\frac{\det(V_{t-1})/\det(V_0)}{\delta}\right) \leq 2\sigma^2\log\left(\frac{(T/d)^{2d}}{\delta}\right)
\end{align*}
Thus, for any $y\geq 0$, we have
\begin{align*}
    \mbP\left(\sqrt{(A_{t-1}\mathcal E_{t-1})^\top  V_{t-1}^{-1}A_{t-1}\mathcal E_{t-1}} \geq x\right) \leq (T/d)^{2d}\exp\left(-\frac{x^2}{2\sigma^2}\right)
\end{align*}
We have, for any $\theta$, 
\begin{align} \label{eq:linear-decompose-any}
    \mbP(R_\theta^\pi(T) \geq x) & \leq \mbP\left(\bigcup_{t\geq 2}\left\{\Delta_t \geq \frac{x-1}{2t\ln T},\ \|a_t\|_t^2\leq d/t \right\}\right) + \mbP\left(\bigcup_{t\geq 2}\left\{\frac{\Delta_t}{\|a_t\|_t^2}\geq \frac{x-1}{4d\ln T}, \ \|a_t\|_t^2 > d/t\right\}\right)
\end{align}
The reason that \eqref{eq:linear-decompose-any} holds is as follows. To prove it, we only need to show that the following events cannot hold simultaneously:
\begin{align*}
    \Delta_t < \frac{x-1}{2t\ln T},\ \text{ if }\|a_t\|_t^2\leq d/t; \quad \frac{\Delta_t}{\|a_t\|_t^2} < \frac{x-1}{4d\ln T}, \ \text{ if }\|a_t\|_t^2 > d/t.
\end{align*}
If not, then
\begin{align*}
    R_{\theta}^\pi(T) & = \theta^\top(a_1^*-a_1) + \sum_{t\geq 2} \Delta_t\mathds 1\{\|a_t\|_t^2\leq d/t\} + \frac{\Delta_t}{\|a_t\|_t^2}\cdot \|a_t\|_t^2\mathds 1\{\|a_t\|_t^2 > d/t\} \\
    & < 1 + \sum_{t\geq 2}\frac{x-1}{2t\ln T} + \sum_{t\geq 2}\frac{x-1}{4d\ln T}\|a_t\|_t^2 \\
    & \leq 1 + \frac{x-1}{2} + \frac{x-1}{2} = x.
\end{align*}
This is a contradiction.
At time $t$, the policy takes action $a_t$, which means
\begin{align}
    & \quad\ \quad\ \hat\theta_{t-1}^\top a_t + \eta(t/d)^\alpha\sqrt{d}\|a_t\|_t^2\wedge\sqrt{f(t)}\|a_t\|_t + \sqrt{d\|a_t\|_t^2} \geq \nonumber\\
    & \quad\quad\quad \hat\theta_{t-1}^\top a_t^* + \eta(t/d)^\alpha\sqrt{d}\|a_t^*\|_t^2\wedge\sqrt{f(t)}\|a_t^*\|_t + \sqrt{d\|a_t^*\|_t^2} \nonumber\\
    & \Leftrightarrow\quad \theta^\top a_t - \theta^\top V_{t-1}^{-1}a_t + (V_{t-1}^{-1}A_{t-1}\mathcal E_{t-1})^\top a_t + \eta(t/d)^\alpha\sqrt{d}\|a_t\|_t^2\wedge\sqrt{f(t)}\|a_t\|_t + \sqrt{d\|a_t\|_t^2} \geq \nonumber\\
    & \quad\quad\quad \theta^\top a_t^* - \theta^\top V_{t-1}^{-1}a_t^* + (V_{t-1}^{-1}A_{t-1}\mathcal E_{t-1})^\top a_t^* + \eta(t/d)^\alpha\sqrt{d}\|a_t^*\|_t^2\wedge\sqrt{f(t)}\|a_t^*\|_t + \sqrt{d\|a_t^*\|_t^2} \nonumber\\
    & \Leftrightarrow\quad (V_{t-1}^{-1}A_{t-1}\mathcal E_{t-1})^\top a_t + \eta(t/d)^\alpha\sqrt{d}\|a_t\|_t^2\wedge\sqrt{f(t)}\|a_t\|_t + \sqrt{d\|a_t\|_t^2} - \theta^\top V_{t-1}^{-1}a_t \geq \nonumber\\
    & \quad\quad\quad \Delta_t + (V_{t-1}^{-1}A_{t-1}\mathcal E_{t-1})^\top a_t^* + \eta(t/d)^\alpha\sqrt{d}\|a_t^*\|_t^2\wedge\sqrt{f(t)}\|a_t^*\|_t + \sqrt{d\|a_t^*\|_t^2} - \theta^\top V_{t-1}^{-1}a_t^* \nonumber\\
    & \Rightarrow\quad a_t^\top V_{t-1}^{-1}A_{t-1}\mathcal E_{t-1} + \eta(t/d)^\alpha\sqrt{d}\|a_t\|_t^2\wedge\sqrt{f(t)}\|a_t\|_t + 2\sqrt{d\|a_t\|_t^2} \geq \nonumber\\
    & \quad\quad\quad \Delta_t + a_t^{*\top}V_{t-1}^{-1}A_{t-1}\mathcal E_{t-1} + \eta(t/d)^\alpha\sqrt{d}\|a_t^*\|_t^2\wedge\sqrt{f(t)}\|a_t^*\|_t \nonumber\\
    & \Rightarrow\quad a_t^\top V_{t-1}^{-1}A_{t-1}\mathcal E_{t-1} \geq \frac{\Delta_t}{2} - \eta(t/d)^\alpha\sqrt{d}\|a_t\|_t^2\wedge\sqrt{f(t)}\|a_t\|_t - 2\sqrt{d\|a_t\|_t^2} \quad \text{or} \nonumber\\
    & \quad\quad\quad -a_t^{*\top}V_{t-1}^{-1}A_{t-1}\mathcal E_{t-1} \geq \frac{\Delta_t}{2} + \eta(t/d)^\alpha\sqrt{d}\|a_t^*\|_t^2\wedge\sqrt{f(t)}\|a_t^*\|_t. \label{eq:linear-action-any}
\end{align}
Note that in \eqref{eq:linear-action-any} we use the following inequality: for any $a\in\cA_t$, 
\begin{align*}
    |\theta^\top V_{t-1}^{-1}a| \leq \sqrt{\theta^\top V_{t-1}^{-1}\theta}\sqrt{a^\top V_{t-1}^{-1}a} \leq \sqrt{d(a^\top V_{t-1}^{-1}a)}.
\end{align*}
Combining \eqref{eq:linear-decompose-any} and \eqref{eq:linear-action-any} yields {\footnotesize
\begin{align*}
    & \quad\; \mbP\left(R_\theta^\pi(T) \geq x/2\right) \\
    & \leq \sum_{t}\mbP\left(\Delta_t \geq \frac{x-1}{2t\ln T},\ \|a_t\|_t^2\leq d/t,\ a_t^\top V_{t-1}^{-1}A_{t-1}\mathcal E_{t-1} \geq \frac{\Delta_t}{2} - \eta(t/d)^\alpha\sqrt{d}\|a_t\|_t^2\wedge\sqrt{f(t)}\|a_t\|_t - 2\sqrt{d\|a_t\|_t^2}\right) \\
    & + \sum_t\mbP\left(\Delta_t \geq \frac{x-1}{2t\ln T},\ \|a_t\|_t^2\leq d/t,\ -a_t^{*\top}V_{t-1}^{-1}A_{t-1}\mathcal E_{t-1} \geq \frac{\Delta_t}{2} + \eta(t/d)^\alpha\sqrt{d}\|a_t^*\|_t^2\wedge\sqrt{f(t)}\|a_t^*\|_t\right) \\
    & + \sum_{t}\mbP\left(\frac{\Delta_t}{\|a_t\|_t^2}\geq \frac{x-1}{4d\ln T},\ \|a_t\|_t^2 > \frac{d}{t},  \ a_t^\top V_{t-1}^{-1}A_{t-1}\mathcal E_{t-1} \geq \frac{\Delta_t}{2} - \eta(t/d)^\alpha\sqrt{d}\|a_t\|_t^2\wedge\sqrt{f(t)}\|a_t\|_t - 2\sqrt{d\|a_t\|_t^2}\right) \\
    & + \sum_t\mbP\left(\frac{\Delta_t}{\|a_t\|_t^2}\geq \frac{x-1}{4d\ln T},\ \|a_t\|_t^2 > \frac{d}{t},\ -a_t^{*\top}V_{t-1}^{-1}A_{t-1}\mathcal E_{t-1} \geq \frac{\Delta_t}{2} + \eta(t/d)^\alpha\sqrt{d}\|a_t^*\|_t^2\wedge\sqrt{f(t)}\|a_t^*\|_t\right).
\end{align*}}
We bound each term separately. 

{\noindent\bf 1. Worst-case scenario.} We have
\begin{align*}
    & \quad\; \mbP\left(\Delta_t \geq \frac{x-1}{2t\ln T},\ \|a_t\|_t^2\leq d/t,\ a_t^\top V_{t-1}^{-1}A_{t-1}\mathcal E_{t-1} \geq \frac{\Delta_t}{2} - \eta(t/d)^\alpha\sqrt{d}\|a_t\|_t^2\wedge\sqrt{f(t)}\|a_t\|_t - 2\sqrt{d\|a_t\|_t^2}\right) \\
    & \leq \mbP\left(\|a_t\|_t^2\leq d/t,\ a_t^\top V_{t-1}^{-1}A_{t-1}\mathcal E_{t-1} \geq \frac{x-1}{4t\ln T} - \eta(d/t)^{1-\alpha}\sqrt{d} - 2d/\sqrt{t}\right) \\
    & \leq \mbP\left(\frac{|a_t^\top V_{t-1}^{-1}A_{t-1}\mathcal E_{t-1}|}{\sqrt{\|a_t\|_t^2}}\geq \frac{\left(\frac{x-1}{4t\ln T} - \eta(d/t)^{1-\alpha}\sqrt{d}-2d/\sqrt{t}\right)_+}{\sqrt{d/t}\ln T}\right) \\
    & \leq \mbP\left(\sqrt{(A_{t-1}\mathcal E_{t-1})^\top  V_{t-1}^{-1}A_{t-1}\mathcal E_{t-1}} \geq \frac{\left(x-1 - 8d\sqrt{T}\ln T - 4\eta d^{\frac{3}{2}-\alpha}T^\alpha\ln T\right)_+}{4\sqrt{dT}}\right) \\
    & \leq (T/d)^{2d}\exp\left(-\frac{\left(x-1 - 8d\sqrt{T}\ln T - 4\eta d^{\frac{3}{2}-\alpha}T^\alpha\ln T\right)_+^2}{32\sigma^2dT\ln^2T}\right)
\end{align*}
and
\begin{align*}
    & \quad\; \mbP\left(\Delta_t \geq \frac{x-1}{2t\ln T},\ \|a_t\|_t^2\leq d/t,\ -a_t^{*\top}V_{t-1}^{-1}A_{t-1}\mathcal E_{t-1} \geq \frac{\Delta_t}{2} + \eta(t/d)^\alpha\sqrt{d}\|a_t^*\|_t^2\wedge\sqrt{f(t)}\|a_t^*\|_t\right) \\
    & \leq \mbP\left(t\geq \frac{x-1}{2\Delta_t\ln T},\ -a_t^{*\top}V_{t-1}^{-1}A_{t-1}\mathcal E_{t-1} \geq \sqrt{2\frac{x-1}{2t\ln T}\eta(t/d)^\alpha\sqrt{d}\|a_t^*\|_t^2}\wedge\sqrt{f(t)}\|a_t^*\|_t\right) \\
    & \leq \mbP\left(\frac{|a_t^{*\top} V_{t-1}^{-1}A_{t-1}\mathcal E_{t-1}|}{\sqrt{\|a_t^*\|_t^2}} \geq \sqrt{\frac{(x-1)_+\eta\sqrt{d}}{2d^{\alpha}T^{1-\alpha}\ln T}}\wedge\sqrt{f(t)}, \ t\geq \frac{x-1}{4\sqrt{d}\ln T}\right) \\
    & \leq \mbP\left(\sqrt{(A_{t-1}\mathcal E_{t-1})^\top  V_{t-1}^{-1}A_{t-1}\mathcal E_{t-1}} \geq \frac{\sqrt{(x-1)_+\eta\sqrt{d}}}{\sqrt{2d^{\alpha}T^{1-\alpha}\ln T}}\wedge\sqrt{f(x\vee t)}\right) \\
    & \leq (T/d)^{2d}\exp\left(-\frac{\eta(x-1)_+}{4\sigma^2 d^{\alpha-\frac{1}{2}}T^{1-\alpha}\ln T}\right) + (T/d)^{2d}\exp\left(-\frac{f(x\vee t)}{2\sigma^2}\right)
\end{align*}
and
\begin{align*}
    & \quad\; \mbP\left(\frac{\Delta_t}{\|a_t\|_t^2}\geq \frac{x-1}{4d\ln T}, \ \|a_t\|_t^2 > \frac{d}{t}, \ a_t^\top V_{t-1}^{-1}A_{t-1}\mathcal E_{t-1} \geq \frac{\Delta_t}{2} - \eta(t/d)^\alpha\sqrt{d}\|a_t\|_t^2\wedge\sqrt{f(t)}\|a_t\|_t - 2\sqrt{d\|a_t\|_t^2}\right) \\
    & = \mbP\left(\frac{\Delta_t}{\|a_t\|_t^2}\geq \frac{x-1}{4d\ln T}, \ \|a_t\|_t^2 > \frac{d}{t}, \ \frac{a_t^\top V_{t-1}^{-1}A_{t-1}\mathcal E_{t-1}}{\sqrt{\|a_t\|_t^2}\sqrt{\|a_t\|_t^2}} \geq \frac{\Delta_t}{2\|a_t\|_t^2} - \eta(t/d)^\alpha\sqrt{d} - 2\sqrt{t}\right) \\
    & \leq \mbP\left(\|a_t\|_t^2 > \frac{d}{t}, \ \frac{a_t^\top V_{t-1}^{-1}A_{t-1}\mathcal E_{t-1}}{\sqrt{\|a_t\|_t^2}\sqrt{\|a_t\|_t^2}} \geq \frac{x-1}{8d\ln T} - \eta(t/d)^\alpha\sqrt{d} - 2\sqrt{t}\right) \\
    & \leq \mbP\left(\frac{|a_t^\top V_{t-1}^{-1}A_{t-1}\mathcal E_{t-1}|}{\sqrt{\|a_t\|_t^2}} \geq \left(\frac{x-1}{8d\ln T} - \eta(t/d)^\alpha\sqrt{d} - 2\sqrt{t}\right)_+\sqrt{\frac{d}{t}}\right) \\
    & \leq \mbP\left(\sqrt{(A_{t-1}\mathcal E_{t-1})^\top  V_{t-1}^{-1}A_{t-1}\mathcal E_{t-1}} \geq \frac{\left(x-1-16d\sqrt{T}\ln T - 8\eta d^{\frac{3}{2}-\alpha}T^\alpha\ln T\right)_+}{8\sqrt{dT}\ln T}\right) \\
    & \leq (T/d)^{2d}\exp\left(-\frac{\left(x-1-16d\sqrt{T}\ln T - 8\eta d^{\frac{3}{2}-\alpha}T^\alpha\ln T\right)_+^2}{128\sigma^2dT\ln^2T}\right)
\end{align*}
and
\begin{align*}
    & \quad\; \mbP\left(\frac{\Delta_t}{\|a_t\|_t^2}\geq \frac{x-1}{4d\ln T},\ \|a_t\|_t^2 > \frac{d}{t}, -a_t^{*\top}V_{t-1}^{-1}A_{t-1}\mathcal E_{t-1}\geq \frac{\Delta_t}{2} + \eta(t/d)^\alpha\sqrt{d}\|a_t^*\|_t^2\wedge\sqrt{f(t)}\|a_t^*\|_t\right) \\
    & \leq \mbP\left(-a_t^{*\top}V_{t-1}^{-1}A_{t-1}\mathcal E_{t-1}\geq \sqrt{2\frac{x-1}{4t\ln T}\eta(t/d)^\alpha\sqrt{d}\|a_t^*\|_t^2}\wedge\sqrt{f(t)}\|a_t^*\|_t\right) \\
    & \leq \mbP\left(\frac{|a_t^{*\top} V_{t-1}^{-1}A_{t-1}\mathcal E_{t-1}|}{\|a_t^*\|_t} \geq \sqrt{\frac{(x-1)\eta\sqrt{d}}{2d^\alpha T^{1-\alpha}\ln T}}\wedge \sqrt{f(t)}\right) \\
    & \leq \mbP\left(\sqrt{(A_{t-1}\mathcal E_{t-1})^\top  V_{t-1}^{-1}A_{t-1}\mathcal E_{t-1}} \geq \frac{\sqrt{(x-1)_+\eta\sqrt{d}}}{\sqrt{2d^\alpha T^{1-\alpha}\ln T}}\wedge\sqrt{f(x\vee t)}\right) \\
    & \leq (T/d)^{2d}\exp\left(-\frac{\eta(x-1)_+}{4\sigma^2 d^{\alpha-\frac{1}{2}} T^{1-\alpha}\ln T}\right) + (T/d)^{2d}\exp\left(-\frac{f(x\vee t)}{2\sigma^2}\right).
\end{align*}

Plugging the four bounds above into \eqref{eq:linear-decompose-any} yields the final result
\begin{align*}
    \sup_{\theta}\mbP(R_{\theta}^\pi(T)\geq x) & \leq 2d(T/d)^{2d+1}\exp\left(-\frac{\left(x-1-16d\sqrt{T}\ln T - 8\eta d^{\frac{3}{2}-\alpha}T^\alpha\ln T\right)_+^2}{128\sigma^2dT\ln^2T}\right) \\
    & \quad\quad + 2d(T/d)^{2d+1}\exp\left(-\frac{\eta(x-1)_+}{4\sigma^2 d^{\alpha-\frac{1}{2}} T^{1-\alpha}\ln T}\wedge\frac{f(x)}{2\sigma^2}\right).
\end{align*}

{\noindent\bf 2. Instance-dependent scenario.} We have
\begin{align*}
    & \quad\; \mbP\left(\Delta_t \geq \frac{x-1}{2t\ln T},\ \|a_t\|_t^2\leq d/t,\ a_t^\top V_{t-1}^{-1}A_{t-1}\mathcal E_{t-1} \geq \frac{\Delta_t}{2} - \eta(t/d)^\alpha\sqrt{d}\|a_t\|_t^2\wedge\sqrt{f(t)}\|a_t\|_t - 2\sqrt{d\|a_t\|_t^2}\right) \\
    & \leq \mbP\left(\Delta_t \geq \frac{x-1}{2t\ln T},\ \|a_t\|_t^2\leq d/t,\ \frac{a_t^\top V_{t-1}^{-1}A_{t-1}\mathcal E_{t-1}}{\|a_t\|_t} \geq \frac{1}{2}\sqrt{\Delta_t}\cdot\sqrt{\frac{\Delta_t}{\|a_t\|_t^2}} - \sqrt{f(t)} - 2\sqrt{d}\right) \\
    & \leq \mbP\left(\frac{|a_t^\top V_{t-1}^{-1}A_{t-1}\mathcal E_{t-1}|}{\sqrt{\|a_t\|_t^2}}\geq \left(\frac{1}{2}\sqrt{\Delta}\cdot\sqrt{\frac{x-1}{2d\ln T}} - \sqrt{f(t)} - 2\sqrt{d}\right)_+\right) \\
    & \leq \mbP\left(\sqrt{(A_{t-1}\mathcal E_{t-1})^\top  V_{t-1}^{-1}A_{t-1}\mathcal E_{t-1}} \geq \frac{\sqrt{\Delta(x-1)} - 2\sqrt{2}\sqrt{f(t)} - 4\sqrt{2}\sqrt{d}}{2\sqrt{2d\ln T}}\right) \\
    & \leq (T/d)^{2d}\exp\left(-\frac{\left(\Delta(x-1)/4 - 64d - 16f(t)\right)_+}{16\sigma^2d\ln T}\right)
\end{align*}
and
\begin{align*}
    & \quad\; \mbP\left(\Delta_t \geq \frac{x-1}{2t\ln T},\ \|a_t\|_t^2\leq d/t,\ -a_t^{*\top}V_{t-1}^{-1}A_{t-1}\mathcal E_{t-1} \geq \frac{\Delta_t}{2} + \eta(t/d)^\alpha\sqrt{d}\|a_t^*\|_t^2\wedge\sqrt{f(t)}\|a_t^*\|_t\right) \\
    & \leq \mbP\left(-a_t^{*\top}V_{t-1}^{-1}A_{t-1}\mathcal E_{t-1} \geq \sqrt{\Delta\eta(t/d)^\alpha\sqrt{d}\|a_t^*\|_t^2}\wedge\sqrt{f(t)}\|a_t^*\|_t, \ t\geq \frac{x-1}{2\Delta_t\ln T}\right) \\
    & \leq \mbP\left(\frac{|a_t^{*\top} V_{t-1}^{-1}A_{t-1}\mathcal E_{t-1}|}{\sqrt{\|a_t^*\|_t^2}} \geq \sqrt{\Delta\eta(t/d)^\alpha\sqrt{d}}\wedge\sqrt{f(t)}\right) \\
    & \leq \mbP\left(\sqrt{(A_{t-1}\mathcal E_{t-1})^\top  V_{t-1}^{-1}A_{t-1}\mathcal E_{t-1}} \geq \frac{\sqrt{\eta\Delta x^\alpha\sqrt{d}}}{d^{\frac{\alpha}{2}}}\wedge\sqrt{f(x\vee t)}\right) \\
    & \leq (T/d)^{2d}\exp\left(-\frac{\eta\Delta x^\alpha}{2\sigma^2 d^{\alpha-\frac{1}{2}}}\right) + (T/d)^{2d}\exp\left(-\frac{f(x\vee t)}{2\sigma^2}\right)
\end{align*}
and
\begin{align*}
    & \quad\; \mbP\left(\frac{\Delta_t}{\|a_t\|_t^2}\geq \frac{x-1}{4d\ln T}, \ \|a_t\|_t^2 > \frac{d}{t}, \ a_t^\top V_{t-1}^{-1}A_{t-1}\mathcal E_{t-1} \geq \frac{\Delta_t}{2} - \eta(t/d)^\alpha\sqrt{d}\|a_t\|_t^2\wedge\sqrt{f(t)}\|a_t\|_t - 2\sqrt{d\|a_t\|_t^2}\right) \\
    & \leq \mbP\left(\frac{\Delta_t}{\|a_t\|_t^2}\geq \frac{x-1}{4d\ln T},\ \|a_t\|_t^2>d/t,\ \frac{a_t^\top V_{t-1}^{-1}A_{t-1}\mathcal E_{t-1}}{\|a_t\|_t} \geq \frac{1}{2}\sqrt{\Delta_t}\cdot\sqrt{\frac{\Delta_t}{\|a_t\|_t^2}} - \sqrt{f(t)} - 2\sqrt{d}\right) \\
    & \leq \mbP\left(\frac{|a_t^\top V_{t-1}^{-1}A_{t-1}\mathcal E_{t-1}|}{\sqrt{\|a_t\|_t^2}}\geq \left(\frac{1}{2}\sqrt{\Delta}\cdot\sqrt{\frac{x-1}{4d\ln T}} - \sqrt{f(t)} - 2\sqrt{d}\right)_+\right) \\
    & \leq \mbP\left(\sqrt{(A_{t-1}\mathcal E_{t-1})^\top  V_{t-1}^{-1}A_{t-1}\mathcal E_{t-1}} \geq \frac{\sqrt{\Delta(x-1)} - 4\sqrt{f(t)} - 8\sqrt{d}}{4\sqrt{d\ln T}}\right) \\
    & \leq (T/d)^{2d}\exp\left(-\frac{\left(\Delta(x-1)/4 - 128d - 32f(t)\right)_+}{32\sigma^2d\ln T}\right)
\end{align*}
and
\begin{align*}
    & \quad\; \mbP\left(\frac{\Delta_t}{\|a_t\|_t^2}\geq \frac{x-1}{4d\ln T},\ \|a_t\|_t^2 > \frac{d}{t}, -a_t^{*\top}V_{t-1}^{-1}A_{t-1}\mathcal E_{t-1}\geq \frac{\Delta_t}{2} + \eta(t/d)^\alpha\sqrt{d}\|a_t^*\|_t^2\wedge\sqrt{f(t)}\|a_t^*\|_t\right) \\
    & \leq \mbP\left(-a_t^{*\top}V_{t-1}^{-1}A_{t-1}\mathcal E_{t-1}\geq \sqrt{\Delta_t\eta(t/d)^\alpha\sqrt{d}\|a_t^*\|_t^2}\wedge\sqrt{f(t)}\|a_t^*\|_t, \ \Delta_t\geq \frac{x-1}{4t\ln T}\right) \\
    & \leq \mbP\left(\frac{|a_t^{*\top} V_{t-1}^{-1}A_{t-1}\mathcal E_{t-1}|}{\sqrt{\|a_t^*\|_t^2}} \geq \sqrt{\Delta\eta(t/d)^\alpha\sqrt{d}}\wedge\sqrt{f(t)}\right) \\
    & \leq \mbP\left(\sqrt{(A_{t-1}\mathcal E_{t-1})^\top  V_{t-1}^{-1}A_{t-1}\mathcal E_{t-1}} \geq \frac{\sqrt{\eta\Delta x^\alpha\sqrt{d}}}{d^{\frac{\alpha}{2}}}\wedge\sqrt{f(x\vee t)}\right) \\
    & \leq (T/d)^{2d}\exp\left(-\frac{\eta\Delta x^\alpha}{2\sigma^2d^{\alpha-\frac{1}{2}}}\wedge\frac{f(x)}{2\sigma^2}\right) + (T/d)^{2d}\exp\left(-\frac{f(x\vee t)}{2\sigma^2}\right).
\end{align*}

Plugging the four bounds above into \eqref{eq:linear-decompose-any} yields the final result
\begin{align*}
    \sup_{\theta}\mbP(R_{\theta}^\pi(T)\geq x) & \leq 2d(T/d)^{2d+1}\exp\left(-\frac{\left(\Delta(x-1)/4 - 128d - 32f(t)\right)_+}{32\sigma^2d\ln T}\right) \\
    & \quad\quad + 2d(T/d)^{2d+1}\exp\left(-\frac{\eta\Delta x^\alpha}{2\sigma^2 d^{\alpha-\frac{1}{2}}}\right) + 2(T/d)^{2d}\int_{0}^T\exp\brs{-\frac{f(x\vee y)}{2\sigma^2}}dy.
\end{align*}

$\hfill\Box$

\end{APPENDICES}
%
%


\end{document}